\documentclass{article}

\PassOptionsToPackage{round}{natbib}

\usepackage[preprint]{neurips_2026}

\usepackage[utf8]{inputenc} 
\usepackage[T1]{fontenc}    
\usepackage{url}            
\usepackage{booktabs}       
\usepackage{amsfonts}       
\usepackage{nicefrac}       
\usepackage{microtype}      
\usepackage{xcolor}         

\usepackage{tabularx}
\usepackage{makecell}
\usepackage{array}
\newcolumntype{Y}{>{\raggedright\arraybackslash}X}

\usepackage{ninecolors}
\usepackage[colorlinks=true,
linkcolor=blue4,
filecolor=blue4,
urlcolor=blue4,
citecolor=blue4]{hyperref}

\usepackage{algorithmic}
\usepackage{algorithm}

\author{%
	Dorian Baudry\\
Univ. Grenoble Alpes, Inria, CNRS, Grenoble INP, LIG, 38000 Grenoble, France\\
	\texttt{dorian.baudry@inria.fr}\\
	\And
	Nathan Boyer\\
	ENS Paris-Saclay, France\\
	\And
	Patrick Rebeschini\\
 Department of Statistics, University of Oxford
}

\usepackage{amsmath}
\usepackage{amssymb}
\usepackage{mathtools}
\usepackage{amsthm}
\usepackage{thmtools}
\usepackage{thm-restate}

\usepackage{tocloft}
\addtocontents{toc}{\protect\setcounter{tocdepth}{0}}

\newcommand{\XMAX}{X_{\mathrm{max}}}

\usepackage{ninecolors}
\newcommand{\antoine}[1]{%
    \ifmmode
    \text{\textcolor{green6}{[Antoine: #1]}}
    \else
    \textcolor{green6}{[Antoine: #1]}
    \fi
}

\newcommand{\tmpassumption}[1]{%
    \ifmmode
    \text{\textcolor{orange}{[Assumption: #1]}}
    \else
    \textcolor{orange}{[Assumption: #1]}
    \fi
}

\usepackage{xspace}
\makeatletter
\DeclareRobustCommand\onedot{\futurelet\@let@token\@onedot}
\def\@onedot{\ifx\@let@token.\else.\null\fi\xspace}

\makeatother

\DeclarePairedDelimiter{\spr}{(}{)}
\DeclarePairedDelimiter{\sbr}{[}{]}

\DeclarePairedDelimiter{\scbr}{\{}{\}}
\DeclarePairedDelimiter{\inp}{\langle}{\rangle}
\DeclarePairedDelimiter{\norm}{\|}{\|}
\DeclarePairedDelimiter{\abs}{|}{|}

\newcommand{\sumaK}{\sum_{a=1}^K}

\newcommand{\sumtT}{\sum_{t=1}^T}
\newcommand{\sumst}{\sum_{s=1}^t}
\newcommand{\sumtn}{\sum_{t=1}^n}

\usepackage{mathtools} 
\newcommand{\given}{\nonscript\mathpunct{}\middle|\nonscript\mathpunct{}}

\newcommand{{\transpose}}{^\mathsf{\scriptscriptstyle T}}

\newcommand{\wh}[1]{\widehat{#1}}
\newcommand{\wt}[1]{\widetilde{#1}}

\usepackage{mathtools}  

\newcommand{\bbE}{\mathbb{E}}

\newcommand{\bbN}{\mathbb{N}}
\newcommand{\bbP}{\mathbb{P}}

\newcommand{\bbR}{\mathbb{R}}
\newcommand{\bbRnnK}{\mathbb{R}_{\geq 0}^K}
\newcommand{\bbRpK}{\mathbb{R}_{>0}^K}

\newcommand{\bfone}{\mathbf{1}}

\newcommand{\cA}{\mathcal{A}}

\newcommand{\cD}{\mathcal{D}}

\newcommand{\cH}{\mathcal{H}}

\newcommand{\cL}{\mathcal{L}}

\newcommand{\cO}{\mathcal{O}}

\newcommand{\cX}{\mathcal{X}}

\DeclareMathOperator*{\argmax}{arg\,max}
\DeclareMathOperator*{\argmin}{arg\,min}

\DeclareMathOperator{\tr}{tr}
\DeclareMathOperator{\Diag}{Diag}  

\newcommand{\simplex}{\Delta}

\newcommand{\pen}{\mathtt{penalty}_T}

\newcommand{\stability}{\mathtt{stability}_T}

\newcommand{\elltiw}{\widehat{\ell}_t^{\texttt{IW}}}
\newcommand{\thetaiw}{\widehat{\theta}^{\texttt{IW}}}
\newcommand{\thetataiw}{\widehat{\theta}_{t, a}^{\texttt{IW}}}

\newcommand{\Thetataiw}{\widehat{\Theta}_{t, a}^{\texttt{IW}}}

\newcommand{\htmax}{h_t^{\max}}
\usepackage{bm}

\newcommand{\bE}{\bbE}
\usepackage{dsfont}
\newcommand{\ind}{\mathds{1}}

\let\epsilon\relax
\DeclareMathOperator{\epsilon}{\varepsilon}

\usepackage{tikz}
\usetikzlibrary{decorations.pathreplacing,calc}

\usepackage{soul}

\usepackage{comment}
\usepackage{enumitem}

\usepackage[textsize=tiny]{todonotes}

\usepackage[capitalize,noabbrev]{cleveref}

\theoremstyle{plain}
\declaretheorem[numberwithin=section,name=Theorem]{theorem}
\declaretheorem[sibling=theorem,name=Proposition]{proposition}
\declaretheorem[sibling=theorem,name=Lemma]{lemma}

\theoremstyle{definition}

\theoremstyle{remark}
\declaretheorem[sibling=theorem,name=Remark]{remark}

\usepackage[textsize=tiny]{todonotes}

\newcommand{\agg}{\texttt{Agg--SPB}}
\newcommand{\loc}{\texttt{CE--SPB}}
\newcommand{\obs}{\texttt{Obs}_T}

\title{Linear contextual bandits with paid observations}

\begin{document}
	\maketitle
	
	
	
	\begin{abstract}
		We study linear contextual bandits with paid observations. At each round, the learner observes a context, chooses an action, and may pay a fixed cost to observe feedback from any subset of arms. We develop two Follow-the-Regularized-Leader algorithms with Best-of-Both-Worlds guarantees. A first approach, \agg, extends the SPB-matching framework of \citet{BOBWhardproblems} by aggregating context-dependent stability quantities, and achieves the characteristic adversarial $T^{2/3}$ rate, together with logarithmic dependence on $T$ in stochastic environments. A second approach, \loc, instead uses arm-dependent observation probabilities and an entropy-adaptive learning rate inspired by \citet{BOBWlinear}. It achieves an entropy-adaptive $\widetilde O(T^{2/3})$ adversarial guarantee and polylogarithmic stochastic regret, while avoiding the minimum-context-mass dependence appearing in the stochastic analysis of \agg. Both methods extend to corrupted stochastic environments with explicit corruption-dependent guarantees. Our results show that logarithmic stochastic regret is compatible with the $T^{2/3}$ adversarial regime in linear contextual bandits with paid observations, and expose a tradeoff between sharp stochastic dependence on the horizon and a path-dependent matching mechanism without explicit minimum-context-mass dependence.
	\end{abstract}

	\section{Introduction} \label{sec:introduction}

Multi-armed bandits (MAB) provide a canonical model for sequential decision-making under uncertainty \citep{BanditBook, BubeckCbook12}: at each round, a learner selects an action and receives feedback only for that action. While this captures the exploration--exploitation tradeoff central in many problems, modern interactive systems often require richer models. In particular, rewards may depend on contextual information, feedback may be costly to acquire, and the environment may not be reliably stochastic.

A motivating example is online recommendation. The value of recommending a video, article, or product depends on the user receiving it, which naturally leads to contextual bandit models \citep{AbeL99, BeygelzimerLLRS11}. A widely studied instance is the linear contextual bandit model \citep{LangfordZ07, LiCLS10}, where the expected reward of each arm is a linear function of the observed context. This model captures heterogeneity in user preferences while preserving enough structure for efficient learning.

In many applications, however, feedback is not automatic. A user may only provide a rating, review, or response if prompted, and requesting such feedback too often may be costly or intrusive. This motivates bandits with paid observations \citep{paidobservations}, where the learner may purchase feedback at a fixed cost and must decide not only which action to play, but also when information is worth acquiring.

Finally, the data-generating process itself may range from a fixed stochastic source to an adversarial one. This has motivated Best-of-Both-Worlds (BoBW) algorithms, which aim to retain adversarial robustness while exploiting stochastic structure when it is present \citep{bubeck12bobw, SeldinS14, zimmerttsallis2022, DannWZ23, BOBWhardproblems}. In the paid-observation setting, this objective has a particularly natural interpretation: in a stochastic environment, the learner should eventually become confident enough to purchase feedback only rarely, whereas in an adversarial environment it must continue acquiring enough information to remain robust.

In this work, we study \emph{linear contextual bandits with paid observations} under adversarial and stochastic/corrupted-stochastic regimes. We show that these problems admit BoBW guarantees: one can retain the characteristic $T^{2/3}$ adversarial behavior of paid-observation bandits while obtaining only logarithmic or polylogarithmic dependence on the horizon in stochastic environments.

\paragraph{Outline and contributions.}
Bandits with paid observations are intrinsically ``hard'' in the sense that, already in the non-contextual case, paying for information changes the optimal horizon dependence from the usual $\sqrt{T}$ regime to $T^{2/3}$, up to logarithmic factors \citep{paidobservations}. This places them in the regime studied by \citet{BOBWhardproblems}, who developed a \emph{stability--penalty--bias} (SPB) matching FTRL framework for best-of-both-worlds guarantees in such multi-armed problems. In parallel, \citet{BOBWlinear} developed entropy-adaptive FTRL algorithms for linear contextual bandits.

Combining these two lines is not immediate. In paid-observation problems, the learner must decide how much feedback to acquire in order to balance estimation variance against observation cost. In the contextual setting, this tradeoff is coupled through the context distribution: the observation probabilities induce weighted covariance matrices whose inverses govern the variance of the linear loss estimates across contexts. Consequently, the scalar arm-wise SPB matching used by \citet{BOBWhardproblems} in the multi-armed setting no longer applies directly, while the entropy-adaptive contextual analysis must be combined with a new observation mechanism and a refined stability analysis.

We address this difficulty through two complementary exact importance-weighted FTRL constructions, corresponding to two different ways of handling this contextual coupling.

The first, \emph{Aggregate-SPB} (\agg, Algorithm~\ref{alg:clipped-aggregate-spb-iw}), follows the hard-problem SPB framework of \citet{BOBWhardproblems}. It aggregates the context-dependent stability quantities over the context distribution, thereby recovering a scalar SPB structure and enabling a \emph{direct} matching of stability, penalty, and observation cost at each round. In \Cref{thm::main}, we prove the characteristic adversarial $T^{2/3}$ rate up to logarithmic and problem-dependent factors, logarithmic dependence on $T$ in stochastic environments, and explicit corruption-dependent guarantees. This construction shows in particular that logarithmic stochastic regret is compatible with the $T^{2/3}$ adversarial regime in linear contextual bandits with paid observations.

The second approach, \emph{Cumulative-Entropy SPB} (\loc, Algorithm~\ref{alg:kuroki-style}), follows the entropy-adaptive route of \citet{BOBWlinear}. Rather than collapsing the contextual geometry into a global scalar quantity, it uses arm and context-dependent observation probabilities together with an entropy-driven learning rate. Stability, penalty, and observation cost are then controlled through the same entropy-dependent regret bound, yielding an \emph{indirect} form of SPB matching, at the final horizon scale. In \Cref{thm:le-spb}, we obtain an entropy-adaptive $\widetilde O(T^{2/3})$ adversarial guarantee and polylogarithmic stochastic regret, again with explicit corruption dependence.

The two approaches perform similarly in the adversarial regime: under fixed cost $c>0$, both obtain a $(cKd)^{1/3}T^{2/3}$ regret upper bound up to logarithmic factors. Their main differences emerge in stochastic environments. \agg{} achieves the sharper $\log T$ horizon dependence, but its self-bounding argument incurs a potentially severe dependence on the context distribution, through the minimum context mass $\mu_{\min}$. In contrast, \loc{} avoids this worst-context dependence, but pays additional logarithmic factors in $T$. Thus, the two constructions expose complementary limitations: aggregate SPB matching is statistically sharp in the horizon but can be unfavorable in the context dimension and support geometry, whereas the matching mechanism based on cumulative entropy has more favorable distributional dependence but a looser stochastic time dependence.

This comparison suggests a clear target for further work: retain the local structure and context-distribution robustness of \loc{} while recovering the logarithmic stochastic scaling exhibited by \agg. The lower-bound discussion in \Cref{subsec:lower-bound-perspective} further shows that the leading adversarial $(cKd)^{1/3}T^{2/3}$ term and stochastic $\frac{cKd}{\Delta^{2}}\;\log T$ information-acquisition term have the expected dependence on the main problem parameters.

The rest of the paper is organized as follows. \Cref{sec::setting} formalizes the linear contextual paid-observation model and the adversarial, stochastic, and corrupted-stochastic regimes. \Cref{sec::algorithm} introduces the shared exact-IW FTRL template and the two instantiations, \agg\ and \loc. \Cref{sec::regret} states their regret guarantees and the accompanying lower-bound benchmarks. Finally, \Cref{sec:discussion} compares the two approaches in light of these benchmarks, while \Cref{subsec:perspectives} discusses their limitations and the main directions for further work.

\subsection{Detailed literature review}

In this section we detail existing results related to the different aspects of the setting that we consider.

\paragraph{Linear Contextual Bandits}
Contextual bandits extend classical multi-armed bandits by allowing the reward distribution to depend on an observed context, which can vary across rounds. To enable efficient decision-making, one must adopt a suitable model to capture how the context influences rewards. In this work we consider the \emph{linear contextual bandit} (LCB) model \citep{LangfordZ07, LiCLS10}. It is closely related to the well-studied \emph{stochastic linear bandit} (SLB) framework: in both settings the average reward is represented by an inner product. However, in SLB actions are associated with known feature vectors and rewards depend on a shared unknown parameter, whereas in our linear contextual setting the context is observed and each arm has its own unknown parameter vector. Most approaches used in linear contextual bandits are adapted from the literature on SLB, including approaches based on \emph{Optimism in Face of Uncertainty} \citep{AbeL99, DaniHK08, Abbasi-YadkoriPS11, FlynnRKP23}, \emph{Thompson Sampling} \citep{AgrawalG13TSlinear, AbeilleL17, AbeilleJP25}, or \emph{information quantification} \citep{LattimoreS17, DegenneSK20, KirschnerL020}. Nonetheless, linear contextual bandits exhibit specific properties compared to stochastic linear bandits. In particular, \citet{BastaniBK21} showed that under suitable assumptions on \emph{context diversity}, even a simple greedy strategy can achieve logarithmic regret.

While the above works assume stochastic rewards, this assumption can be restrictive in practice. To address this, \citet{NeuO20} introduced an adversarial formulation of linear contextual bandits, in which arm parameters are fixed by an oblivious adversary. They derived a $\sqrt{KdT}$ regret bound for an exponential-weights algorithm \citep{auer2002nonstochastic}, where $d$ is the parameter dimension, $K$ is the number of arms, and $T$ is the horizon. Building on this, \citet{OlkhovskayaMENW23} obtained refined first- and second-order bounds. Best-of-Both-Worlds guarantees for linear contextual bandits were subsequently developed in the recent works of \citet{BOBWlinear} and \citet{KatoI24}, showing that polylogarithmic regret in stochastic environments can be combined with $\widetilde O(\sqrt{T})$ dependence in adversarial environments. More recently, \citet{KatoI25} introduced LC-Tsallis-INF, an FTRL algorithm based on Tsallis entropy that achieves $O(\log T)$ regret under a uniform positive-gap condition while retaining $O(\sqrt{T})$ adversarial dependence. Their analysis also extends to margin conditions, which relax the requirement that the suboptimality gap be uniformly bounded away from zero.

\paragraph{Bandits with Paid Observations.}
This framework was introduced by \citet{paidobservations} to capture a feedback structure lying between the standard multi-armed bandit and full-information settings. In this model, the learner may choose to observe the reward of \emph{any} arm at a fixed cost. They prove matching upper and lower bounds of order $T^{2/3}$. Prior to this, several related models were proposed to account for the possibility of observing additional feedback beyond the chosen arm \citep{MannorS11, AvnerMS12, AlonCGM13}, though these formulations do not explicitly capture the cost of information acquisition.

A closely related model was subsequently studied by \citet{TuckerBWJ23}, who consider \emph{costly reward observations}: the learner pays a cost to observe the reward of the action it selected. They establish an $\Omega(c^{1/3}T^{2/3})$ lower bound and a matching general non-adaptive construction, and also propose adaptive algorithms for multi-armed and contextual bandits. Their feedback model differs from that of \citet{paidobservations} and from ours, since the decision to acquire feedback concerns the played action rather than an independently chosen set of arms.

Another approach is to impose an explicit \emph{budget} on feedback acquisition. \citet{YunPASY18} study additional arm observations with acquisition costs subject to a total budget, while \citet{EfroniMSM21} consider a possibly time-dependent hard budget on the number of reward queries, including in linear contextual bandits. In such budget-constrained formulations, information acquisition is treated as an external resource constraint. By contrast, the bandits-with-paid-observations framework places learning performance and information acquisition on the same scale by incorporating observation costs directly into the objective.

\paragraph{Best-of-Both-Worlds (BoBW).}
The design of algorithms that perform well simultaneously in stochastic and adversarial regimes has become a central theme in the bandit literature. The foundational work of \citet{bubeck12bobw, SeldinS14} initiated this line of research by asking whether one can achieve logarithmic regret in the stochastic setting while retaining $\sqrt{T}$ regret in the adversarial case. Their results provided only partial success, either with suboptimal bounds or with algorithms of limited practicality. Later, \citet{zimmerttsallis2022} first obtained the optimal best-of-both-worlds guarantees in the $K$-armed bandit setting. This breakthrough has since inspired the development of BoBW algorithms across a variety of bandit problems \citep{amir2022better, RouyerHCS22, saha2022versatile, tsuchiya2023further, jin2023improved, zimmert24prod, KatoI25}.

Of particular relevance to our work, \citet{BOBWlinear} developed two complementary approaches for linear contextual bandits. Their black-box reduction based on \citet{DannWZ23} yields refined data-dependent BoBW guarantees, while their FTRL-LC algorithm uses a Shannon-entropy regularizer together with context-dependent loss estimation and obtains polylogarithmic stochastic regret alongside $\widetilde O(dK\sqrt{T})$ adversarial regret. Independently, \citet{KatoI24} obtained BoBW guarantees for linear contextual bandits under adversarial corruption through an FTRL-based approach. The subsequent LC-Tsallis-INF algorithm of \citet{KatoI25} is particularly related to our regularization strategy: it replaces Shannon entropy by Tsallis entropy, sharpens the stochastic dependence to $O(\log T)$ under a uniform gap condition, and extends the analysis to more general margin conditions.

These works concern linear contextual bandits whose adversarial minimax regret scales as $\sqrt{T}$. In contrast, \citet{BOBWhardproblems} proposed an SPB/FTRL recipe for ``hard'' online learning problems, namely those with minimax regret of order $T^{2/3}$, and use multi-armed bandits with paid observations as an application of their framework. Our setting lies at the intersection of these two lines: it combines the covariance geometry of linear contextual feedback with the variance--cost tradeoff of paid observations. Since neither framework applies off the shelf, we therefore develop and compare two exact importance-weighted adaptations: \agg, which extends hard-problem SPB matching through aggregate context-distribution quantities, and \loc, which adapts the entropy-driven contextual route through local observation probabilities.
	\section{Problem Definition} \label{sec::setting}

In this section we formalize the setting of \emph{linear contextual bandits with paid observations}, and state the standing assumptions used in the exact importance-weighted analysis.

\paragraph{Interaction protocol.}
The learner interacts with the environment for $T\in\bbN$ rounds. Contexts are drawn i.i.d. from a fixed distribution $\cD$ fully supported on a \emph{finite}\footnote{This assumption is only needed for the stochastic bound of \agg.} set $\cX\subseteq\bbR^d$. At each round $t\in [T]$, the loss parameters $\theta_{t,a}\in\bbR^d$ define the conditional mean loss of each arm $a\in[K]$, for $K\geq 2$. A context $X_t\sim\cD$ is drawn and revealed to the learner, who then chooses both an action $A_t\in[K]$ and an observation set $O_t\subseteq[K]$. The learner incurs mean loss $\ell_t(X_t,A_t)+c\abs{O_t}$ and observes the queried feedback scalars $\{Y_t(o):o\in O_t\}$. Here $c\ge0$ is the known unit cost of observation.

Following \citet{paidobservations}, the learner may query multiple arms in each round, paying $c$ per queried arm. As in \citet{BOBWhardproblems}, for simplicity of the presentation we assume a common observation cost $c$ for all arms. However, this assumption can be relaxed in our results (see Section~\ref{sec:discussion}), where $c$ can essentially be replaced by the arm-averaged cost $\bar c$.

\paragraph{Linear losses and environment regimes.}
For every round $t$, context $x\in\cX$, and arm $a\in[K]$, the conditional mean loss is linear: $\ell_t(x,a)=\inp{x,\theta_{t,a}}$.
The observed scalar for a queried arm satisfies
\begin{equation*}
    Y_t(a)\in[-1,1]\quad\text{a.s.},
    \qquad
    \bbE\sbr*{Y_t(a)\mid\cH_{t-1},X_t}
    =
    \ell_t(X_t,a).
\end{equation*}
Exact mean-loss observations correspond to the special case $Y_t(a)=\ell_t(X_t,a)$. The almost-sure boundedness of $Y_t(a)$ is part of the model used in the exact-IW stability proofs; bounded conditional means alone would not suffice. We consider three regimes. In the adversarial regime, the sequence $(\theta_{t,a})_{t\in[T],a\in[K]}$ is arbitrary but fixed before the interaction starts (oblivious adversary). In the stochastic regime, there exist fixed unknown vectors $(\theta_a)_{a\in[K]}$ such that $\theta_{t,a}=\theta_a$ for all $t$ and $a$. In the corrupted stochastic regime, the entire corrupted sequence $(\theta_{t,a})_{t\in[T],a\in[K]}$ is also fixed before the interaction starts (oblivious), and may deviate from such a fixed stochastic instance with the total deviation controlled as specified below.

For any predictable observation rule and each round $t$, write $p_{t,a}(x):=\bbP_t\sbr*{a\in O_t\mid X_t=x}$. We assume the observation randomization has no selection bias with respect to the feedback noise: for every arm $a$,
\begin{equation*}
    \bbE\sbr*{Y_t(a)\ind\scbr*{a\in O_t}\mid\cH_{t-1},X_t}
    =
    p_{t,a}(X_t)\ell_t(X_t,a).
\end{equation*}
This condition is implied, for example, by $O_t\perp (Y_t(1),\ldots,Y_t(K))\mid \cH_{t-1},X_t$ together with the conditional-mean assumption above.

\paragraph{Assumptions.}
We assume that $\ell_t(x,a)\in[-1,1]$ for all $(t,x,a)\in[T]\times\cX\times[K]$, consistently with the bounded-feedback condition above. Since $\cX$ is finite, we can write $\XMAX:=\max_{x\in\cX}\norm{x}_2$. Since $\cD$ is fully supported on $\cX$, we further define
\begin{equation*}
	\mu_{\min} := \min_{x\in\cX}\bbP[X=x] > 0, \quad 	\Sigma := \bbE_{X\sim\cD}\sbr{XX\transpose}\succ0, \text{ and} \quad 	B_\star := \sup_{x,y\in\cX}\abs{x\transpose\Sigma^{-1}y}<+\infty.
\end{equation*}

The exact-IW guarantees assume exact access to the distributional quantities used by the algorithms: the learner can form $\Sigma^{-1}$, use $B_\star$ or a valid upper bound for tuning, and evaluate the expectations over $X\sim\cD$ appearing in the update rules. Since $\cX$ is finite, these expectations reduce to finite sums. Approximate covariance inverses or Monte Carlo estimates of the aggregate quantities are not covered by the theorems below: the stability and penalty arguments use one-sided inequalities that can be broken by uncontrolled perturbations. Such approximations require a separate sensitivity analysis. In the following, we therefore assume that all inverse-covariance and distributional tuning quantities are computed exactly.

\paragraph{Stochastic and corrupted stochastic regimes.}
The stochastic regime is the special case where $\theta_{t,a}=\theta_a$ for fixed unknown vectors $(\theta_a)_{a\in[K]}$. For the gap-dependent guarantees, we assume positive context mass and a unique optimal arm at every context. We write $a^\star(x):=\argmin_{a\in[K]}\inp{x,\theta_a}$,
and assume a non-zero minimum gap,
\begin{equation*}
	\Delta_{\min}
	:=
	\min_{x\in\cX}\min_{a\ne a^\star(x)}
	\inp{x,\theta_a-\theta_{a^\star(x)}} > 0 .
\end{equation*}
The corrupted stochastic regime is defined relative to the same fixed vectors $(\theta_a)_{a\in[K]}$: while the actual parameters $(\theta_{t,a})$ may vary with time, their total deviation is measured by
\begin{equation*}
	C_\theta
	:=
	\sum_{t=1}^T \max_{a\in[K]}\norm{\theta_{t,a}-\theta_a}_2,
	\qquad
	C_{\mathrm{corr}} := 2\XMAX C_\theta.
\end{equation*}
The uncorrupted stochastic regime corresponds to $C_{\mathrm{corr}}=0$.

\paragraph{Regret.} Let $\Pi$ be the set of deterministic policies $\pi:\cX\to[K]$. The comparator is
\[
\pi_T^\star(x)
\in
\argmin_{a\in[K]}
\sum_{t=1}^T \ell_t(x,a) \qquad \text{(best contextual policy in hindsight)},
\]
with ties broken arbitrarily. The learner is evaluated by its expected regret against $\pi_T^\star$,
\begin{equation}\label{eq::regret_def}
R_T
:=
\bbE\!\left[
\sum_{t=1}^T
\left(
\ell_t(X_t,A_t)
-
\ell_t(X_t,\pi_T^\star(X_t))
+
c\abs{O_t}
\right)
\right],
\end{equation}
where the expectation is over the contexts, the learner's randomization, and any bounded feedback noise. This directly adapts the regret definitions in bandits with paid observations \citep{paidobservations} and linear contextual bandits \citep{BOBWlinear}.

Finally, let $\cH_t:=\sigma(X_s,A_s,O_s,\{Y_s(o):o\in O_s\},s\le t)$ be the history up to round $t$. We write $\bbP_t[\cdot]=\bbP[\cdot\mid\cH_{t-1}]$ and $\bbE_t[\cdot]=\bbE[\cdot\mid\cH_{t-1}]$. We use $a=\cO(b)$ and $a\lesssim b$ interchangeably, with constants independent of the problem parameters $T,d,K,c,\Sigma,\cD,B_\star,\mu_{\min},\Delta_{\min},C_\theta$, and $C_{\mathrm{corr}}$.

	\section{Algorithms} \label{sec::algorithm}

We study two instantiations of the same FTRL scheme with exact importance-weighted (IW) observations. At each round, the learner makes two distinct randomized decisions. The distribution $q_t(\cdot\mid x)$ determines which arm is \emph{played}, while $p_{t,a}(x)$ determines whether arm $a$ is \emph{observed}. Importantly, these two probabilities need not coincide. The FTRL component controls the action distribution $q_t$, while the observation rule $p_t$ is designed to balance the variance of the loss estimates against the cost of acquiring feedback.

Our two algorithms implement this balance in different ways. \emph{Aggregate-SPB} (\agg) aggregates local stability quantities over the context distribution before choosing a single observation rate, thereby recovering a scalar SPB-matching problem. \emph{Cumulative-Entropy SPB} (\loc) instead keeps the observation rule local, using arm/context-dependent probabilities together with a learning rate driven by the realized Tsallis entropy. Complete pseudocode appears as Algorithms~\ref{alg:clipped-aggregate-spb-iw} and~\ref{alg:kuroki-style} in Appendix~\ref{app::algos}.

\subsection{Shared FTRL and exact-IW template}

Both algorithms share the same IW estimator, FTRL policy, and local stability quantities. They differ only in how these quantities are used to choose the observation probabilities and update the inverse learning rate.

\paragraph{Exact importance weighting.}
At each round, both algorithms choose predictable observation probabilities $p_{t,a}(x)\in[0,1]$ and independently include each arm $a$ in $O_t$ with probability $p_{t,a}(X_t)$. The corresponding observation covariance is
\[
\Sigma_{t,a}
:=
\bbE_t\!\left[p_{t,a}(X_t)X_tX_t\transpose\right].
\]
This matrix plays the role of the scalar observation probability in standard importance weighting: it describes which directions of the context space are observed when querying arm $a$. When $\Sigma_{t,a}$ is invertible, the exact-IW estimate of the linear loss parameter is
\begin{equation*}
	\widehat{\theta}_{t,a}^{\mathtt{IW}}
	:=
	\ind\{a\in O_t\}\Sigma_{t,a}^{-1}X_tY_t(a),
	\quad \text{and we set} \quad
	\widehat{\ell}_t^{\mathtt{IW}}(x,a)
	:=
	\inp{x,\widehat{\theta}_{t,a}^{\mathtt{IW}}}.
\end{equation*}
Under the no-selection-bias condition of Section~\ref{sec::setting},
$\bbE_t[\widehat{\theta}_{t,a}^{\mathtt{IW}}]=\theta_{t,a}$, so the estimator is unbiased. We write
$\widehat{\ell}_t^{\mathtt{IW}}(x):=(\widehat{\ell}_t^{\mathtt{IW}}(x,a))_{a\in[K]}$
for the vector of estimated losses at context $x$.

\paragraph{FTRL policy.}
Fix $\alpha\in(0,1)$ and let $\bar\alpha:=1-\alpha$. For $q\in\triangle_K$, write
$H_\alpha(q):=\alpha^{-1}\sum_{a=1}^K(q_a^\alpha-q_a)$.
Following \citet{BOBWhardproblems}, both algorithms use the composite Tsallis regularizer
$\psi_t:=-\beta_t H_\alpha$ and $\bar\psi:=-\bar\beta H_{\bar\alpha}$, where $\beta_t$ is the inverse learning rate and $\bar\beta$ is fixed. Given the past exact-IW loss estimates, the policy for context $x$ is
\begin{equation}\label{eq::ftrl_alloc_x}
	q_t(\cdot \mid x)
	\in
	\argmin_{q\in\simplex_K}
	\scbr*{
		\sum_{s=1}^{t-1}\inp*{q,\widehat{\ell}_s^{\mathtt{IW}}(x)}
		+ \psi_t(q)
		+ \bar\psi(q)
	} .
\end{equation}
After observing $X_t$, the learner samples $A_t\sim q_t(\cdot\mid X_t)$.

\paragraph{Local Tsallis stability weights.}
The observation rules are driven by the local geometry of the FTRL policy. For each context $x$, define
$h_t(x):=H_\alpha(q_t(\cdot\mid x))$, let
$I_t(x)\in\argmax_{a\in[K]}q_t(a\mid x)$, and set
\[
s_t(x)
:=
\min\{q_t(I_t(x)\mid x),1-q_t(I_t(x)\mid x)\}.
\]
The corresponding local \emph{stability weights} are
\begin{equation*}
	z_{t,a}(x)
	:=
	\begin{cases}
		q_t(a\mid x)^{2-\alpha}, & a\neq I_t(x),\\[1mm]
		s_t(x)^{2-\alpha}, & a=I_t(x),
	\end{cases}
	\qquad
	Z_t(x):=\sum_{a=1}^K z_{t,a}(x).
\end{equation*}
These quantities arise directly in the local Tsallis stability bound: up to constants, the one-step stability is controlled by a weighted quadratic form involving
$z_{t,a}(x)(\widehat{\ell}_t^{\mathtt{IW}}(x,a))^2$.
The observation mechanisms of both algorithms are designed precisely to control these weighted second moments. The difference is whether this control is performed after aggregating over contexts, as in \agg, or locally at the arm/context level, as in \loc.

\subsection{Aggregate-SPB (\agg)}

A direct adaptation of the hard-problem SPB recipe of \citet{BOBWhardproblems} would try to match stability and observation cost context-wise, as in the multi-armed paid-observation setting. In the contextual model this does not lead to a scalar matching problem: the stability contribution depends both on the local allocation $q_t(\cdot\mid x)$ and on the covariance geometry of the context distribution. In particular, when the observation rate $p_t$ is context-independent, the local stability bound takes the schematic form
\begin{equation*}
	\operatorname{stab}_t(x)
	\lesssim
	\frac{L(x)Z_t(x)}{\beta_t p_t},
	\qquad
	L(x):=x\transpose\Sigma^{-1}x.
\end{equation*}
Thus, the natural scalar quantity only appears after averaging over $X\sim\cD$. The role of \agg\ is precisely to perform this aggregation before applying SPB matching.

\agg\ takes $\alpha\in[1/2,1)$ and uses a context-independent observation rate:
$p_{t,a}(x)\coloneqq p_t$ for all arms and contexts. Hence
$\Sigma_{t,a}=p_t\Sigma$, where
$\Sigma:=\bbE_{X\sim\cD}[XX\transpose]$, and the exact-IW estimator reduces to
\begin{equation*}
	\widehat{\theta}_{t,a}^{\mathtt{IW}}
	=
	\frac{\ind\{a\in O_t\}}{p_t}\Sigma^{-1}X_tY_t(a).
\end{equation*}
Averaging the preceding local stability term motivates the aggregate quantity
\begin{equation*}
	z_t
	:=
	\frac{4cK}{1-\alpha}\,
	\bbE_{X\sim\cD}\sbr*{L(X)Z_t(X)}.
\end{equation*}
It represents the stability/variance scale that enters the scalar SPB problem in the framework of \cite{BOBWhardproblems}.

To control not only second moments but also the pathwise magnitude of the IW estimates, we introduce a second quantity. Let
$\htmax:=\sup_{x\in\cX}h_t(x)$ and
$m_t:=\sup_{x\in\cX}s_t(x)$, and define
\begin{equation*}
	u_t^{\mathrm{mag}}
	:=
	\frac{8cKB_\star}{1-\alpha}\,m_t^{1-\alpha}.
\end{equation*}
The role of $u_t^{\mathrm{mag}}$ is to guarantee the deterministic magnitude condition required by the classical FTRL regret decomposition, more precisely for its stability component. Because the observation probability must also be clipped at one, we further define
\begin{equation*}
	\widetilde u_t
	:=
	u_t^{\mathrm{mag}}+\frac{z_t}{cK},
\end{equation*}
where the additional term accounts for the clipping correction.

Let
$\gamma_t:=cKp_t$ denote the expected observation cost at round $t$. Ignoring for a moment the magnitude safeguard and clipping, the stability-plus-observation-cost contribution has the scalar form
\begin{equation*}
	\frac{z_t}{\beta_t\gamma_t}+\gamma_t,
\end{equation*}
which is balanced for
$\gamma_t\asymp\sqrt{z_t/\beta_t}$.
This motivates the observation rule
\begin{equation*}
	\gamma_t
	:=
	\min\scbr*{
		\sqrt{\frac{z_t}{\beta_t}}
		+
		\frac{u_t^{\mathrm{mag}}}{\beta_t},
		cK
	}.
\end{equation*}
The corresponding SPB envelope is
\begin{equation*}
	b_t
	:=
	2\sqrt{\frac{z_t}{\beta_t}}
	+
	\frac{\widetilde u_t}{\beta_t},
\end{equation*}
and to complete the SPB matching, the inverse learning rate is updated according to
\begin{equation*}
	\beta_{t+1}
	:=
	\beta_t+\frac{b_t}{2\htmax}.
\end{equation*}
\paragraph{Round-by-round SPB-matching} Thus, \agg\ performs \emph{direct} stability-penalty-bias (observation cost) matching: the three quantities are matched \emph{at every round}, as they are all controlled through the same scalar $b_t$, involving terms computed using the full distribution of FTRL allocations across contexts. While this can be burdensome for large context spaces, this aggregate control yields the sharper stochastic bound in \Cref{thm::main}. More precisely, the price is that the tuning rule depends on the global distributional quantities
$\bbE_{X\sim\cD}[L(X)Z_t(X)]$ and $h_t^{\max}$, which themselves require evaluating the FTRL allocation \eqref{eq::ftrl_alloc_x} over the support of $\cD$. Approximate versions would require a separate perturbation analysis.

\subsection{Cumulative-Entropy SPB (\loc)}

The \loc\ instantiation keeps the same FTRL policy and exact-IW estimator, but avoids collapsing the contextual stability quantities into a single aggregate scalar. Instead, the observation probabilities are chosen locally so as to cancel the stability weights $z_{t,a}(x)$ directly. Here, ‘local’ refers to the arm and context-dependent form of the observation rule and matching argument. However, the exact implementation still requires distributional expectations through $\Sigma_{t,a}^{-1}$. 

Starting from an initial entropy scale $\Lambda_0>0$, define the cumulative realized entropy
\begin{equation}\label{eq::update_le}
	\Lambda_t
	:=
	\Lambda_0+\sum_{s=1}^t h_s(X_s),
	\qquad
	\beta_{t+1}
	:=
	\beta_t
	+
	\zeta\Lambda_t^{-1/2}
	+
	\xi\Lambda_t^{-1/3},
	\qquad
	\eta_t:=\beta_t^{-1},
\end{equation}
for some parameters $\xi, \zeta>0$.
The two increments play complementary roles: $\Lambda_t^{-1/2}$ controls the full-information-like part of the regret, while the $\Lambda_t^{-1/3}$ term is tailored to the $T^{2/3}$ scaling induced by paid observations. Replacing the deterministic time scale by the realized cumulative entropy is key to obtain BoBW guarantees, as the cumulative entropy will admit different scaling in the adversarial and stochastic regimes.

For additional parameters $\nu,u>0$, \loc\ observes arm $a$ at context $x$ with probability
\begin{equation*}
	p_{t,a}(x)
	:=
	\min\scbr*{
		1,
		\underbrace{\frac{z_{t,a}(x)}{\nu\Lambda_{t-1}^{1/3}}}_{\text{local variance control}}
		+
		\underbrace{\frac{u}{\beta_t}}_{\text{magnitude safeguard}}
	}.
\end{equation*}
The two components have distinct roles. The first assigns more observations to arm/context pairs with larger Tsallis stability weights. In the covariance-based second-moment calculation, this choice cancels the local dependence on $z_{t,a}(x)$ and leaves a common paid-observation scale of order $\Lambda_{t-1}^{1/3}$. The second term guarantees $	\Sigma_{t,a}^{-1}
	\preceq
	\frac{\beta_t}{u}\Sigma^{-1}$. Under the initialization conditions of Section~\ref{subsec:analysis-kuroki-style}, this keeps the exact-IW losses within the deterministic range required by the Tsallis stability lemma.

Thus, unlike \agg, \loc\ does not first aggregate the contextual geometry into a scalar observation rate. It controls the variance locally through $p_{t,a}(x)$, while the inverse learning rate adapts to the realized cumulative entropy $\Lambda_t$. This algorithmic choice is the reason why the theoretical guarantees of \loc{} can be expressed as a function of this quantity.

When $c=0$, observations are free and we simply take $p_{t,a}(x)=1$ for every $t\in[T],a\in [K],x\in \cX$. The paid-observation component then disappears, while the $\Lambda_t^{-1/2}$ update retains the full-information-like behavior of the algorithm. More generally, the tuning introduced in Section~\ref{subsec:analysis-kuroki-style} interpolates continuously with the observation cost.

\paragraph{Final-horizon SPB matching.}
Although \loc\ does not explicitly match the SPB terms round by round, it still implements the same balancing principle more indirectly, at a final horizon and context-averaged level. More precisely, the two components of the update~\eqref{eq::update_le} are chosen so that the three terms of the SPB regret decomposition can all be controlled as functions of the cumulative entropy, more precisely $\Lambda_T^{1/2}$ and $\Lambda_T^{2/3}$. 

Hence, \agg\ performs SPB matching \emph{locally in time after globally aggregating over contexts}, whereas \loc\ keeps the observation rule \emph{local in $(x,a)$}, but its learning rate update based on $\Lambda_T$ balances the three components at the level of the full regret bound.

A natural question is why \loc\ uses Tsallis entropy rather than the Shannon entropy employed by FTRL-LC~\citep{BOBWlinear}. In the present setting, the action probabilities $q_t$ and observation probabilities $p_t$ are decoupled. Consequently, the variance of the IW estimates is governed by the observation rule rather than directly by the action distribution. The local Tsallis stability bound provides the compensating weights $z_{t,a}(x)$, which can be matched explicitly by the observation probabilities above. This local stability structure is the key ingredient allowing the observation cost and estimation variance to be controlled on the same entropy-dependent scales.

	\section{Regret Analysis} \label{sec::regret}

We provide BoBW regret upper bounds for the two exact-IW FTRL instantiations introduced in \Cref{sec::algorithm}. \Cref{thm::main} studies \agg, while \Cref{thm:le-spb} studies \loc. Together, the two results allow us to discuss the theoretical tradeoffs between the global aggregate matching and local entropy-adaptive matching.

\subsection{BOBW regret guarantees of Aggregate-SPB}
\label{subsec:analysis-exact-iw-aggregate-spb}

We first analyze \agg, which performs direct SPB matching after aggregating the local stability quantities over the context distribution. The theorem below states the main scaling of the resulting regret bounds. The fully explicit version, including numerical constants, initialization conditions, and all lower-order terms, is stated and proved in \Cref{thm:exact-iw-aggregate-spb}.

\begin{theorem}[Main guarantee for \agg]
	\label{thm::main}
	Consider Algorithm~\ref{alg:clipped-aggregate-spb-iw} under the boundedness and covariance assumptions of Section~\ref{sec::setting}. Assume $c>0$ and initialization chosen as in \Cref{thm:exact-iw-aggregate-spb}, and let
	$\alpha=\max\left\{1-\frac{1}{\log(K)},1/2\right\}$.
	Then, for any $T\geq 1$, the following bounds hold up to universal numerical constants, initialization terms, and the gap- and corruption-dependent lower-order remainder made explicit in \Cref{thm:exact-iw-aggregate-spb}:
	\begin{enumerate}[label=\textnormal{(\roman*)}]
		\item In the adversarial linear regime,
		\[
		R_T
		\lesssim
		(cKd\log K)^{1/3}T^{2/3}
		+
		\sqrt{(cKB_\star+d)\log K\,T}.
		\]
		
		\item Suppose, in addition, that the stochastic and corrupted stochastic conditions of Section~\ref{sec::setting} hold. Let $\Delta:=\min\{\Delta_{\min},1\}$, let $C_{\mathrm{corr}}$ be the corruption budget, and define
		\[
		\rho_{\mathrm{IW}}
		:=
		\log K\cdot
		\max\scbr*{
			\frac{cKd}{\mu_{\min}^2},
			\frac{cKB_\star+d}{\mu_{\min}}
		},
		\qquad
		\tau:=\Delta^{-3}+\frac{C_{\mathrm{corr}}}{\Delta}.
		\]
		Then, for every $T\geq \tau$,
		\[
		R_T
		\lesssim
		\frac{\rho_{\mathrm{IW}}}{\Delta^2}\log_+(T\Delta^3)
		+
		\spr*{
			\frac{C_{\mathrm{corr}}^2\rho_{\mathrm{IW}}}{\Delta^2}
			\log_+\spr*{\frac{T\Delta}{C_{\mathrm{corr}}}}
		}^{1/3}.
		\]
		The corruption-dependent term is interpreted as zero when $C_{\mathrm{corr}}=0$. Thus, in the purely stochastic case, the only dependence on $T$ is logarithmic, up to the $T$-independent gap- and initialization-dependent terms made explicit in \Cref{thm:exact-iw-aggregate-spb}.
	\end{enumerate}
\end{theorem}

The adversarial bound separates two effects. The leading $T^{2/3}$ term is the characteristic contribution of paid observations: it results from balancing the variance of the importance-weighted estimates against the cost of acquiring feedback. The second term is of order $\sqrt{T}$ and arises from the magnitude-control component of the SPB analysis. In particular, it remains of order $\sqrt{T}$ as $c\downarrow 0$, consistently with the full-information limit, although the theorem itself is stated only for $c>0$.

The second part gives the best-of-both-worlds conversion. In a stochastic environment, the probability of selecting suboptimal actions controls both the stability scale and the amount of information that needs to be purchased. This self-bounding mechanism converts the adversarial $T^{2/3}$ guarantee into logarithmic dependence on the horizon when $C_{\mathrm{corr}}=0$, while degrading smoothly with the corruption budget. The finite-support assumption enters this conversion through $\mu_{\min}$: it allows us to control worst-context suboptimal-action probabilities from their average under the context distribution. When $c=0$, querying all arms is free and the problem instead reduces to the full-information regime.

\begin{proof}[Proof sketch]
	The main difficulty of the proof is to verify the necessary conditions to apply the ``BoBW for hard problems'' framework of \cite{BOBWhardproblems}, which inspired the design of \agg. It has three main steps. First, exact importance weighting and the ghost-context argument reduce the regret to an FTRL stability--penalty decomposition with zero estimation bias. Second, the aggregate observation rule turns the context-dependent stability contributions into a single scalar SPB envelope $b_t$ that simultaneously controls stability, observation cost, and penalty. Third, in the stochastic regime, the same SPB quantities can be self-bounded by the probability of playing suboptimal actions, allowing us to invoke the BoBW transfer theorem \ref{thm:spb-transfer}.
	
	For the first step, since \agg\ uses a context-independent observation rate, $\Sigma_{t,a}=p_t\Sigma$ for every arm and $\widehat\theta_{t,a}^{\mathtt{IW}}$ is conditionally unbiased. The ghost-context argument with an independent $X_0\sim\cD$ therefore reduces regret against $\pi_T^\star$ to the stability and penalty terms for the estimated losses, plus the paid-observation cost, with zero bias.
	
	For the second step, fix a context $x$. The local Tsallis stability lemma produces a $Z_t(x)$-weighted quadratic term. Its pathwise magnitude condition follows from $\abs*{x\transpose\Sigma^{-1}X_t}\le B_\star$ together with the lower bound $\gamma_t\ge u_t^{\mathrm{mag}}/\beta_t$. Moreover,
	\[
	\bbE_t\sbr*{\elltiw(x,a)^2}
	\le
	\frac{L(x)}{p_t}.
	\]
	Averaging this bound over the context distribution yields a stability contribution $z_t/(\beta_t\gamma_t)$. After adding the observation cost, the one-round contribution becomes
	\[
	a_t
	:=
	\frac{z_t}{\beta_t\gamma_t}+\gamma_t
	\le
	b_t,
	\qquad
	b_t
	:=
	2\sqrt{\frac{z_t}{\beta_t}}
	+
	\frac{\widetilde u_t}{\beta_t},
	\]
	where the term $z_t/(cK)$ in $\widetilde u_t$ accounts for clipping the observation probability at one.
	
	The same envelope $b_t$ also controls the penalty. The composite Tsallis regularizer guarantees
	$h_{t+1}^{\max}\le 2h_t^{\max}$, and with
	\[
	\beta_{t+1}
	=
	\beta_t+\frac{b_t}{2h_t^{\max}},
	\]
	the penalty is bounded by the initialization cost plus $\sum_t b_t$. Hence
	\[
	R_T
	\le
	\beta_1h_\star+\bar\beta\bar h_\star
	+
	2\bbE\left[\sum_{t=1}^T b_t\right] \;.
	\]
	Applying the SPB summation theorem with deterministic bounds on $z_t$, $\widetilde u_t$, and the available next-entropy bound $\widehat h_{t+1}=2h_t^{\max}$ gives the adversarial rate.
	
	For the stochastic conversion, define
	\[
	e_t(x)
	:=
	1-q_t(a^\star(x)\mid x),
	\qquad
	\bar e_t
	:=
	\bbE_{X\sim\cD}[e_t(X)].
	\]
	The finite-support assumption implies
	$e_t^{\max}\le \bar e_t/\mu_{\min}$.
	Combining this with the local entropy and stability self-bounds yields
	\[
	\sqrt{z_t\widehat h_{t+1}}
	\lesssim
	\sqrt{\rho_{\mathrm{IW}}}\,\bar e_t,
	\qquad
	\widetilde u_t\widehat h_{t+1}
	\lesssim
	\rho_{\mathrm{IW}}\,\bar e_t.
	\]
	Thus both SPB scales are controlled linearly by the average probability of playing a suboptimal arm. Finally, the gap and corruption assumptions give
	\[
	R_T
	\ge
	\Delta\bbE\sum_{t=1}^T\bar e_t
	-
	C_{\mathrm{corr}}.
	\]
	The self-bounding part of \Cref{thm:spb-transfer} then yields the stated logarithmic/corrupted-stochastic bound with
	$\tau=\Delta^{-3}+C_{\mathrm{corr}}/\Delta$.
\end{proof}

\subsection{BOBW regret guarantees of Cumulative-Entropy SPB}
\label{subsec:analysis-kuroki-style}

We next state the guarantees for \loc. All algorithmic quantities,
including the entropy clock $\Lambda_t$, the local weights
$z_{t,a}(x)$, and the observation probabilities $p_{t,a}(x)$,
are defined in Section~\ref{sec::algorithm}. We only specify here
the horizon-dependent tuning used in the main result.

Assume $T\ge 3$ and $2\le K\le T$, and choose
\begin{equation*}
	\alpha
	:=
	1-\frac{1}{\log T}.
\end{equation*}
The complementary regularizer in~\eqref{eq::ftrl_alloc_x} is not
needed for \loc, so we set $\overline\beta=0$. We further define
\begin{equation}
	\begin{aligned}
		u
		&:=
4B_\star \log(T),
		\qquad &\nu
		:=
		\left(
		\frac{4c^2}
		{3Kd}
		\right)^{1/3}, \qquad &\Lambda_0=\max\{1, h_\star\},
		 \\ 
		\xi&
		:=
		\left(
		\frac{16cKd}
		{9}
		\right)^{1/3} \log(T),
		\quad
		&\zeta
		:=
		\sqrt{K(d+cB_\star)} \; \log(T), \quad
		&\beta_1
		:=
		\max\left\{u,
		\,2\zeta,
		\,2\xi
		\right\} .
	\end{aligned}
	\label{eq:le-main-tuning}
\end{equation}
We now detail the BoBW guarantees obtained for \loc{} under this tuning. 

%

\begin{theorem}[Main guarantee for \loc]
	\label{thm:le-spb}
	Assume $c\ge0$, $T\ge 3$, $2\le K\le T$, and that the horizon
	$T$ is known. Consider Algorithm~\ref{alg:kuroki-style} with its parameters tuned according to Equation~\eqref{eq:le-main-tuning}, under the boundedness
	and covariance assumptions of Section~\ref{sec::setting}.
	 Then the following guarantees hold.
	
	\begin{enumerate}[label=\textnormal{(\roman*)}]
		\item
	The regret of \loc{} can be upper bounded as follows,
		\begin{equation}
			R_T
			\lesssim
			\zeta\,
			\bbE\sbr{\Lambda_T^{1/2}}
			+
			\xi\,
			\bbE\sbr{\Lambda_T^{2/3}}
			+
				B_\star \log(K),
			\label{eq:le-main-adaptive}
		\end{equation}
\item In particular, in the adversarial regime it holds that 
		\begin{equation}
			\begin{aligned}
				R_T
				\lesssim\;&
				\sqrt{
					K(d+cB_\star)
					T\log T\log K
				}
				&+
				(cKd\log T)^\frac{1}{3}(T\log(K))^\frac{2}{3}
				+
				B_\star\log K.
			\end{aligned}
			\label{eq:le-main-adversarial}
		\end{equation}
		
		\item
		Suppose, in addition, that the corrupted-stochastic
		conditions of Section~\ref{sec::setting} hold, and let
		\begin{equation*}
			\Delta
			:=
			\min\{\Delta_{\min},1\}.
		\end{equation*}
		Then
		\begin{equation}
			R_T
			\lesssim
			\frac{
				K(d+cB_\star)\log^2 T
			}{
				\Delta
			}
			+
			\frac{
				cKd\log^3 T
			}{
				\Delta^2
			}
			+
			B_\star\log K
			+
			C_{\mathrm{corr}}.
			\label{eq:le-main-stochastic}
		\end{equation}
	\end{enumerate}
\end{theorem}

The first part makes explicit the entropy adaptivity of \loc. Unlike a worst-case analysis indexed
directly by the horizon, the bound~\eqref{eq:le-main-adaptive} depends on the realized cumulative
entropy $\Lambda_T$. Whenever the FTRL policies concentrate rapidly, $\Lambda_T$ can be
substantially smaller than its worst-case bound $\Lambda_0+h_\star T$, which is the key mechanism
behind the BoBW guarantee.

As for \agg, the paid-observation component has the characteristic
$(cKd)^{1/3}T^{2/3}$ scaling, up to logarithmic factors. The
$\Lambda_t^{-1/2}$ component of the inverse-learning-rate update ensures that the bound remains
meaningful as $c\to0$. Under the unified tuning above, the resulting worst-case full-information
term scales as
$\sqrt{KdT\log T\log K}$. However, as shown in \Cref{rem::full_info_loc}, when $c=0$ the stability
bound can be sharpened by a factor $K$, and retuning $\zeta$ accordingly yields a
$\sqrt{dT\log T\log K}$ bound, which admits the expected scaling in $K$ in this regime.

The stochastic guarantee relies crucially on the horizon-dependent choice
$\alpha=1-1/\log T$. Indeed, the entropy self-bound contains a factor $	(K-1)^{1-\alpha} T^{1-\alpha}$,
which is bounded by a universal constant under $K\le T$ for this choice of $\alpha$. This prevents the cumulative entropy from introducing a polynomial dependence on $T$. The price is that the factors
$(1-\alpha)^{-1}$ appearing in the stability and observation-cost bounds become logarithmic in
$T$, and propagate through the tuning of $\zeta$ and $\xi$, producing the additional logarithmic
factors in \eqref{eq:le-main-adversarial}-\eqref{eq:le-main-stochastic}. These logarithmic powers need not
be optimal. However, they cannot be removed within the present mechanism with a fixed, horizon-independent
$\alpha<1$, as we show in \Cref{lem::unavoidable_log_loc}: the current cumulative
entropy update incurs polynomial stochastic regret even on a deterministic two-arm
full-information instance.

\begin{proof}[Proof sketch]
	As for \Cref{thm::main}, exact importance weighting and the
	ghost-context argument of Appendix~A reduce the regret to
	stability, penalty, and paid-observation cost, with zero
	estimation bias. The difference is that these terms are controlled
	through the realized entropy clock $\Lambda_t$ rather than by a
	common scalar SPB envelope.
	
	For the stability term, the local Tsallis stability lemma produces
	the weights $z_{t,a}(x)$. The observation rule guarantees the
	covariance lower bound
	\begin{equation*}
		\Sigma_{t,a}
		\succeq
		\frac{u}{\beta_t}\Sigma,
	\end{equation*}
	which controls the pathwise magnitude of the exact-IW estimates.
	Moreover, writing
	$r_t:=\nu\Lambda_{t-1}^{1/3}$, clipping-aware covariance
	arguments yield Lemma~\ref{lem:hybrid-stability}, which states 
	\begin{equation*}
\stability
		\lesssim
		\frac{Kd}{1-\alpha} \; 
		\bbE \left[
		\sum_{t=1}^T
		\frac{
			1+\nu\Lambda_{t-1}^{1/3}
		}{
			\beta_t
		} \right] .
	\end{equation*}
	The two inverse-learning-rate sums are then controlled separately by Lemma~\ref{lem:hybrid-beta-sums}: the first yields a $\Lambda_T^{1/2}$ contribution, while the
	second yields a $\Lambda_T^{2/3}$ contribution.
	
	The same two entropy scales appear in the remaining terms. The
	penalty satisfies
	\begin{equation*}
		\pen
		\lesssim
		\beta_1h_{\star}
		+
		\zeta\bbE\sbr{\Lambda_T^{1/2}}
		+
		\xi\bbE\sbr{\Lambda_T^{2/3}},
	\end{equation*}
	while the observation-cost analysis yields
	\begin{align*}
		\obs&\le \;
		\frac{2c \alpha}{\nu (1-\alpha)}
		\bbE\left[\sum_{t=1}^T
		\frac{h_t(X_t)}{\Lambda_{t-1}^{1/3}}\right]
		+cKu \;\bbE\left[\sum_{t=1}^T\frac1{\beta_t}\right],
	\end{align*}
	where the two terms already appeared in the penalty and stability bounds, respectively. To achieve this, we had to replace the stability weights by entropies, using the Tsallis inequality $Z_t(x)
		\le \frac{2}{1-\alpha}
		H_\alpha(q_t(\cdot\mid x))$ (Lemma~\ref{lem:ub_CalphaZ}), yielding a $\Lambda_T^{2/3}$ contribution from the local
	observation term, together with a $\Lambda_T^{1/2}$ contribution
	from the magnitude safeguard. 
	
	Combining the three upper bounds yields the
	generic entropy-adaptive bound of Proposition~\ref{prop:hybrid-master}, which takes the form 
	\[R_T \leq  a_1 \bE[\Lambda_T^{2/3}] + a_2  \bE[\Lambda_T^{1/2}] + a_3 \bE[\Lambda_T^{1/3}] + a_4  \;, \]
	for some coefficients $(a_i)_{i\in [4]}$ made explicit in the proposition, which depend on all parameters of the algorithm, as well as $c,d, K, \alpha, \log(T)$. The tuning proposed in \eqref{eq:le-main-tuning} comes from optimizing these terms. Simplifying a bit: $a_1$ gives the tuning of $\xi$ and $\nu$, while $a_2$ gives the tuning of $\zeta$. After this, we use Young's inequality to show that the term scaling with $\Lambda_T^{1/3}$ can be absorbed in the two other polynomial terms in $\Lambda_T$ without degrading their constants, as well as a large part of the terms independent of $T$, which leaves only the $B_\star \log(K)$ term in the end.  
	
	For the stochastic conversion, let $S_T
		:=
		\bbE\sum_{t=1}^T
		\spr*{
			1-q_t(a^\star(X_t)|X_t)
		}$. The entropy self-bound gives
	\begin{equation*}
		\bbE\left[\sum_{t=1}^T h_t(X_t) \right]
		\le
		\frac{(K-1)^{1-\alpha}}{\alpha}
		T^{1-\alpha}S_T^\alpha,
	\end{equation*}
which, for $r\in\{1/2,2/3\}$, we combine with the upper bound
	\begin{equation*}
		\bbE\sbr{\Lambda_T^r}
		\le
		\Lambda_0^r
		+
		\left(
		\left[\bbE\sum_{t=1}^T  h_t(X_t) \right]
		\right)^r.
	\end{equation*}
	On the other hand, the gap and corruption assumptions imply
	\begin{equation*}
		S_T
		\le
		\frac{R_T+C_{\mathrm{corr}}}{\Delta}.
	\end{equation*}
	For $\alpha=1-1/\log T$, the factor
	$T^{1-\alpha}(K-1)^{1-\alpha}/\alpha$ is bounded by a universal
	constant under $K\le T$. Using additionally
	$x^p\le 1+x^q$ for $0<p\le q$ to replace $\alpha$ by $1$ in the remaining terms, the entropy-adaptive bound reduces
	to a fixed-point inequality of the form
	\begin{equation*}
		R_T
		\lesssim
		\zeta
		\left(
		\frac{R_T+C_{\mathrm{corr}}}{\Delta}
		\right)^{1/2}
		+
		\xi
		\left(
		\frac{R_T+C_{\mathrm{corr}}}{\Delta}
		\right)^{2/3}
		+
		\zeta+\xi+B_\star\log K.
	\end{equation*}
	Solving this inequality gives
	\begin{equation*}
		R_T
		\lesssim
		\frac{\zeta^2}{\Delta}
		+
		\frac{\xi^3}{\Delta^2}
		+
		\zeta+\xi
		+
		B_\star\log K
		+
		C_{\mathrm{corr}}.
	\end{equation*}
	Substituting the tuning~\eqref{eq:le-main-tuning} proves
	\eqref{eq:le-main-stochastic}; see
	\Cref{thm:le-spb-stochastic} for the complete argument.
\end{proof}

\subsection{Discussion on lower bounds}
\label{subsec:lower-bound-perspective}

We briefly discuss the natural lower bound benchmarks for our setting. We show that there are several reductions to standard analyses that already identify the main difficulty introduced by paid observations.

\paragraph{Adversarial regime.}
The lower bound of \citet{paidobservations} extends directly to the contextual setting. Consider the context set
$\mathcal X=\{e_1,\ldots,e_d\}$, where $(e_j)_{j\in[d]}$ is the canonical basis of $\mathbb R^d$, and let $\mathcal D$ be uniform over $\mathcal X$. Each context then induces an independent $K$-armed bandit with paid observations encountered on $\Omega(T/d)$ rounds with high probability. Applying the non-contextual lower bound on each coordinate and summing over the $d$ contexts gives the $(cKd)^{1/3}T^{2/3}$ term.  
Combining the same reduction with the standard adversarial full-information lower bound
$\Omega(\sqrt{n\log K})$ on each coordinate further gives the following benchmark.
\[R_T \gtrsim
(cKd)^{1/3}T^{2/3}
+
\sqrt{dT\log K}\;.\]
Thus, for $c>0$ the characteristic paid-observation term $(cKd)^{1/3}T^{2/3}$ appearing in our upper bounds is minimax-optimal in its polynomial dependence on $c,K,d$, and $T$, up to logarithmic factors. \agg{} further matches the right rate in full information.

\paragraph{Stochastic regime.} A similarly simple adaptation of the standard lower bound of \citet{lairobbins85} translates to the acquisition cost in the stochastic setting. Consider again $d$ independent contexts, where each context induces a
stochastic $K$-armed instance. Let one arm have a two-point loss distribution
supported on $\{-1,+1\}$ with mean $-\Delta$, and let the remaining $K-1$
arms have the same support and mean $0$. Thus the first arm is uniquely optimal and all other arms
have gap $\Delta$. For every suboptimal arm $a$, consider the alternative
instance in which only the distribution of arm $a$ is changed to have mean
$-2\Delta$. For $\Delta$ sufficiently small, the KL divergence between the
original and alternative distributions is of order $\Delta^2$. A standard
change-of-measure argument therefore implies that any uniformly efficient
algorithm must query each suboptimal arm at rate
\begin{equation*}
	\bbE[N_a^{\rm obs}(T)]
	=
	\Omega\!\left(\frac{\log T}{\Delta^2}\right).
\end{equation*}using the standard quadratic approximation of KL divergences for the chosen distributions. In standard MAB, the $\Delta^{-2}$ dependence would turn into $\Delta^{-1}$ as the regret would be the gap-weighted number of observations. In the paid-observation setting, however, this simplification does not occur for the observation cost, which yields the following statement,  
\[ \text{There exists a stochastic instance for which} \quad \obs \gtrsim 
\frac{cdK}{\Delta^2}
\log\left(\frac{T}{d}
\right),
\]
where we simplify $\log(T/d)$ by $\log(T)$ in the subsequent discussions since we don't try to match asymptotic constant exactly. 
Finally, in the costless limit the corresponding full-information benchmark is $\frac{d\log(K)}{\Delta}$. It is not clear whether the sum of the two bounds forms a tight stochastic lower bound for all regimes of $c\geq 0$, but this provides a basis for discussions in the next section. 
	\section{Discussion and perspectives}
\label{sec:discussion}

The two constructions developed above achieve the same qualitative
best-of-both-worlds objective, but their guarantees differ substantially
in how they depend on the context distribution, on the horizon, and on the corruption level in the stochastic case.
The lower-bound reductions of
Section~\ref{subsec:lower-bound-perspective} provide useful benchmarks
for interpreting these differences.

\subsection{Comparison of the guarantees}

For clarity, Table~\ref{tab:main-comparison} summarizes the leading terms of the guarantees in
\Cref{thm::main,thm:le-spb}. We omit the initialization and lower-order terms made explicit in the corresponding appendix results. For \loc, we report the worst-case specialization of its entropy-adaptive adversarial bound under $2\le K\le T$. For \agg, we recall that 
\[	\rho_{\mathrm{IW}}
:=
\max\scbr*{
	\frac{cKd}{\mu_{\min}^2},
	\frac{cKB_\star+d}{\mu_{\min}}} \; \log K \;.
\]

\begin{table}[H]
	\centering
	\small
	\renewcommand{\arraystretch}{1.55}
	\begin{tabular}{@{}c|cc@{}}
		\toprule
		& \textbf{\agg} & \textbf{\loc} \\
		\midrule
		
		\textbf{Adversarial}
		&
		$\begin{aligned}
			&(cKd\log K)^{1/3}T^{2/3}\\
			&\quad+\sqrt{\log K(d+cKB_\star)\,T}
		\end{aligned}$
		&
		$\begin{aligned}
			&(cKd\log T)^{1/3}(\log K)^{2/3}T^{2/3}\\
			&\quad+\sqrt{K\log K \log T(d+cB_\star)T}
		\end{aligned}$
		\\[5mm]
		
		\textbf{Stochastic}
		&
		$\displaystyle
		\frac{\rho_{\mathtt{IW}}}{\Delta^2}
		\log_+(T\Delta^3)
		$
		&
		$\displaystyle
		\frac{K(d+cB_\star)}{\Delta}\log^2 T
		+
		\frac{cKd}{\Delta^2}\log^3 T
		$
		\\[5mm]
		
		\makecell{\textbf{Corruption} \\ \textbf{(additional term)}}
		&
		$\displaystyle
		\left(
		\frac{C_{\mathrm{corr}}^2\rho_{\mathtt{IW}}}{\Delta^2}
		\log_+\!\left(\frac{T\Delta}{C_{\mathrm{corr}}}\right)
		\right)^{1/3}
		$
		&
		$\displaystyle C_{\mathrm{corr}}$
		\\
		
		\bottomrule
	\end{tabular}
	\caption{Leading regret scalings of \agg\ and \loc. Additive initialization and lower-order terms are omitted. The corruption row reports the leading explicit corruption-dependent contribution.}
	\label{tab:main-comparison}
\end{table}

\paragraph{Adversarial regime.}
In the adversarial regime, the two algorithms achieve broadly similar
guarantees under non-negligible costs. Both recover the minimax-optimal $(cKd)^{1/3}T^{2/3}$ paid-observation rate
up to logarithmic factors, as discussed in
Section~\ref{subsec:lower-bound-perspective}. The main difference between the
two guarantees therefore lies in logarithmic factors and in the
additional $\sqrt{T}$ contribution.

For \agg, the additional $\sqrt{\log K(d+cKB_\star)T}$ term
also recovers the correct $\sqrt{dT\log K}$ scaling as the observation
cost vanishes. The worst-case bound for \loc\ instead contains a $\sqrt{K(d+cB_\star)T\log T\log K}$. By investigating the proof, we can see that the additional $\sqrt K$ factor at $c=0$ comes from the
stability upper bound in Lemma~\ref{lem:hybrid-stability}, which implies a $\sqrt{K}$ factor in the tuning of $\zeta$. However, in Remark~\ref{rem::full_info_loc} we observe that, in full information, a factor $K$ can be shaved from
the stability bound so that, after retuning the $\Lambda_t^{-1/2}$ component the $\sqrt{K}$ term is removed from the final bound. Obtaining a smoother interpolation for all $c>0$ remains challenging with current analysis, and might require introducing an adaptive $\zeta_t$. We leave this question open, as the improvements only concern costs of order $c\leq \sqrt{\frac{Kd}{T}}$, as otherwise the term $(cKd)^{1/3}T^{2/3}$ would dominate $\sqrt{KdT}$ anyway.

Hence, away from the costless limit, the adversarial guarantees of the
two methods are comparable up to logarithmic
factors. The more significant differences appear in the stochastic
regime.

\paragraph{Stochastic regime.}
The lower-bound discussion of
Section~\ref{subsec:lower-bound-perspective} suggests that the
observation cost should scale, on equal-gap instances, as $\frac{cKd}{\Delta^2}\log T$.
Both algorithms recover this expected $\Delta^{-2}$ dependence,
but they depart from this benchmark in very different ways.

For \agg, the stochastic bound is logarithmic in $T$:
\[
R_T
\lesssim
\frac{\rho_{\mathtt{IW}}}{\Delta^2}
\log_+(T\Delta^3),
\qquad
\rho_{\mathtt{IW}}
=
\log K\,
\max\left\{
\frac{cKd}{\mu_{\min}^2},
\frac{cKB_\star+d}{\mu_{\min}}
\right\}.
\]
Thus direct SPB matching shows that the optimal logarithmic horizon
dependence is compatible with the adversarial $T^{2/3}$ regime.
However, this comes with a severe dependence on the context
distribution. For example, for the canonical construction
$\mathcal X=\{e_1,\ldots,e_d\}$ with uniform contexts, one has
$\mu_{\min}=1/d$ and $B_\star=d$, so the first term in
$\rho_{\mathtt{IW}}$ scales as
\[ \rho_{\mathtt{IW}} \gtrsim 
cKd^3\log K,
\]
which is the most favorable case for the value of both $\mu_{\min}$ and $B_\star$. Thus the current
self-bounding analysis of \agg\ leaves a factor of order $d^2$ even
for this well-conditioned context distribution, and can deteriorate
arbitrarily for large or highly imbalanced supports.

The behavior of \loc\ is almost complementary. It pays an extra $\log(T)^2$ term compared with the lower-bound benchmark, but does not involve $\mu_{\min}$. In contrast to \agg, its bound also recovers the $\Delta^{-1}$ rate in full-information, although it pays a sub-optimal $K\log^2(T)$ factor.

This contrast between the two algorithms is interesting: the stochastic limitation of
\agg\ comes from converting average suboptimal-action probabilities
into worst-context quantities through $\mu_{\min}$, whereas the
limitation of \loc\ comes from the entropy clock and its stochastic
self-bounding conversion. In particular, Lemma~\ref{lem::unavoidable_log_loc} shows that keeping
the Tsallis parameter $\alpha<1$ fixed cannot remove this issue for the
current \loc\ update: it already leads to polynomial stochastic regret
on a deterministic two-arm full-information instance. The logarithmic
overhead is therefore tied to the present entropy-based mechanism,
although the powers of $\log T$ in Theorem~\ref{thm:le-spb} need not
be optimal.

\paragraph{Corruption.}
The corrupted-stochastic guarantees provide a further distinction.
For \agg, the leading explicit corruption contribution is sublinear, but gap and horizon-dependent, while for \loc, corruption simply enters
additively as $C_{\mathrm{corr}}$. Therefore, the improved dependence of \agg{} in the corruption budget pays off at least for
\[C_{\mathrm{corr}} \gtrsim \frac{\rho_{\mathtt{IW}}}{\Delta^2} , \]
which depends on both the stochastic instance and the context distribution.

\paragraph{Heterogeneous observation costs.} Although we assume a common observation cost $c$ for simplicity, both methods extend naturally to arm-dependent costs $(c_a)_{a\in[K]}$.
For \agg, keeping a common observation probability across arms simply
replaces each occurrence of $c$ by the average cost
$\bar c:=K^{-1}\sum_a c_a$. The local structure of \loc\ can additionally exploit
the heterogeneity of the costs through arm-dependent observation scales $(\nu_a)_{a \in [K]}$, replacing the $c^{2/3}$ by $c_a/\bar c^{1/3}$.
We omit these direct extensions to keep the presentation focused.

\subsection{What do the two constructions tell us?}

The main difference between the algorithms is therefore not their
adversarial behavior, which is similar up to secondary factors, but the
mechanism by which they obtain BoBW guarantees through their corrupted-stochastic regret upper bound.

The aggregate construction demonstrates that an
$O(\log T)$ stochastic scaling is attainable in contextual
paid-observation problems while retaining the characteristic
$T^{2/3}$ adversarial guarantee. In this sense, \agg\ provides an
important benchmark: the additional logarithmic factors appearing in
\loc\ are not imposed by the problem itself. On the other hand, the
price paid by the current aggregate self-bounding argument is substantial.
Its dependence on $\mu_{\min}$ can dominate the stochastic rate even for
simple context distributions, and the use of worst-context quantities
makes the analysis difficult to extend naturally beyond finite support.

The local/path-dependent construction of \loc{} avoids this difficulty. Its stochastic bound
depends on the context distribution through covariance geometry rather
than through a minimum context mass, and its observation rule retains
the local arm--context structure of the stability bound. Moreover, the
entropy-adaptive guarantee is a natural pathwise measure, which is lost by the aggregate construction. These
features make \loc\ a natural starting point for further refinements,
with the main theoretical objective being to reduce its logarithmic
stochastic overhead while preserving its local structure.

From an implementation perspective, the comparison is more subtle. With exact distributional quantities, \loc\ is not necessarily cheaper:
it requires forming and inverting $|O_t|$ weighted covariance matrices at
each round, whereas \agg\ only uses the fixed inverse $\Sigma^{-1}$.
The local structure of \loc\ may nevertheless be more amenable to
sampling-based approximations, such as matrix-geometric resampling (MGR) \citep{NeuO20, BOBWlinear},
while approximating the aggregate quantities entering the direct SPB
matching rule requires controlling their errors inside the matching
mechanism itself; see Appendix~\ref{app::implem}. 

We believe that MGR estimates can be incorporated into our analysis using techniques similar to those of \cite{BOBWlinear}, with the additional approximation terms absorbed through an appropriate retuning of the algorithmic parameters. We leave this extension for future work, as our goal here is to establish a precise benchmark for the regret rates achievable in linear contextual bandits with paid observations under exact distributional information.

Overall, we view the two approaches as identifying complementary pieces
of a sharper solution: \agg\ shows that logarithmic stochastic
dependence on $T$ is possible, while \loc\ provides a context-local
mechanism with substantially more favorable dependence on the context
distribution. A natural target is therefore an algorithm combining
these two properties.

\subsection{Perspectives}\label{subsec:perspectives}

\paragraph{Beyond finite-support contexts and uniform gaps.}
The finite-support assumption is particularly restrictive for \agg,
whose stochastic analysis depends explicitly on $\mu_{\min}$. Replacing
this worst-context control by a covariance or distribution-dependent
quantity would both sharpen the finite-support result and open the way
to continuous context spaces.

For continuous context distributions, the uniform positive-gap
assumption also becomes restrictive: the minimum gap may vanish even
when nearly ambiguous contexts have very small probability. Extending
the results to margin conditions or other distribution-dependent gap
assumptions, as in recent work on linear contextual bandits \citep{KatoI25}, is therefore
a natural next step.


\paragraph{Closing the stochastic gap.}
The comparison above suggests a more specific theoretical objective:
retain the local context dependence of \loc\ while recovering the
$O(\log T)$ stochastic scaling exhibited by \agg. This would require a
self-bounding mechanism that avoids both the worst-context conversion
through $\mu_{\min}$ and the entropy-clock overhead responsible for the
extra logarithmic factors of \loc. Determining whether such a local SPB
matching rule exists appears to be the main open question raised by our
results.
		

	\bibliography{biblio}
	\bibliographystyle{plainnat}
	
	\newpage
	\appendix
	\numberwithin{equation}{section}
	\providecommand*{\theHtheorem}{}
	\providecommand*{\theHproposition}{}
	\providecommand*{\theHlemma}{}
	\providecommand*{\theHcorollary}{}
	\providecommand*{\theHdefinition}{}
	\providecommand*{\theHassumption}{}
	\providecommand*{\theHremark}{}
	\renewcommand*{\theHequation}{\thesection.\arabic{equation}}
	\renewcommand*{\theHtheorem}{\thesection.\arabic{theorem}}
	\renewcommand*{\theHproposition}{\thesection.\arabic{theorem}}
	\renewcommand*{\theHlemma}{\thesection.\arabic{theorem}}
	\renewcommand*{\theHcorollary}{\thesection.\arabic{theorem}}
	\renewcommand*{\theHdefinition}{\thesection.\arabic{theorem}}
	\renewcommand*{\theHassumption}{\thesection.\arabic{theorem}}
	\renewcommand*{\theHremark}{\thesection.\arabic{theorem}}

	\addtocontents{toc}{\protect\setcounter{tocdepth}{2}}
	{
	\setlength{\cftbeforesecskip}{.8em}
	\setlength{\cftbeforesubsecskip}{.5em}
	\tableofcontents
	}

	\clearpage
	\section{Algorithm template and regret decomposition}
\label{sec:template-decomposition}

Both algorithms considered in this paper are instances of Follow-the-Regularized-Leader (FTRL) with a composite Tsallis entropy regularizer and the exact importance-weighted estimator introduced in Section~3.1. This appendix collects the shared notation and derives the regret decomposition used in the subsequent analyses. The decomposition is based on the ghost-sample technique introduced by \citet{NeuO20}.

\subsection{Shared FTRL and exact-IW template}
\label{sec:notation-algorithm}

We recall the notation introduced in the main paper. For any scalar $\alpha \in (0,1)$, define the $\alpha$-Tsallis entropy over the nonnegative orthant $\bbRnnK$ as
\begin{equation*}
	H_\alpha \spr*{q}
	:=
	\frac{1}{\alpha}
	\sumaK \spr*{q_a^\alpha - q_a}.
\end{equation*}
It is concave on $\bbRnnK$, so the convex regularizer used by FTRL is $-H_\alpha$. We write $\bar\alpha := 1-\alpha$. 

At time $t$, for any context $x$ and arm $a$, the conditional mean loss is $\ell_t\spr*{x,a}=\inp*{x,\theta_{t,a}}$, while the observation is $Y_t(a)$, which may equal the exact mean loss $\ell_t(X_t,a)$ or be a bounded noisy observation with conditional mean $\ell_t(X_t,a)$. The bound $\abs*{Y_t(a)}\le1$ holds almost surely.

We denote by $p_{t,a}\spr*{x}$ the probability of observing arm $a$ at time $t$ conditional on context $x$,
\begin{equation*}
	p_{t,a}\spr*{x}
	:=
	\bbP_t\sbr*{a\in O_t\given X_t=x},
\end{equation*}
and define the corresponding observation covariance matrix
\begin{equation*}
	\Sigma_{t,a}
	:=
	\bbE_t\sbr*{p_{t,a}\spr*{X_t}X_tX_t\transpose}.
\end{equation*}

$\Sigma_{t,a}$ is invertible, because in both algorithms $p_{t,a}(x)>0$ on the support of $\mathcal D$, hence $\Sigma_{t,a}\succ0$ since $\Sigma\succ0$. Thus, we can use the exact importance-weighted estimator 
\begin{equation}
	\thetataiw
	:=
	\ind\scbr*{a\in O_t}\Sigma_{t,a}^{-1}X_tY_t(a),
	\qquad
	\elltiw\spr*{x,a}
	:=
	\inp*{x,\thetaiw_{t,a}}.
	\label{eq:exact-iw-estimator}
\end{equation}
We write $\wh{\ell}^{\mathrm{IW}}_t\spr*{x}\in\bbR^K$ for the vector with components $\elltiw\spr*{x,a}$.

Under the no-selection-bias condition of Section~\ref{sec::setting}, the estimator is conditionally unbiased. Indeed,
\begin{align*}
	\bbE_t\sbr*{\thetaiw_{t,a}}
	&=
	\Sigma_{t,a}^{-1}
	\bbE_t\sbr*{X_tY_t(a)\ind\scbr*{a\in O_t}}\\
	&=
	\Sigma_{t,a}^{-1}
	\bbE_t\sbr*{
		X_t\,
		\bbE_t\sbr*{Y_t(a)\ind\scbr*{a\in O_t}\given X_t}
	}\\
	&=
	\Sigma_{t,a}^{-1}
	\bbE_t\sbr*{
		p_{t,a}\spr*{X_t}X_t\ell_t\spr*{X_t,a}
	}\\
	&=
	\Sigma_{t,a}^{-1}
	\bbE_t\sbr*{
		p_{t,a}\spr*{X_t}X_tX_t\transpose
	}\theta_{t,a}
	=
	\theta_{t,a}.
\end{align*}
Thus, throughout the exact-IW analysis, there is no estimation-bias contribution in the regret decomposition. Note that biased estimates (such as MGR approximations) would introduce an additional term in the following regret bound, following the analysis of \cite{NeuO20}.

For any context $x$ and round $t$, define the FTRL policy
\begin{equation}
	q_t\spr*{\cdot\given x}
	:=
	\argmin_{q\in\simplex_K}
	\scbr*{
		\sum_{s=1}^{t-1}\inp*{q,\wh{\ell}^{\mathrm{IW}}_s\spr*{x}}
		+
		\psi_t\spr*{q}
		+
		\bar\psi\spr*{q}
	},
	\label{eq:shared-ftrl-policy}
\end{equation}
where
\begin{equation*}
	\psi_t:=-\beta_tH_\alpha,
	\qquad
	\bar\psi:=-\bar\beta H_{\bar\alpha},
	\qquad
	\beta_t:=\eta_t^{-1},
\end{equation*}
and $\bar\beta\geq 0$ is fixed.

We also recall the local quantities entering the stability bounds. For every context $x$, define
\begin{equation*}
	h_t\spr*{x}
	:=
	H_\alpha\spr*{q_t\spr*{\cdot\given x}},
	\qquad
	I_t\spr*{x}
	\in
	\argmax_{a\in\sbr*{K}}q_t\spr*{a\given x},
\end{equation*}
and
\begin{equation*}
	s_t\spr*{x}
	:=
	\min\scbr*{
		q_t\spr*{I_t\spr*{x}\given x},
		1-q_t\spr*{I_t\spr*{x}\given x}
	}.
\end{equation*}
The corresponding local Tsallis stability weights are
\begin{equation*}
	z_{t,a}\spr*{x}
	:=
	\begin{cases}
		q_t\spr*{a\given x}^{2-\alpha}, & a\neq I_t\spr*{x},\\
		s_t\spr*{x}^{2-\alpha}, & a=I_t\spr*{x},
	\end{cases}
	\qquad
	Z_t\spr*{x}
	:=
	\sumaK z_{t,a}\spr*{x}.
\end{equation*}
Define also 
\[ h_\star \coloneqq \sup_{q \in \triangle_K} H_\alpha(q) =  \frac{K^{1-\alpha}-1}{\alpha} \;,\]
where we dropped $\alpha$ from the notation for convenience as it will always be clear from context.

Finally, recall
\begin{equation*}
	\Sigma
	:=
	\bbE_{X\sim\cD}\sbr*{XX\transpose},
	\qquad
	B_\star
	:=
	\sup_{x,y\in\cX}\abs*{x\transpose\Sigma^{-1}y}.
\end{equation*}
Since $\cX$ is bounded and $\Sigma\succ0$, we have $B_\star<\infty$. For example, if $\norm*{x}_2\le\XMAX$ for all $x\in\cX$, then
\begin{equation*}
	B_\star
	\le
	\frac{\XMAX^2}{\lambda_{\min}\spr*{\Sigma}},
\end{equation*}
while we also have 
\[B_\star \geq \bE[X^\top \Sigma^{-1} X] =d \;.\]
Hence, if the context distribution is sufficiently uniform so that $\lambda_{\min}(\Sigma)$ is of order $1/d$, $B_\star = \Theta(d)$.

\agg{} and \loc{} specialize this shared template by choosing different observation probabilities $p_{t,a}$ and inverse-learning-rate updates.

\subsection{Regret decomposition}
\label{sec:regret-decomposition}

We recall the definition of the regret, identifying the arm-valued comparator $\pi_T^\star\spr*{x}$ with the corresponding deterministic distribution for simplicity: 
$\pi_T^\star\spr*{a\given x}
	:=
	\ind\scbr*{a=\pi_T^\star\spr*{x}}$. Then,
\begin{align*}
	R_T
	&=
	\bbE\sbr*{
		\sumtT
		\spr*{
			\ell_t\spr*{X_t,A_t}
			-
			\ell_t\spr*{X_t,\pi_T^\star\spr*{X_t}}
		}
	}
	+
	c\,\bbE\sbr*{\sumtT\abs*{O_t}}\\
	&=
	\bbE\sbr*{
		\sumtT
		\sumaK
		\spr*{
			q_t\spr*{a\given X_t}
			-
			\pi_T^\star\spr*{a\given X_t}
		}
		\inp*{X_t,\theta_{t,a}}
	}
	+
	c\,\bbE\sbr*{
		\sumtT
		\sumaK p_{t,a}\spr*{X_t}
	}.
\end{align*}
Here $\ell_t\spr*{x,a}=\inp*{x,\theta_{t,a}}$ denotes the conditional mean loss in all regimes. If the queried feedback is noisy, the observed scalar is $Y_t(a)$, while regret is still measured in terms of $\ell_t$. Moreover,
\begin{equation*}
	\bbE_t\sbr*{\abs*{O_t}\given X_t}
	=
	\sumaK p_{t,a}\spr*{X_t}.
\end{equation*}

Let $X_0\sim\cD$ be independent of $\cH_T$ and of all learner randomization. Conditional on $\cH_{t-1}$, both $X_t$ and $X_0$ are distributed according to $\cD$, and $p_{t,a}$ is $\cH_{t-1}$-measurable. Hence
\begin{equation*}
	\bbE_t\sbr*{p_{t,a}\spr*{X_t}}
	=
	\bbE_t\sbr*{p_{t,a}\spr*{X_0}}
\end{equation*}
for every arm $a$. Since the exact-IW estimator is conditionally unbiased, Lemma~\ref{lem:ghost-sample-neu} gives
\begin{equation*}
	R_T
	=
	\bbE\sbr*{
		\sumtT
		\sumaK
		\spr*{
			q_t\spr*{a\given X_0}
			-
			\pi_T^\star\spr*{a\given X_0}
		}
		\inp*{X_0,\thetaiw_{t,a}}
	}
	+
	c\,\bbE\sbr*{
		\sumtT
		\sumaK p_{t,a}\spr*{X_0}
	}.
\end{equation*}
By the definition of $q_t$, we may apply the standard FTRL regret decomposition from Lemma~\ref{lem::ftrl_regret_decomp}, which yields
\begin{align}
	R_T
	&\le
	\underbrace{
		\bbE\sbr*{
			\sumtT
			\inp*{
				q_t\spr*{\cdot\given X_0}
				-
				q_{t+1}\spr*{\cdot\given X_0},
				\wh{\ell}^{\mathrm{IW}}_t\spr*{X_0}
			}
			-
			D_{\psi_t+\bar\psi}
			\spr*{
				q_{t+1}\spr*{\cdot\given X_0},
				q_t\spr*{\cdot\given X_0}
			}
		}
	}_{=\stability}
	\label{eq:ftrl-decomposition}\\
	&\quad+
	\underbrace{
		\bbE\sbr*{
			\spr*{\psi_{T+1}+\bar\psi}
			\spr*{\pi_T^\star\spr*{\cdot\given X_0}}
			-
			\spr*{\psi_1+\bar\psi}
			\spr*{q_1\spr*{\cdot\given X_0}}
			+
			\sumtT
			\spr*{\psi_t-\psi_{t+1}}
			\spr*{q_{t+1}\spr*{\cdot\given X_0}}
		}
	}_{=\pen}
	\nonumber\\
	&\quad+
	\underbrace{c\,\bbE\sbr*{
		\sumtT
		\sumaK p_{t,a}\spr*{X_0} }
	}_{\texttt{obs}_t},
	\nonumber
\end{align}
where we defined the stability and penalty terms. In the remainder of the paper, the two algorithms differ only in how the observation probabilities and learning-rate updates control these three contributions.
	
	\section{Analysis of \agg}
\label{sec:analysis-aggspb-iw}
In this section, we derive the BOBW guarantees of \agg{} and justify its parameter tuning.

\subsection{Parameters and main bound}

We start by recalling the main properties of \agg{} and related quantities, before stating the exact result we are proving in this section.

The main feature of \agg{} is that its observation rule is context-independent: each arm is included in $O_t$ with probability $p_t$, so for every arm and context we have
\begin{equation*}
	\thetataiw
	=
	\frac{\ind \scbr*{a \in O_t}}{p_t}
	\Sigma^{-1} X_t Y_t(a).
\end{equation*}
As shown in Appendix~\ref{sec:notation-algorithm}, $\thetataiw$ is conditionally unbiased.

We now introduce the quantities specific to the aggregate-SPB construction, in addition to the notation already introduced in \ref{sec:notation-algorithm}. For every context $x$ and time $t\geq 1$, let
\begin{equation*}
	\htmax
	:=
	\sup_{x \in \cX} h_t \spr*{x},
	\qquad
	m_t
	:=
	\sup_{x \in \cX} s_t \spr*{x},
\end{equation*}
and define
\begin{equation*}
	L \spr*{x}
	:=
	x\transpose\Sigma^{-1}x.
\end{equation*}
The aggregate stability quantity is
\begin{equation}
	z_t
	:=
	\frac{4cK}{1-\alpha}
	\bbE_{X\sim\cD}
	\sbr*{
		L \spr*{X} Z_t \spr*{X}
	}.
	\label{eq:def-z}
\end{equation}
To control the pathwise magnitude of the exact-IW losses, define
\begin{equation}
	u_t^{\mathrm{mag}}
	:=
	\frac{8cKB_\star}{1-\alpha}
	m_t^{1-\alpha}.
	\label{eq:def-u-mag}
\end{equation}
Because the observation probability is clipped at one, we further set
\begin{equation*}
	\wt{u}_t
	:=
	u_t^{\mathrm{mag}}
	+
	\frac{z_t}{cK}.
\end{equation*}
Let $\gamma_t:=cKp_t$ denote the expected observation cost at round $t$. The observation rate is chosen according to
\begin{equation}
	\gamma_t
	:=
	\min\scbr*{
		\sqrt{\frac{z_t}{\beta_t}}
		+
		\frac{u_t^{\mathrm{mag}}}{\beta_t},
		cK
	},
	\qquad
	p_t
	:=
	\frac{\gamma_t}{cK}.
	\label{eq:def-gamma}
\end{equation}
The corresponding SPB envelope is
\begin{equation}
	b_t
	:=
	2\sqrt{\frac{z_t}{\beta_t}}
	+
	\frac{\wt{u}_t}{\beta_t}
	=
	2\sqrt{\frac{z_t}{\beta_t}}
	+
	\frac{u_t^{\mathrm{mag}}}{\beta_t}
	+
	\frac{z_t}{\beta_t cK},
	\label{eq:def-b}
\end{equation}
and the inverse learning rate is updated as
\begin{equation}
	\beta_{t+1}
	=
	\beta_t
	+
	\frac{b_t}{2\htmax},
	\label{eq:beta-update}
\end{equation}
with the convention that $\beta_{t+1}=\beta_t$ if $\htmax=0$. The role of the quantities above will be made explicit in the stability and penalty bounds below.

Finally, we define the shortcuts 
\begin{equation*}
	h_\star
	:=
	\frac{K^{1-\alpha}-1}{\alpha},
	\qquad
	\bar h_\star
	:=
	\frac{K^\alpha-1}{1-\alpha}.
\end{equation*}

The main result of this appendix is the following best-of-both-worlds guarantee for the exact-IW version of \agg. The statement keeps the leading dependences on $c$, $K$, and $T$ explicit.
\begin{theorem}[Exact-IW aggregate-SPB guarantee]
	\label{thm:exact-iw-aggregate-spb}
	Consider Algorithm~\ref{alg:clipped-aggregate-spb-iw} under the boundedness and covariance assumptions of Section~\ref{sec::setting}, with the exact-IW FTRL template of Appendix~\ref{sec:notation-algorithm} and the aggregate-SPB quantities defined above. Assume $c>0$, $K\ge2$, and $\alpha\in[1/2,1)$. 
	Choose
	\begin{equation*}
		\beta_1
		\ge
		\frac{8B_\star}{1-\alpha}, \quad 		\bar\beta
		\ge
		\frac{\sqrt{2}A_\Delta}
		{1-\spr*{\sqrt{2}}^{-\alpha}},
	\end{equation*}
with
	\begin{equation*}
		A_\Delta
		:=
		\frac{8\alpha K}{1-\alpha}
		\sqrt{\frac{cd}{(1-\alpha)\beta_1}}
		+
		\frac{8\alpha K\spr*{2cB_\star+d}}
		{(1-\alpha)^2\beta_1}.
	\end{equation*}
Set
	\begin{equation*}
		\kappa_{\mathrm{IW}}
		:=
		3\beta_1h_\star
		+
		\bar\beta\bar h_\star
		+
		\frac{2}{\beta_1}\sqrt{\frac{cKd}{1-\alpha}}
		+
		\frac{8cKB_\star+4d}{(1-\alpha)\beta_1}.
	\end{equation*}
	Then the following guarantees hold.
	\begin{enumerate}[label=\textnormal{(\roman*)}]
		\item In the adversarial linear regime, for every horizon $T\ge1$,
		\begin{equation*}
			R_T
			\lesssim R_T^\text{Adv} 
			\coloneqq	\spr*{
					\frac{cKd\,K^{1-\alpha}}
					{\alpha(1-\alpha)}
				}^{1/3}
				T^{2/3}
				+
				\sqrt{
					\frac{
						\spr*{cKB_\star+d}
						K^{1-\alpha}
					}{
						\alpha(1-\alpha)
					}
					T
				}
				+
				\kappa_{\mathrm{IW}}.
		\end{equation*}
		In particular, for
		\begin{equation*}
			\alpha
			=
			\max\scbr*{
				1-\frac{1}{\log K},
				\frac12
			},
		\end{equation*}
		the leading terms scale as
		\begin{equation*}
			R_T \lesssim
				(cKd\log K)^{1/3}T^{2/3}
				+
				\sqrt{
					\spr*{cKB_\star+d}
					\log(K) \,T
				}
				+
				\kappa_{\mathrm{IW}}
			.
		\end{equation*}
		
		\item Suppose, in addition, that the finite-support stochastic or corrupted-stochastic conditions of Section~\ref{sec::setting} hold. Let	$\Delta
			:=
			\min\scbr*{\Delta_{\min},1}$,
		and $C_{\mathrm{corr}}$ be the corruption budget from Section~\ref{sec::setting}. Further define
		\begin{equation*}
			\rho_{\mathrm{IW}}
			:=
			\max\scbr*{
				\frac{
					cKd\,K^{1-\alpha}
				}{
					\alpha(1-\alpha)\mu_{\min}^2
				},
				\frac{
					\spr*{cKB_\star+d}
					K^{1-\alpha}
				}{
					\alpha(1-\alpha)\mu_{\min}
				}
			},
			\qquad
			\tau
			:=
			\Delta^{-3}
			+
			\frac{C_{\mathrm{corr}}}{\Delta}.
		\end{equation*}
		Then, for every $T\ge1$,
		\begin{align*}
			R_T
			&\lesssim
				\frac{\rho_{\mathrm{IW}}}{\Delta^2}
				\log_+\spr*{T\Delta^3}
				+
				\spr*{
					\frac{
						C_{\mathrm{corr}}^2
						\rho_{\mathrm{IW}}
					}{
						\Delta^2
					}
					\log_+\spr*{
						\frac{T\Delta}
						{C_{\mathrm{corr}}}
					}
				}^{1/3}
				+
				\kappa_{\mathrm{IW}} + R_\tau^{\text{Adv}}
			.
		\end{align*}
		The corruption-dependent term is interpreted as zero when $C_{\mathrm{corr}}=0$.
	\end{enumerate}
\end{theorem}
The rest of Appendix~\ref{sec:analysis-aggspb-iw} proves this theorem: Appendix~\ref{sec:bound-stability-exact-iw} bounds the stability term, Appendix~\ref{sec:bound-penalty-exact-iw} bounds the penalty term. Those two results are then used in Appendix~\ref{sec:putting-together-exact-iw} to derive the adversarial regret bound, and Appendix~\ref{subsec::corrupted_agg} builds on this to prove the gap-dependent bound in the corrupted stochastic regime. The proof template follows \cite{BOBWhardproblems}, and we reuse several of their results. For completeness, we recall them in Appendix~\ref{app::technical_tools}.

\subsection{Bounding the stability term}
\label{sec:bound-stability-exact-iw}

In this section we control the stability term appearing in the FTRL decomposition~\eqref{eq:ftrl-decomposition}. 

\begin{lemma}[Stability term]\label{lem::stab_agg} The stability term of the regret decomposition of \agg{} satisfies 
\begin{equation*}
	\stability
	\le
	\bbE \sbr*{\sumtT \frac{z_t}{\beta_t \gamma_t}}.
\end{equation*}

\end{lemma}

\begin{proof}
For a fixed context $x$, define the one-step stability contribution
\begin{equation*}
    \operatorname{stab}_t \spr*{x}
    :=
    \inp*{q_t \spr*{\cdot \given x} - q_{t+1} \spr*{\cdot \given x}, \elltiw \spr*{x}} - D_{\psi_t} \spr*{q_{t+1} \spr*{\cdot \given x}, q_t \spr*{\cdot \given x}},
\end{equation*}
where we drop the $\bar \psi$ term from the divergence for convenience as $D_{\psi+\bar \psi} \geq D_\psi$.
The stability term in~\eqref{eq:ftrl-decomposition} is therefore
\begin{equation*}
    \stability
    =
    \bbE \sbr*{\sumtT \operatorname{stab}_t \spr*{X_0}}.
\end{equation*}
Throughout this proof, when a ghost context appears, we first condition on its value and then average over $X_0 \sim \cD$. Fix a round $t$ and a context $x \in \cX$. We apply Lemma~\ref{lem:stability-tsallis-bound} with
\begin{equation*}
    q = q_t \spr*{\cdot \given x},
    \qquad
    \ell = \frac{\elltiw \spr*{x}}{\beta_t},
    \qquad
    \bar I = I_t \spr*{x}.
\end{equation*}
The lemma requires the deterministic magnitude condition, for every arm $a$,
\begin{equation*}
    \abs*{\frac{\elltiw \spr*{x, a}}{\beta_t}}
    \le
    \frac{1 - \alpha}{4} s_t \spr*{x}^{\alpha-1}.
\end{equation*}
We verify this condition pathwise. By the definition of the exact-IW estimator, for any arm $a \in \sbr*{K}$,
\begin{equation*}
    \abs*{\elltiw \spr*{x, a}}
    =
    \frac{\ind \scbr*{a \in O_t}}{p_t} \abs*{x\transpose \Sigma^{-1} X_t\,Y_t(a)}
    \le
    \frac{B_\star}{p_t},
\end{equation*}
where we used $\abs*{Y_t(a)} \le 1$ and the definition of $B_\star$. The composite Tsallis FTRL objective has minimizers in the relative interior of the simplex, so $s_t \spr*{x} > 0$ for every $x$ and hence $m_t > 0$. 

Since $p_t = \frac{\gamma_t}{\spr*{c K}}$, and Equation~\eqref{eq:def-gamma} gives $\gamma_t \ge \frac{u_t^{\mathrm{mag}}}{\beta_t} \wedge cK \geq \frac{u_t^{\mathrm{mag}}}{\beta_t}$, where the last inequality follows from the initialization, so we obtain
\begin{equation*}
    \abs*{\elltiw \spr*{x, a}}
    \le
    \frac{B_\star c K \beta_t}{u_t^{\mathrm{mag}}}
    =
    \frac{1 - \alpha}{8}\,\beta_t m_t^{\alpha-1},
\end{equation*}
where the equality uses the definition of $u_t^{\mathrm{mag}}$ in Equation~\eqref{eq:def-u-mag}. Because $\alpha - 1 < 0$, the map $u \mapsto u^{\alpha - 1}$ is decreasing on $\bbR_{>0}$. Since $s_t \spr*{x} \le m_t$, this gives $m_t^{\alpha - 1} \le s_t \spr*{x}^{\alpha - 1}$. Therefore,
\begin{equation} \label{eq::stab_cond_agg}
    \abs*{\frac{\elltiw \spr*{x, a}}{\beta_t}}
    \le
    \frac{1 - \alpha}{8} m_t^{\alpha-1}
    \le
    \frac{1 - \alpha}{4} s_t \spr*{x}^{\alpha-1}.
\end{equation}
Thus the condition of Lemma~\ref{lem:stability-tsallis-bound} holds for the scaled loss vector $\elltiw \spr*{x} / \beta_t$.
%
We can therefore apply the lemma and obtain the pathwise bound
\begin{equation*}
    \operatorname{stab}_t \spr*{x}
    \le
    \frac{4}{\spr*{1 - \alpha}\beta_t} \spr*{\sum_{a \ne I_t \spr*{x}} q_t \spr*{a \given x}^{2 - \alpha} \elltiw \spr*{x, a}^2 + s_t \spr*{x}^{2 - \alpha} \elltiw \spr*{x, I_t \spr*{x}}^2}.
\end{equation*}
We next control the conditional second moment of the exact-IW estimated loss. For the fixed context $x \in \cX$ and arm $a \in \sbr*{K}$,
\begin{align*}
    \bbE_t \sbr*{\elltiw \spr*{x, a}^2}
    &=
    \frac{1}{p_t^2} \bbE_t \sbr*{\ind \scbr*{a \in O_t} \spr*{x\transpose \Sigma^{-1} X_t}^2 Y_t(a)^2} \\
    &\le
    \frac{1}{p_t^2} \bbE_t \sbr*{\ind \scbr*{a \in O_t} \spr*{x\transpose \Sigma^{-1} X_t}^2} \\
    &=
    \frac{1}{p_t} \bbE_{X \sim \cD} \sbr*{\spr*{x\transpose \Sigma^{-1} X}^2} \\
    &=
    \frac{1}{p_t} x\transpose \Sigma^{-1} \bbE \sbr*{X X\transpose} \Sigma^{-1} x
    =
    \frac{L \spr*{x}}{p_t}.
\end{align*}
The inequality uses $\abs*{Y_t(a)} \le 1$, and the next equality uses $\bbP_t \sbr*{a \in O_t \given X_t} = p_t$. Since the second-moment bound is uniform over arms, taking conditional expectation in the pathwise stability bound yields, still for fixed $x$,
\begin{equation*}
    \bbE_t \sbr*{\operatorname{stab}_t \spr*{x}}
    \le
    \frac{4}{\spr*{1 - \alpha} \beta_t p_t}
    L \spr*{x} Z_t \spr*{x}.
\end{equation*}
We can now average this fixed-context inequality over the independent ghost context $X_0 \sim \cD$, and obtain
\begin{align*}
    \bbE_t \sbr*{\operatorname{stab}_t \spr*{X_0}}
    &\le
    \frac{4}{\spr*{1 - \alpha} \beta_t p_t} \bbE_{X_0 \sim \cD} \sbr*{L \spr*{X_0} \spr*{\sum_{a \ne I_t \spr*{X_0}} q_t \spr*{a \given X_0}^{2 - \alpha} + s_t \spr*{X_0}^{2 - \alpha}}} \\
    &=
    \frac{4}{\spr*{1 - \alpha} \beta_t p_t} \bbE_{X \sim \cD} \sbr*{L \spr*{X} Z_t \spr*{X}}.
\end{align*}
By the definition of $z_t$ in Equation~\eqref{eq:def-z} and the identity $\gamma_t = c K p_t$,
\begin{align*}
    \bbE_t \sbr*{\operatorname{stab}_t \spr*{X_0}}
    &\le
    \frac{1}{\beta_t c K p_t}
    \frac{4 c K}{1 - \alpha}
    \bbE_{X \sim \cD} \sbr*{L \spr*{X} Z_t \spr*{X}} \\
    &=
    \frac{z_t}{\beta_t \gamma_t}.
\end{align*}
Summing over $t$ and applying the tower rule yields the desired result.
\end{proof}

\subsection{Bounding the penalty term}
\label{sec:bound-penalty-exact-iw}

We now control the penalty term in the FTRL decomposition. The main point is to show that the composite Tsallis regularizer guarantees the pointwise (for any $x$) entropy growth bound
\begin{equation*}
    H_\alpha \spr*{q_{t+1} \spr*{\cdot \given x}}
    \le
    2 H_\alpha \spr*{q_t \spr*{\cdot \given x}}
\end{equation*}
Once this is proved, the learning-rate update $\beta_{t+1} - \beta_t = b_t / \spr*{2 \htmax}$ immediately gives a penalty contribution bounded by $\sumtT b_t$. We prove the following result.
\begin{lemma}[Penalty term of \agg]\label{lem::pen_agg} If $\alpha \in [1/2, 1)$, then
\begin{equation*}
    \pen
    \le
    \beta_1 \frac{K^{1 - \alpha} - 1}{\alpha}
+
\bar\beta \frac{K^\alpha - 1}{1 - \alpha}
+
\bbE \sbr*{\sumtT \spr*{2 \sqrt{\frac{z_t}{\beta_t}} + \frac{\wt{u}_t}{\beta_t}}}.
\end{equation*}
\end{lemma}
Because the regularizers $- H_\alpha$ and $- H_{\bar \alpha}$ have infinite one-sided slope at the boundary of the simplex, the FTRL solution lies in the relative interior of $\simplex_K$ whenever the cumulative estimated losses are finite. Thus $s_t \spr*{x} > 0$ and $\htmax > 0$ in the analysis below. The convention $\beta_{t+1} = \beta_t$ when $\htmax = 0$ is only an implementation safeguard and will not be used in the proof.

\begin{proof}
Fix $t$ and $x \in \cX$. We will apply Lemma~\ref{lem::ftrl_smooth_bound} with
\begin{equation*}
    L
    =
    \sum_{s=1}^{t-1}\wh{\ell}_s^{\mathtt{IW}} \spr*{x},
    \quad
    \ell
    =
    \elltiw \spr*{x},
    \quad
    q
    =
    q_t \spr*{\cdot \given x},
    \quad
    r
    =
    q_{t+1} \spr*{\cdot \given x},
    \quad
    \beta
    =
    \beta_t,
    \quad
    \beta'
    =
    \beta_{t+1}.
\end{equation*}
To do that, we need to verify the two assumptions of the lemma, namely Equations~\eqref{eq:penalty-det-loss-condition} and \eqref{eq:penalty-det-beta-condition}. Since both assumptions are stated with a maximum, it is enough to satisfy one branch of each maximum. We first verify the loss-magnitude condition, by reusing Eq.~\eqref{eq::stab_cond_agg} from the proof of the stability bound,
%
%
\begin{equation*}
    \abs*{\elltiw \spr*{x, a}}
    \le
    \frac{1 - \alpha}{8}\,\beta_t s_t \spr*{x}^{\alpha-1}.
\end{equation*}
We then use that, for $\alpha \in \spr*{0,1}$,
\begin{equation*}
    \frac{1 - \alpha}{8}
    \le
    \frac{1 - \spr*{\sqrt{2}}^{\alpha - 1}}{2}.
\end{equation*}
Indeed, with $y = 1 - \alpha$, this is equivalent to $1 - 2^{- y/2} \ge y/4$, which follows by concavity of $y \mapsto 1 - 2^{- y/2}$ on $\sbr*{0, 1}$ and $1 - 2^{-1/2} \ge 1/4$. Consequently,
\begin{equation*}
    \norm*{\elltiw \spr*{x}}_\infty
    \le
    \frac{1-\spr*{\sqrt{2}}^{\alpha-1}}{2} \beta_t s_t \spr*{x}^{\alpha-1},
\end{equation*}
hence Equation~\eqref{eq:penalty-det-loss-condition} holds. 

We next verify Equation~\eqref{eq:penalty-det-beta-condition}. From its update rule, we have $\beta_{t+1} - \beta_t = b_t / \spr*{2\htmax} \ge 0$, and Lemma~\ref{lem:tsallis-entropy-lower-bound} gives, for every $x \in \cX$,
\begin{equation*}
    h_t \spr*{x}
    =
    H_\alpha \spr*{q_t \spr*{\cdot \given x}}
    \ge
    c_\alpha s_t \spr*{x}^\alpha,
    \qquad
    c_\alpha := \frac{1 - \alpha}{4\alpha}.
\end{equation*}
Taking the supremum over $x \in \cX$ together with the definitions of $\htmax$ and $m_t$, yields
\begin{equation*}
    \htmax
    =
    \sup_{x \in \cX} h_t \spr*{x}
    \ge
    c_\alpha \sup_{x \in \cX} s_t \spr*{x}^\alpha
    =
    c_\alpha m_t^\alpha.
\end{equation*}
We now upper bound the three components entering $b_t$, namely $\sqrt{z_t}$, $u_t^{\mathrm{mag}}$, and $v_t = z_t / \spr*{cK}$. For every $a \ne I_t \spr*{x}$, one has $q_t \spr*{a \given x} \le s_t \spr*{x}$. Indeed, if $q_t \spr*{I_t \spr*{x} \given x} \le 1/2$, then $s_t \spr*{x} = q_t \spr*{I_t \spr*{x} \given x}$ and the claim follows by definition of $I_t$. If $q_t \spr*{I_t \spr*{x} \given x} > 1/2$, then $s_t \spr*{x} = 1 - q_t \spr*{I_t \spr*{x} \given x} = \sum_{a \ne I_t \spr*{x}} q_t \spr*{a \given x}$, so each component is smaller than the sum. Combining these observations, we have
\begin{equation*}
    Z_t \spr*{x}
    \le
    K s_t \spr*{x}^{2 - \alpha}
    \le
    K m_t^{2 - \alpha}.
\end{equation*}
Since $\bbE_{X \sim \cD} \sbr*{L \spr*{X}} = \tr \spr*{\Sigma^{-1}\bbE \sbr*{X X\transpose}} = d$, the definition of $z_t$ gives
\begin{equation*}
    z_t
    \le
    \frac{4 c K^2 d}{1 - \alpha} m_t^{2 - \alpha}.
\end{equation*}
Introduce the proof constants
\begin{equation*}
    C_z := 2 K \sqrt{\frac{c d}{1 - \alpha}},
    \qquad
    C_u := \frac{8 c K B_\star}{1 - \alpha},
    \qquad
    C_{\mathrm{clip}} := \frac{4 K d}{1 - \alpha}.
\end{equation*}
Then
\begin{equation*}
    \sqrt{z_t} \le C_z m_t^{1 - \alpha/2},
    \qquad
    u_t^{\mathrm{mag}} \le C_u m_t^{1 - \alpha},
    \qquad
    \frac{z_t}{c K} \le C_{\mathrm{clip}} m_t^{2 - \alpha}.
\end{equation*}
Recall the definition of $b_t$ in Equation~\eqref{eq:def-b},
\begin{equation*}
    b_t
    =
    2 \sqrt{\frac{z_t}{\beta_t}} + \frac{u_t^{\mathrm{mag}}}{\beta_t} + \frac{z_t}{\beta_t cK}.
\end{equation*}
Since $\beta_t \ge \beta_1$, the bounds above imply
\begin{equation*}
    b_t
    \le
    \frac{2 C_z}{\sqrt{\beta_1}} m_t^{1 - \alpha/2} + \frac{C_u}{\beta_1} m_t^{1 - \alpha} + \frac{C_{\mathrm{clip}}}{\beta_1} m_t^{2 - \alpha}.
\end{equation*}
Combining the last display with $\htmax \ge c_\alpha m_t^\alpha$ and the update $\beta_{t+1} - \beta_t = b_t / \spr*{2 \htmax}$ gives
\begin{align*}
    \beta_{t+1} - \beta_t
    &\le
    \frac{C_z}{c_\alpha \sqrt{\beta_1}} m_t^{1 - 3 \alpha / 2}
    +
    \frac{C_u}{2 c_\alpha \beta_1} m_t^{1 - 2 \alpha}
    +
    \frac{C_{\mathrm{clip}}}{2 c_\alpha \beta_1} m_t^{2 - 2 \alpha}.
\end{align*}
Since $\alpha \in [1/2, 1)$, the exponents satisfy $1 - 3\alpha/2 \ge 1 - 2\alpha$ and $2 - 2\alpha \ge 1 - 2\alpha$. As $0 < m_t \le 1$, the map $r \mapsto m_t^r$ is nonincreasing, so both $m_t^{1 - 3 \alpha / 2}$ and $m_t^{2 - 2\alpha}$ are at most $m_t^{1 - 2\alpha}$. Hence
\begin{equation*}
    \beta_{t+1} - \beta_t
    \le
    A_\Delta m_t^{1 - 2 \alpha},
    \qquad
    A_\Delta
    :=
    \frac{C_z}{c_\alpha \sqrt{\beta_1}} + \frac{C_u + C_{\mathrm{clip}}}{2c_\alpha \beta_1}.
\end{equation*}
Because $1 - 2 \alpha \le 0$ and $s_t \spr*{x} \le m_t$, we have $m_t^{1 - 2 \alpha} \le s_t \spr*{x}^{1 - 2 \alpha}$. Therefore
\begin{equation*}
    \beta_{t+1} - \beta_t
    \le
    A_\Delta s_t \spr*{x}^{1 - 2 \alpha}.
\end{equation*}
We choose the complementary regularizer weight so that
\begin{equation*}
    \bar\beta
    \ge
    \frac{\sqrt{2} A_\Delta}{1 - 2^{-\frac{\alpha}{2}}}.
\end{equation*}
The denominator is positive because $\bar\alpha \in \spr*{0,1}$. Since $\bar \alpha - \alpha = 1 - 2 \alpha$, the previous displays imply
\begin{equation*}
    \beta_{t+1} - \beta_t
    \le
    \frac{1 - \spr*{\sqrt{2}}^{\bar \alpha - 1}}{\sqrt{2}} \bar \beta s_t \spr*{x}^{\bar \alpha - \alpha}.
\end{equation*}
This is the second branch in Equation~\eqref{eq:penalty-det-beta-condition}, and therefore the beta-increment condition holds for every context $x$. Equations~\eqref{eq:penalty-det-loss-condition} and \eqref{eq:penalty-det-beta-condition} are therefore satisfied, so Lemma~\ref{lem::ftrl_smooth_bound} gives, for any $x \in \cX$,
\begin{equation*}
    H_\alpha \spr*{q_{t+1} \spr*{\cdot \given x}}
    \le
    2 H_\alpha \spr*{q_t \spr*{\cdot \given x}}.
\end{equation*}
Consequently,
\begin{equation*}
    h_{t+1}^{\max}
    \le
    2 \htmax.
\end{equation*}
We now turn to the penalty term itself. By the FTRL decomposition,
\begin{align*}
    \pen
    &= \bbE \sbr*{\spr*{\psi_{T+1}+\bar\psi} \spr*{\pi_T^\star \spr*{\cdot \given X_0}} - \spr*{\psi_1+\bar\psi} \spr*{q_1 \spr*{\cdot \given X_0}}}\\
    &\quad+ \bbE \sbr*{\sumtT \spr*{\psi_t-\psi_{t+1}} \spr*{q_{t+1} \spr*{\cdot \given X_0}}}.
\end{align*}
The comparator $\pi_T^\star \spr*{\cdot \given X_0}$ is deterministic, hence it is a vertex of the simplex. Since $H_\rho \spr*{e_a} = 0$ for every $\rho \in \spr*{0,1}$ and every arm $a$, the first term is zero. Moreover,
\begin{equation*}
    - \spr*{\psi_1 + \bar\psi} \spr*{q_1 \spr*{\cdot \given X_0}}
    =
    \beta_1 H_\alpha \spr*{q_1 \spr*{\cdot \given X_0}} + \bar\beta H_{\bar\alpha} \spr*{q_1 \spr*{\cdot \given X_0}}.
\end{equation*}
Finally, only $\psi_t$ changes with $t$, and therefore
\begin{equation*}
    \spr*{\psi_t-\psi_{t+1}} \spr*{q_{t+1} \spr*{\cdot \given X_0}}
    =
    \spr*{\beta_{t+1}-\beta_t} H_\alpha \spr*{q_{t+1} \spr*{\cdot \given X_0}}.
\end{equation*}
Thus
\begin{align*}
    \pen
    =
    \bbE \sbr*{
        \beta_1 H_\alpha \spr*{q_1 \spr*{\cdot \given X_0}}
        +
        \bar\beta H_{\bar\alpha} \spr*{q_1 \spr*{\cdot \given X_0}}
        +
        \sumtT
        \spr*{\beta_{t+1} - \beta_t}
        H_\alpha \spr*{q_{t+1} \spr*{\cdot \given X_0}}
    }.
\end{align*}
Using $H_\alpha \spr*{q_{t+1} \spr*{\cdot \given X_0}} \le h_{t+1}^{\max} \le 2\htmax$ and the learning-rate update in Equation~\eqref{eq:beta-update}, we obtain
\begin{align*}
    \spr*{\beta_{t+1} - \beta_t} H_\alpha \spr*{q_{t+1} \spr*{\cdot \given X_0}}
    &\le
    \spr*{\beta_{t+1}-\beta_t} 2 \htmax \\
    &=
    b_t.
\end{align*}
Therefore
\begin{equation*}
    \pen
    \le
    \beta_1 \bbE \sbr*{H_\alpha \spr*{q_1 \spr*{\cdot \given X_0}}}
    +
    \bar\beta \bbE \sbr*{H_{\bar\alpha} \spr*{q_1 \spr*{\cdot \given X_0}}}
    +
    \bbE \sbr*{\sumtT b_t}.
\end{equation*}
At $t=1$, FTRL plays the uniform distribution. Therefore
%
\begin{equation*}
    H_\alpha \spr*{q_1 \spr*{\cdot \given X_0}}
    =
    \frac{K^{1 - \alpha}-1}{\alpha},
    \qquad
    H_{\bar\alpha} \spr*{q_1 \spr*{\cdot \given X_0}}
    =
    \frac{K^\alpha-1}{1 - \alpha}.
\end{equation*}
Consequently, using the definition of $b_t$,
\begin{align*}
    \pen
    &\le
    \beta_1 \frac{K^{1 - \alpha} - 1}{\alpha}
    +
    \bar\beta \frac{K^\alpha - 1}{1 - \alpha}
    +
    \bbE \sbr*{\sumtT b_t}\\
    &=
    \beta_1 \frac{K^{1 - \alpha} - 1}{\alpha}
    +
    \bar\beta \frac{K^\alpha - 1}{1 - \alpha}
    +
    \bbE \sbr*{\sumtT \spr*{2 \sqrt{\frac{z_t}{\beta_t}} + \frac{\wt{u}_t}{\beta_t}}}.
\end{align*}
This concludes the proof of the lemma.
\end{proof}

\subsection{Adversarial bound}
\label{sec:putting-together-exact-iw}

We now combine the stability and penalty bounds and derive the adversarial and corrupted-stochastic guarantees for the exact importance-weighted estimator. We assume throughout this subsection that $c > 0$, as the clipped observation rate is written in terms of $\gamma_t = c K p_t$. 

\begin{proof}{Proof of the adversarial bound in Theorem~\ref{thm:exact-iw-aggregate-spb}}
We start by treating the observation term by combining it with the stability upper bound. By definition, we have 
\begin{align*}
	\obs =   c \bbE \sbr*{\sumtT \abs*{O_t}} \leq cK \bE\left[\sum_{t=1} p_t\right]
	= \bbE \sbr*{\sumtT \gamma_t} \;.
\end{align*}
Then, adding this with the stability bound of Lemma~\ref{lem::stab_agg} and the definitions of $\gamma_t, b_t$, we directly obtain 
\begin{align*}
	\stability + \obs &\leq 	\bbE \sbr*{\sumtT \spr*{\frac{z_t}{\beta_t \gamma_t} + \gamma_t}} \\ 
	& \leq \bbE \sbr*{\sumtT \spr*{\frac{z_t}{\beta_t \min \left\{\sqrt{\frac{z_t}{\beta_t}, cK}\right\} } +\sqrt{\frac{z_t}{\beta_t}}
		+
		\frac{u_t^{\mathrm{mag}}}{\beta_t}}} \\
	& \leq \bbE \sbr*{\sumtT \spr*{	2\sqrt{\frac{z_t}{\beta_t}}
			+
			\frac{u_t^{\mathrm{mag}}}{\beta_t}		+
			\frac{z_t}{\beta_t cK}} } = \bbE \sbr*{\sumtT b_t}\;.
\end{align*}

We first record deterministic bounds on the SPB quantities. Since
$2-\alpha>1$, one has $Z_t(x)\le1$ for every $x\in\cX$.
Moreover,
\begin{equation*}
	\bbE_{X\sim\cD}[L(X)]
	=
	\tr\!\spr*{
		\Sigma^{-1}\bbE[XX\transpose]
	}
	=
	d.
\end{equation*}
Therefore,
\begin{equation}
	z_t
	\le
	z_{\max}
	:=
	\frac{4cKd}{1-\alpha},
	\qquad
	\wt u_t
	\le
	\wt u_{\max}
	:=
	\frac{8cKB_\star+4d}{1-\alpha}.
	\label{eq:agg-deterministic-bounds}
\end{equation}

We invoke \Cref{thm:spb-transfer} directly, with the following variable map:
\begin{equation}
    \label{eq:b3-spb-variable-map}
    u_t \leftarrow \widetilde u_t,
    \qquad
    u_{\max} \leftarrow \widetilde u_{\max},
    \qquad
    \widehat h_{t + 1} \leftarrow 2h_t^{\max},
    \qquad
    \widehat h_{\max}^{\mathrm{IW}} := 2h_\star,
\end{equation}
where we recall that $h_\star$ is the maximum entropy for the fixed $\alpha$.
The quantities $z_t$ and $\widetilde u_t$ are nonnegative and adapted, and the deterministic bounds give the required $z_{\max}$ and $u_{\max}$. Since $q_t$ is determined by the history before round $t$, $h_t^{\max}$ is predictable. The composite Tsallis FTRL solution is interior, so $h_t^{\max} > 0$ for $K \ge 2$, and thus $\widehat h_{t+1}$ is positive. Appendix~\ref{sec:bound-penalty-exact-iw} proves $h_{t + 1}^{\max} \le 2h_t^{\max}$, so $\widehat h_{t+1}$ is an available upper bound on the next entropy component. Finally, the update \eqref{eq:beta-update} is exactly the Rule~2 SPB update under the map \eqref{eq:b3-spb-variable-map}:
\begin{equation*}
    \beta_{t+1}
    =
    \beta_t
    +
    \frac{1}{\widehat h_{t+1}}
    \spr*{
        2\sqrt{\frac{z_t}{\beta_t}}
        +
        \frac{\widetilde u_t}{\beta_t}
    }.
\end{equation*}

Combining Lemma~\ref{lem::stab_agg} and Appendix~\ref{lem::pen_agg} gives
\begin{align}
    R_T
    &\le
    \stability
    +
    \pen
    + \obs \nonumber \\
    &\le
    \beta_1h_\star
    +
    \bar\beta\bar h_\star
    +
    2\bbE \sbr*{\sum_{t=1}^T b_t}, \label{eq::RT_agg_bt}
\end{align}
where, by definition,
\begin{equation*}
    b_t
    =
    2\sqrt{\frac{z_t}{\beta_t}}
    +
    \frac{\widetilde u_t}{\beta_t}.
\end{equation*}
Thus the regret upper-bound hypothesis of \Cref{thm:spb-transfer} holds with
\begin{equation}
    \label{eq:b3-R0-def}
    R_0
    :=
    \beta_1h_\star
    +
    \bar\beta\bar h_\star.
\end{equation}

Applying the adversarial-regime part of \Cref{thm:spb-transfer}, and using $\widehat h_{\max}^{\mathrm{IW}} = 2h_\star$, yields
\begin{equation}
    \label{eq:b3-adversarial-bound}
    R_T
    \lesssim 
        (z_{\max}h_\star)^{1/3}T^{2/3}
        +
        \sqrt{\widetilde u_{\max}h_\star T}
        +
        \kappa_{\mathrm{IW}}
,
\end{equation}
where
\begin{equation*}
    \kappa_{\mathrm{IW}}
    :=
    R_0
    +
    \frac{\sqrt{z_{\max}}}{\beta_1}
    +
    \frac{\widetilde u_{\max}}{\beta_1}
    +
    \beta_1\widehat h_{\max}^{\mathrm{IW}}.
\end{equation*}
Substituting $R_0=\beta_1h_\star+\bar\beta\bar h_\star$, $\widehat h_{\max}^{\mathrm{IW}}=2h_\star$, $z_{\max}=4cKd/(1-\alpha)$, and $\widetilde u_{\max}=(8cKB_\star+4d)/(1-\alpha)$ gives exactly the definition of $\kappa_{\mathrm{IW}}$ in \Cref{thm:exact-iw-aggregate-spb}.

Using the definition of the constants, the bound in \eqref{eq:b3-adversarial-bound} can be written as
\begin{equation}
    \label{eq:b3-adversarial-expanded}
    R_T
    \lesssim 
        \spr*{
            \frac{cKd\,h_\star}{1-\alpha}
        }^{1/3}T^{2/3}
        +
        \sqrt{
            \frac{(cKB_\star+d)h_\star}{1-\alpha}T
        }
        +
        \kappa_{\mathrm{IW}}
.
\end{equation}
For the choice $\alpha_K = 1 - 1/(\log K \vee 2)$, we get that $h_\star\lesssim K^{1-\alpha}\lesssim 1$, since $K \mapsto K^{1-\alpha_K}$ admits finite values and converges to $e$ as $K$ goes to $+\infty$. We therefore obtain the bound presented in the theorem. 
\end{proof}

\subsection{Corrupted stochastic bound}\label{subsec::corrupted_agg}

We conclude the proof of Theorem~\ref{thm:exact-iw-aggregate-spb} by deriving the regret bound in the corrupted stochastic regime. 

\begin{proof}{Proof of Theorem~\ref{thm:exact-iw-aggregate-spb} (gap and corruption-dependent bound)} 
We recall that, for this part, we assume the finite-support conditions of Section~\ref{sec::setting}, and use the notation $\mu_{\min}$, $\Delta_{\min}$, $a^\star \spr*{x}$, and $\spr*{\theta_a}_{a \in \sbr*{K}}$ defined there. To match the normalization in \Cref{thm:spb-transfer}, set
\begin{equation*}
    \Delta
    :=
    \min \scbr*{\Delta_{\min}, 1}
    \in
    (0,1].
\end{equation*}
Our goal is to apply the stochastic conversion in Theorem~\ref{thm:spb-transfer}, after verifying the necessary conditions. We start from the regret bound \eqref{eq::RT_agg_bt}. For every $t$ and $x$, define
\begin{equation*}
    e_t \spr*{x}
    :=
    1 - q_t \spr*{a^\star \spr*{x} \mid x},
    \qquad
    \bar e_t
    :=
    \bbE_{X \sim \cD} \sbr*{e_t \spr*{X}},
    \qquad
    e_t^{\max}
    :=
    \max_{x \in \cX}e_t \spr*{x}.
\end{equation*}
Since every context has probability at least $\mu_{\min}$,
\begin{equation*}
    e_t^{\max}
    \le
    \frac{\bar e_t}{\mu_{\min}}.
\end{equation*}
We first relate the quantities $s_t \spr*{x}$ and $Z_t \spr*{x}$ to $e_t \spr*{x}$. For every context $x$,
\begin{equation}
    \label{eq:b3-s-leq-e}
    s_t \spr*{x}
    \le
    e_t \spr*{x}, 
\end{equation}
since we either have equality or $s_t\leq 1/2$ while $e_t\geq 1/2$. We also have
\begin{equation}
    \label{eq:b3-z-local-self-bound}
    Z_t \spr*{x}
    \le
    4e_t \spr*{x}^{2 - \alpha}.
\end{equation}
To prove this, first suppose $e_t \spr*{x} < 1/2$. Then $q_t \spr*{a^\star \spr*{x} \mid x} > 1/2$, so $I_t \spr*{x} = a^\star \spr*{x}$ and $s_t \spr*{x} = e_t \spr*{x}$. Since $\sum_{a \ne a^\star \spr*{x}}q_t \spr*{a \mid x} = e_t \spr*{x}$ and $2 - \alpha > 1$,
\begin{equation*}
    \sum_{a \ne a^\star \spr*{x}}q_t \spr*{a \mid x}^{2 - \alpha}
    \le
    e_t \spr*{x}^{2 - \alpha}.
\end{equation*}
Therefore, $Z_t \spr*{x} \le 2e_t \spr*{x}^{2 - \alpha}$. If $e_t \spr*{x} \ge 1/2$, then $Z_t \spr*{x} \le 1$, while $1 \le 2^{2 - \alpha}e_t \spr*{x}^{2 - \alpha} \le 4e_t \spr*{x}^{2 - \alpha}$. This proves \eqref{eq:b3-z-local-self-bound}.

Invoking Lemma~\ref{lem::entropy_ub} and taking the supremum over $x$ and using the definition of $e_t^{\max}$ gives
\begin{equation}
    \label{eq:b3-hmax-self-bound}
    h_t^{\max}
    \le
    \frac{(K-1)^{1-\alpha}}{\alpha} (e_t^{\max})^\alpha.
\end{equation}
Combining \eqref{eq:b3-z-local-self-bound} with the definition of $z_t$ and $\bbE \sbr*{L \spr*{X}} = d$, we get
\begin{equation}
    \label{eq:b3-z-self-bound}
    z_t
    \le
    \frac{16 c K d}{1 - \alpha}(e_t^{\max})^{2 - \alpha}.
\end{equation}
Moreover, by \eqref{eq:b3-s-leq-e},
\begin{equation*}
    u_t^{\mathrm{mag}}
    \le
    \frac{8 c K B_\star}{1 - \alpha}(e_t^{\max})^{1 - \alpha},
\end{equation*}
and \eqref{eq:b3-z-self-bound} gives
\begin{equation*}
    \frac{z_t}{cK}
    \le
    \frac{16d}{1 - \alpha}(e_t^{\max})^{2 - \alpha}.
\end{equation*}
Hence
\begin{equation}
    \label{eq:b3-utilde-self-bound}
    \widetilde u_t
    =
    u_t^{\mathrm{mag}}
    +
    \frac{z_t}{cK}
    \le
    \frac{8 c K B_\star}{1 - \alpha}(e_t^{\max})^{1 - \alpha}
    +
    \frac{16d}{1 - \alpha}(e_t^{\max})^{2 - \alpha}.
\end{equation}
We now verify the self-bounding sequence condition \eqref{eq:spb-transfer-self-sequence} of \Cref{thm:spb-transfer} with $\widehat h_{t + 1} = 2h_t^{\max}$ and theorem error sequence $e_t^{\mathrm{SPB}} := \bar e_t$. Equations \eqref{eq:b3-hmax-self-bound} and \eqref{eq:b3-z-self-bound} imply
\begin{equation*}
    \sqrt{z_t\widehat h_{t+1}}
    =
    \sqrt{2z_t h_t^{\max}}
    \le
    \sqrt{
        \frac{32 c K d\,(K-1)^{1-\alpha}}{\alpha (1 - \alpha)}
    }\,
    e_t^{\max}
    \le
    \sqrt{\rho_1}\,\bar e_t,
\end{equation*}
where
\begin{equation*}
    \rho_1
    :=
    \frac{32 c K d\,(K-1)^{1-\alpha}}{\alpha(1 - \alpha)\mu_{\min}^2}.
\end{equation*}
Similarly, using \eqref{eq:b3-hmax-self-bound}, \eqref{eq:b3-utilde-self-bound}, and $e_t^{\max} \le 1$, we have
\begin{equation*}
\begin{aligned}
    \widetilde u_t\widehat h_{t+1}
    =
    2\widetilde u_t h_t^{\max}
    &\le
    2 \frac{(K-1)^{1-\alpha}}{\alpha}
    \spr*{
        \frac{8 c K B_\star}{1 - \alpha}e_t^{\max}
        +
        \frac{16d}{1 - \alpha}(e_t^{\max})^2
    }  \\
    &\le
    \frac{2(K-1)^{1-\alpha} \spr*{8 c K B_\star + 16d}}{\alpha(1 - \alpha)}e_t^{\max}
    \le
    \rho_2\bar e_t,
\end{aligned}
\end{equation*}
where
\begin{equation*}
    \rho_2
    :=
    \frac{2(K-1)^{1-\alpha} \spr*{8 c K B_\star + 16d}}{\alpha(1 - \alpha)\mu_{\min}}.
\end{equation*}
Thus \eqref{eq:spb-transfer-self-sequence} holds with
\begin{equation}
    \label{eq:b3-rho-def}
    \rho
    :=
    \max \scbr*{\rho_1,\rho_2}.
\end{equation}
It remains to prove the self-bounding lower bound on regret. Under the corrupted-stochastic conditions of Section~\ref{sec::setting}, use the corruption budget $C_{\mathrm{corr}}=2\XMAX C_\theta$ defined there.
In the purely stochastic regime, $C_\theta = 0$ and hence $C_{\mathrm{corr}} = 0$. For any context $x$ and arm $a \ne a^\star \spr*{x}$,
\begin{equation*}
\begin{aligned}
    \inp*{x, \theta_{t,a} - \theta_{t,a^\star \spr*{x}}}
    &=
    \inp*{x, \theta_a - \theta_{a^\star \spr*{x}}}
    +
    \inp*{x, \theta_{t,a} - \theta_a}
    -
    \inp*{x, \theta_{t,a^\star \spr*{x}} - \theta_{a^\star \spr*{x}}} \\
    &\ge
    \Delta_{\min}
    -
    2\XMAX
    \max_{b \in \sbr*{K}}\norm*{\theta_{t,b} - \theta_b}_2.
\end{aligned}
\end{equation*}
The best policy in hindsight has cumulative conditional mean loss no larger than the fixed uncorrupted optimal policy $a^\star \spr*{\cdot}$, and the paid observation cost is nonnegative. Therefore,
\begin{equation*}
    R_T
    \ge
    \Delta_{\min}\,
    \bbE \sbr*{\sum_{t=1}^T\bar e_t}
    -
    C_{\mathrm{corr}}
    \ge
    \Delta\,
    \bbE \sbr*{\sum_{t=1}^T\bar e_t}
    -
    C_{\mathrm{corr}}.
\end{equation*}
This is the self-bounding lower-bound hypothesis \eqref{eq:spb-transfer-self-lower} of \Cref{thm:spb-transfer} with gap parameter $\Delta$, theorem error sequence $e_t^{\mathrm{SPB}} = \bar e_t$, and corruption level $C_{\mathrm{corr}}$.

Applying the BOBW-transfer part of \Cref{thm:spb-transfer} with the variable map \eqref{eq:b3-spb-variable-map}, $R_0$ from \eqref{eq:b3-R0-def}, and $\rho$ from \eqref{eq:b3-rho-def}, we obtain the corrupted-stochastic bound. Namely, for
\begin{equation}
    \label{eq:b3-tau-def}
    \tau
    :=
    \Delta^{-3}
    +
    \frac{C_{\mathrm{corr}}}{\Delta},
\end{equation}
and every $T \ge \tau$,
\begin{equation}
    \label{eq:b3-corrupted-final}
    R_T
    \le
    C
    \spr*{
        \frac{\rho}{\Delta^2}
        \log_+ \spr*{T\Delta^3}
        +
        \spr*{
            \frac{C_{\mathrm{corr}}^2\rho}{\Delta^2}
            \log_+ \spr*{\frac{T\Delta}{C_{\mathrm{corr}}}}
        }^{1/3}
        +
        \kappa_{\mathrm{IW}}'
    },
\end{equation}
where the second term is interpreted as zero when $C_{\mathrm{corr}} = 0$, and
\begin{equation}
    \label{eq:b3-kappa-iw-prime}
    \kappa_{\mathrm{IW}}'
    :=
    \kappa_{\mathrm{IW}}
    +
    \spr*{
        (z_{\max}\widehat h_{\max}^{\mathrm{IW}})^{1/3}     \tau^{2/3}
        +
        \sqrt{\widetilde u_{\max}\widehat h_{\max}^{\mathrm{IW}}\tau}
    }
.
\end{equation}
For $T < \tau$, the adversarial bound from \Cref{thm:spb-transfer} with horizon $\tau$ is dominated by the same remainder term in \eqref{eq:b3-kappa-iw-prime}, since $\tau \ge 1$.

In the purely stochastic regime, $C_{\mathrm{corr}} = 0$, and \eqref{eq:b3-corrupted-final} reduces to
\begin{equation*}
    R_T
    \le
    C
    \spr*{
        \frac{\rho}{\Delta^2}
        \log_+ \spr*{T\Delta^3}
        +
        \kappa_{\mathrm{IW}}'
    }.
\end{equation*}
Finally, the self-bounding constant can be summarized as
\begin{equation*}
    \rho
    \le
    C
    \max \scbr*{
        \frac{c K d\,K^{1 - \alpha}}
             {\alpha(1 - \alpha)\mu_{\min}^2},
        \frac{(c K B_\star + d)K^{1 - \alpha}}
             {\alpha(1 - \alpha)\mu_{\min}}
    }.
\end{equation*}
For $\alpha = 1 - 1/\log K$, this becomes, up to universal constants,
\begin{equation*}
    \rho
    =
    O\!\spr*{
        \log K\cdot
        \max \scbr*{
            \frac{c K d}{\mu_{\min}^2},
            \frac{c K B_\star + d}{\mu_{\min}}
        }
    }.
\end{equation*}
Together, these estimates give the desired best-of-both-worlds guarantee for the exact importance-weighted estimator. The scaling stated in the theorem follows by substituting the deterministic bounds and the bound on $\rho$. 

\end{proof}

	\section{Analysis of CE--SPB}
\label{sec:analysis-le-spb}

We now analyze \loc, formally proving its regret bound and justifying all the parameter tuning introduced in Section~\ref{sec::regret}.  

\subsection{Algorithm parameters and entropy-adaptive regret bound}
\label{sec:le-hybrid-choices}

Before detailing the analysis, we recall the main algorithm design. Let
\begin{equation}
 \Lambda_t
 :=
 \Lambda_0+\sum_{s=1}^t h_s(X_s),
 \qquad
 \Lambda_0= \max\{1,h_\star\},
 \label{eq:hybrid-Lambda}
\end{equation}
We use the inverse-learning-rate update
\begin{equation}
 \beta_{t+1}
 =
 \beta_t
 +\zeta\Lambda_t^{-1/2}
 +\xi\Lambda_t^{-1/3},
 \qquad \text{so} \quad
 \eta_t:=\beta_t^{-1},
 \label{eq:hybrid-beta-update}
\end{equation}
where $\zeta>0$ and $\xi\ge0$. 

For $c\geq 0$ and parameters $\nu,u>0$, we recall the local observation rule
\begin{equation}
 p_{t,a}(x)
 :=
 \min\left\{
 1,
 \frac{z_{t,a}(x)}{\nu\Lambda_{t-1}^{1/3}}
 +\frac{u}{\beta_t}
 \right\}.
 \label{eq:hybrid-observation-rule}
\end{equation}
When $c=0$, we set $p_{t,a}(x)\equiv1$ for all $(t,a,x)$; in this case we also
set $\nu=\xi=0$ and omit all terms involving $\nu$ or $\xi$ below.  We assume
\begin{equation}
 \beta_1\ge \max\{1,u\},
 \qquad
 u\ge \frac{4B_\star}{1-\alpha},
 \label{eq:hybrid-basic-init}
\end{equation}
and set $\bar\beta=0$. For every arm $a$, define
\begin{equation*}
 \Sigma_{t,a}
 :=
 \bbE_t[p_{t,a}(X_t)X_tX_t\transpose].
\end{equation*}
Since $u/\beta_t\le1$, the observation rule satisfies
$p_{t,a}(x)\ge u/\beta_t$. 
\begin{equation}
 \Sigma_{t,a}
 \succeq
 \frac{u}{\beta_t}\Sigma,
 \qquad
 \Sigma_{t,a}^{-1}
 \preceq
 \frac{\beta_t}{u}\Sigma^{-1},
 \label{eq:hybrid-safety-covariance}
\end{equation}
In particular, the exact-IW estimator is well-defined.

\paragraph{Generic upper bound} We now provide the main regret bound for \loc, obtained as a function of its parameters and the entropy clock $\Lambda_T$. This result will serve as a basis to prove all the theoretical guarantees satisfied by \loc. It is based on the stability-penalty-observation cost decomposition, and upper bounding each of these terms individually. We state the result first, and detail the supporting technical lemmas in the remainder of this section.

\begin{proposition}[Generic entropy-adaptive bound]
	\label{prop:hybrid-master}
	Assume $c\geq 0$, Equation~\eqref{eq:hybrid-basic-init}, and $\xi, \zeta>0$. Then, the regret of \loc{} is upper bounded as follows,
	\begin{align}
		R_T
		\le& 	\quad	
		\beta_1h_\star
		+ \frac{4Kd}{(1-\alpha)\beta_1} + \frac{cK u }{\beta_1}
				+ \frac{4Kd \nu}{(1-\alpha)\beta_1} \;
		\bbE[\Lambda_T^{1/3}]\nonumber \\
		&+\left[
		4\zeta
		+
		\frac{
			4Kd/(1-\alpha)+cKu
		}{\zeta}
		\log\left(
		1+
		\frac{\zeta T}
		{\beta_1\Lambda_0^{1/2}}
		\right)
		\right]
		\bbE[\Lambda_T^{1/2}]
		\nonumber\\
		&+
		\left[
		3\xi
		+
		\frac{4Kd\nu}{(1-\alpha)\xi}
		\log\left(
		1+
		\frac{\xi T}
		{\beta_1\Lambda_0^{1/3}}
		\right)
		+
		\frac{4c}{\nu(1-\alpha)}
		\right]
		\bbE[\Lambda_T^{2/3}].
		\label{eq:hybrid-master-bound}
	\end{align}
Additionally, when $c=0$ we set $p_{t,a}(x)\equiv 1$ and the tuning $\nu=\xi=0$ becomes valid. Corresponding terms can be omitted, and they vanish smoothly under the tuning of Section~\ref{sec:le-hybrid-adaptive} as $c\to0$.
\end{proposition}

Before proving the result, we note that the condition $\beta_1\geq \frac{4B_\star}{1-\alpha}$ combined with $B_\star\geq d$ could already allow us to simplify the middle terms of the first line to $(1+c)K$, but we will show a refinement in $K$ under the final tuning of $\beta_1$, hence we keep this general expression for the moment.

\begin{proof}
To start, we apply the stability-penalty-observation cost decomposition from Appendix~\ref{sec:notation-algorithm}: exact importance weighting and the
ghost-context argument provide
\begin{equation}
	R_T
	\le
	\texttt{stability}_T
	+
	\texttt{penalty}_T
	+
	\texttt{obs}_T,
	\label{eq:le-hybrid-decomposition}
\end{equation}

The rest of the proof consists in upper bounding each of the three terms separately: Lemma~\ref{lem::stability_loc} gives an upper bound on the stability term, Lemma~\ref{lem:hybrid-penalty} upper bounds the penalty term, while Lemma~\ref{lem:hybrid-observation-cost} upper bounds the observation cost. The design of the algorithm guarantees that these bounds display terms of the same order of magnitude in terms of $\Lambda_T$. The bound displayed above simply regroups all terms of the same scaling from these three upper bounds. The lemmas are presented and formally proved in the remainder of this section.
\end{proof}

Proposition~\ref{prop:hybrid-master} is the most important result of this section: it allows us to tune all the algorithm parameters in order to balance the multiplicative factors in front of the entropy terms, which then yields the adversarial scaling introduced in the main paper. We detail this in Appendix~\ref{sec:le-hybrid-adaptive}. Second, the entropy-adaptive bound further allows us to prove a gap-dependent bound for the corrupted stochastic case, via self-bounding arguments. We detail the conversion in Appendix~\ref{sec:le-hybrid-stochastic}.

\subsection{Stability term}
\label{sec:le-hybrid-stability}

Thanks to the algorithmic design recalled in the previous section, we prove the following upper bound on the stability term. 

\begin{lemma}[Upper bound on the stability term]\label{lem::stability_loc}
Under the algorithmic choices of Section~\ref{sec:le-hybrid-choices}, the stability term of the regret  upper bound of \loc{} satisfies
	\begin{align}
		\stability
		\le \frac{4Kd}{1-\alpha}\bigg(
		&\frac1{\beta_1}
		+\frac{1}{\zeta}\;\log\left(
		1+\frac{\zeta T}{\beta_1\Lambda_0^{1/2}}
		\right)
		\bbE[\Lambda_T^{1/2}]
		\nonumber\\
		&+\frac{\nu}{\beta_1}\bbE[\Lambda_T^{1/3}]
		+\frac{\nu}{\xi} \log\left(
		1+\frac{\xi T}{\beta_1\Lambda_0^{1/3}}
		\right)
		\bbE[\Lambda_T^{2/3}]
		\bigg).
		\label{eq:hybrid-stability-adaptive}
	\end{align}
\end{lemma}

\begin{proof}
We first verify the deterministic condition of the local Tsallis stability lemma \ref{lem:stability-tsallis-bound}.
By \eqref{eq:hybrid-safety-covariance}, the boundedness $|Y_t(a)|\le1$, and
Cauchy--Schwarz in the $\Sigma_{t,a}^{-1}$ metric,
\begin{align*}
 \abs*{\wh\ell_t^{\mathrm{IW}}(x,a)}
 &\le
 \abs*{x\transpose\Sigma_{t,a}^{-1}X_t}
 \\
 &\le
 \sqrt{x\transpose\Sigma_{t,a}^{-1}x}
 \sqrt{X_t\transpose\Sigma_{t,a}^{-1}X_t}
 \\
 &\le
 \frac{\beta_t}{u}B_\star.
\end{align*}
Hence \eqref{eq:hybrid-basic-init} gives
\begin{equation}
 \eta_t\abs*{\wh\ell_t^{\mathrm{IW}}(x,a)}
 \le
 \frac{1-\alpha}{4}
 \le
 \frac{1-\alpha}{4}s_t(x)^{\alpha-1},
 \label{eq:hybrid-local-stability-condition}
\end{equation}
because $s_t(x)\le1$ and $\alpha-1<0$. The deterministic condition is satisfied, hence we can apply Lemma~\ref{lem:stability-tsallis-bound} to bound the stability term according to the usual variance terms.

The rest of the proof consists in combining Lemmas~\ref{lem:hybrid-stability} and~\ref{lem:hybrid-beta-sums}, which we introduce separately below for clarity. The first lemma develops the stability bound from Lemma~\ref{lem:stability-tsallis-bound} to express it in terms of cumulative entropies and the inverse-learning rates. The second result further upper bounds it by final cumulative entropies via standard sum/integral comparisons.
\end{proof}

We start with the stability bound derived from Lemma~\ref{lem:stability-tsallis-bound}.

\begin{lemma}[stability bound]
 \label{lem:hybrid-stability}
 Under \eqref{eq:hybrid-basic-init}, for $c\geq 0$,
 \begin{equation}
 \stability
 \le
 \frac{4Kd}{1-\alpha}
 \bbE\left[
 \sum_{t=1}^T
 \frac{1+\nu\Lambda_{t-1}^{1/3}}{\beta_t}
 \right].
 \label{eq:hybrid-stability-first}
 \end{equation}
\end{lemma}

\begin{proof}
 Thanks to \eqref{eq:hybrid-local-stability-condition}, we use  Lemma~\ref{lem:stability-tsallis-bound} and first obtain
 \begin{equation*}
 \stability
 \le
 \frac{4}{1-\alpha}
 \bbE\sum_{t=1}^T\frac1{\beta_t}
 \bbE_t\left[
 \sum_{a=1}^K
 z_{t,a}(X_0)
 \wh\ell_t^{\mathrm{IW}}(X_0,a)^2
 \right],
 \end{equation*}
 where $X_0\sim\cD$ is an independent ghost context.

 Fix $t$ and $a$, and write
 \begin{equation*}
 r_t:=\nu\Lambda_{t-1}^{1/3},
 \qquad
 M_{t,a}
 :=
 \bbE_t[z_{t,a}(X_0)X_0X_0\transpose].
 \end{equation*}
 For $c\geq 0$, the observation probabilities satisfy, for every $x$,
 \begin{equation*}
 p_{t,a}(x)
 \ge
 \min\left\{1,\frac{z_{t,a}(x)}{r_t}\right\}
 \ge
 \frac{z_{t,a}(x)}{r_t+z_{t,a}(x)},
 \end{equation*}
which is valid even when $r_t=0$ under the convention $1/0 =+\infty$. Since $0\le z_{t,a}(x)\le1$, we obtain
 \begin{equation*}
 z_{t,a}(x)xx\transpose
 \preceq
 (r_t+1)p_{t,a}(x)xx\transpose.
 \end{equation*}
 Taking expectations over contexts yields
 \begin{equation}
 M_{t,a}
 \preceq
 (1+r_t)\Sigma_{t,a}.
 \label{eq:hybrid-clipping-psd}
 \end{equation}
 Using the exact-IW estimator and $|Y_t(a)|\le1$,
 \begin{align*}
 \bbE_t\left[
 z_{t,a}(X_0)\wh\ell_t^{\mathrm{IW}}(X_0,a)^2
 \right]&\le
 \text{Tr}\left(
 \Sigma_{t,a}^{-1}M_{t,a}
 \right)
 \\
 &\le
 (1+r_t)\text{Tr}\left(
 \Sigma_{t,a}^{-1}\Sigma_{t,a}
 \right)
 =
 d(1+r_t),
 \end{align*}
 where the second inequality follows from \eqref{eq:hybrid-clipping-psd}.
 Summing over $a\in[K]$ proves \eqref{eq:hybrid-stability-first}.
\end{proof}

Before going to the next result, we introduce a remark on the $Kd$ factor before the stability bound, showing that it can be removed in the full information case. 

\begin{remark}[Full-information stability]\label{rem::full_info_loc}
	When $c=0$, the stability bound can be sharpened by removing the factor $K$.
	Indeed, in this case $p_{t,a}(x)=1$ for every $(t,a,x)$, so that
	$\Sigma_{t,a}=\Sigma$ for every arm. Therefore,
	\begin{align*}
		\bbE_t\left[
		\sum_{a=1}^K
		z_{t,a}(X_0)
		\wh\ell_t^{\mathrm{IW}}(X_0,a)^2
		\right]
		&\le
		\sum_{a=1}^K
		\text{Tr}\left(
		\Sigma^{-1}
		\bbE_t[z_{t,a}(X_0)X_0X_0\transpose]
		\right)
		\\
		&=
		\bbE_t\left[
		X_0\transpose\Sigma^{-1}X_0
		\,Z_t(X_0)
		\right]
		\le d,
	\end{align*}
	where the last inequality follows from $Z_t(x)\le1$ and
	$\bbE[X\transpose\Sigma^{-1}X]=d$. Consequently,
	\begin{equation*}
		\stability
		\le
		\frac{4d}{1-\alpha}
		\bbE\left[
		\sum_{t=1}^T\frac1{\beta_t}
		\right],
	\end{equation*}
	instead of the $Kd$ dependence appearing in
	Lemma~\ref{lem:hybrid-stability}.
Thus, in the full-information regime the local stability complexity is $d$ rather than $Kd$. Retuning $\zeta$ accordingly recovers the usual $\sqrt{\log K}$ dependence on the number of arms, up to the additional logarithmic factor induced by the adaptive learning-rate analysis.
\end{remark}

We next prove the two inverse-learning-rate sums used throughout the proof. Similar upper bounds can be found in previous works, but we restate a proof for completeness.

\begin{lemma}[Inverse-learning-rate sums]
 \label{lem:hybrid-beta-sums}
 Under \eqref{eq:hybrid-beta-update}, pathwise,
 \begin{equation}
 \sum_{t=1}^T\frac1{\beta_t}
 \le
 \frac1{\beta_1}
 +
 \frac{\Lambda_T^{1/2}}{\zeta}
\log\left(
1+\frac{\zeta T}{\beta_1\Lambda_0^{1/2}}
\right).
 \label{eq:hybrid-sum-half}
 \end{equation}
 If $\xi>0$, then also
 \begin{equation}
 \sum_{t=1}^T
 \frac{\Lambda_{t-1}^{1/3}}{\beta_t}
 \le
 \frac{\Lambda_T^{1/3}}{\beta_1}
 +
 \frac{\Lambda_T^{2/3}}{\xi}
 \log\left(
 1+\frac{\xi T}{\beta_1\Lambda_0^{1/3}}
 \right).
 \label{eq:hybrid-sum-third}
 \end{equation}
\end{lemma}

\begin{proof}
 Since $\Lambda_s\le\Lambda_T$ for $s\le T$,
 \begin{equation*}
 \beta_t
 \ge
 \beta_1+(t-1)\frac{\zeta}{\Lambda_T^{1/2}}.
 \end{equation*}
 Harmonic comparison therefore gives
 \begin{equation*}
 \sum_{t=1}^T\frac1{\beta_t}
 \le
 \frac1{\beta_1}
 +
 \frac{\Lambda_T^{1/2}}{\zeta}
 \log\left(
 1+\frac{\zeta T}{\beta_1\Lambda_T^{1/2}}
 \right),
 \end{equation*}
 which implies \eqref{eq:hybrid-sum-half} because $\Lambda_T\ge\Lambda_0$.
 Similarly, when $\xi>0$,
 \begin{equation*}
 \beta_t
 \ge
 \beta_1+(t-1)\frac{\xi}{\Lambda_T^{1/3}},
 \end{equation*}
 and $\Lambda_{t-1}^{1/3}\le\Lambda_T^{1/3}$, yielding
 \eqref{eq:hybrid-sum-third}.
\end{proof}

\subsection{Penalty term}
\label{sec:le-hybrid-penalty}

Using the update rule of the inverse-learning rate, we obtain the following upper bound on the penalty term. 
\begin{lemma}[Penalty]
 \label{lem:hybrid-penalty}
 Under \eqref{eq:hybrid-Lambda}--\eqref{eq:hybrid-beta-update},
 \begin{equation}
 \pen
 \le
 \beta_1h_\star
 +4\zeta\;\bbE[\Lambda_T^{1/2}]
 +3\xi\;\bbE[\Lambda_T^{2/3}].
 \label{eq:hybrid-penalty-bound}
 \end{equation}
\end{lemma}

\begin{proof}
 Since $\bar\beta=0$ and the comparator is deterministic, the penalty term from
 Appendix~\ref{sec:notation-algorithm} satisfies
 \begin{equation*}
 \pen
 \le
 \beta_1h_\star
 +
 \bbE \left[\sum_{t=1}^T 
 (\beta_{t+1}-\beta_t)
 H_\alpha(q_{t+1}(\cdot\mid X_0))\right].
 \end{equation*}
 Extend the i.i.d. context sequence by one independent context $X_{T+1}\sim\cD$.
 Since $q_{t+1}$ is $\cH_t$-measurable, $X_0$ can be replaced in expectation
 by $X_{t+1}$.  Thus, using \eqref{eq:hybrid-beta-update},
 \begin{align*}
 \texttt{penalty}_T
 \le
 \beta_1h_\star
 +\zeta\;\bbE \left[\sum_{t=1}^T
 \frac{h_{t+1}(X_{t+1})}{\Lambda_t^{1/2}}\right]
+\xi\;\bbE \left[\sum_{t=1}^T
 \frac{h_{t+1}(X_{t+1})}{\Lambda_t^{1/3}} \right].
 \end{align*}
 By $h_{t+1}(X_{t+1})\le h_\star\le\Lambda_0\le\Lambda_t$,
 \begin{equation*}
 \Lambda_{t+1}
 =\Lambda_t+h_{t+1}(X_{t+1})
 \le2\Lambda_t.
 \end{equation*}
 Hence, for $r\in\{1/2,1/3\}$,
 \begin{equation*}
 \frac{h_{t+1}(X_{t+1})}{\Lambda_t^r}
 \le
 2^r\frac{\Lambda_{t+1}-\Lambda_t}{\Lambda_{t+1}^r}.
 \end{equation*}
 Integral comparison gives
 \begin{equation*}
 \sum_{t=1}^T
 \frac{\Lambda_{t+1}-\Lambda_t}{\Lambda_{t+1}^r}
 \le
 \frac{\Lambda_{T+1}^{1-r}-\Lambda_1^{1-r}}{1-r}.
 \end{equation*}
 Finally, $\Lambda_{T+1}\le2\Lambda_T$, so the sums are bounded by
 $4\Lambda_T^{1/2}$ and $3\Lambda_T^{2/3}$
 respectively, as in both cases the multiplicative constant becomes $2^r \frac{1}{1-r} 2^{1-r}=\frac{2}{1-r}$.
 This proves \eqref{eq:hybrid-penalty-bound}.
\end{proof}

\subsection{Observation cost}
\label{sec:le-hybrid-observation}

We obtain the following upper bound for the observation. 

\begin{lemma}[Observation cost]
 \label{lem:hybrid-observation-cost}
 For $c\geq 0$ and $\nu, \xi>0$ we have that 
 \begin{equation}
 \obs
 \le
 \frac{4c}{\nu(1-\alpha)}\bbE[\Lambda_T^{2/3}]
 +\frac{cKu}{\beta_1}
 +\frac{cKu \log\left(
 	1+\frac{\zeta T}{\beta_1\Lambda_0^{1/2}}
 	\right)}{\zeta}
 \bbE[\Lambda_T^{1/2}].
 \label{eq:hybrid-observation-adaptive}
 \end{equation}
Additionally, if $c=0$ then $\obs=0$ by definition.
\end{lemma}

\begin{proof}
The observation rule directly gives
 \begin{align*}
 \obs
 &\le
 \frac{c}{\nu} \;
 \bbE\left[\sum_{t=1}^T
 \frac{Z_t(X_t)}{\Lambda_{t-1}^{1/3}}\right]
 +cKu \; \bbE\left[\sum_{t=1}^T\frac1{\beta_t}\right] \;.
\end{align*}
We remark that the second term is already present in the stability upper bound from Lemma~\ref{lem:hybrid-stability}, while the first term is very similar to a component of the penalty upper bound, except that $Z_t(X_t)$ should be replaced by $h_t(X_t)$. From Lemma~\ref{lem:ub_CalphaZ}, we know that this replacement costs at most a factor $C_\alpha^Z\coloneqq \sup_{q \in \simplex_K} \frac{Z_\alpha(q)}{H_\alpha(q)}\leq \frac{2\alpha}{1-\alpha}$. We can thus write that
\begin{align*}
 \obs&\le \;
 \frac{2c \alpha}{\nu (1-\alpha)}
 \bbE\left[\sum_{t=1}^T
 \frac{h_t(X_t)}{\Lambda_{t-1}^{1/3}}\right]
 +cKu \;\bbE\left[\sum_{t=1}^T\frac1{\beta_t}\right],
 \end{align*}
and then use respectively the proof of Lemma~\ref{lem:hybrid-penalty} for the first term,  and \eqref{eq:hybrid-sum-half} for the second term, to obtain the final result.
\end{proof}

\subsection{Parameter tuning and simplification of the upper bound}
\label{sec:le-hybrid-adaptive}

We now tune the parameters directly from
Proposition~\ref{prop:hybrid-master}. We first choose the magnitude-control
parameter at its smallest admissible value,
\begin{equation}
	u:=\frac{4B_\star}{1-\alpha},
	\label{eq:hybrid-u-choice}
\end{equation}
and take
\begin{equation}
	\Lambda_0= \max\{1,h_\star\}.
	\label{eq:hybrid-Lambda0-choice}
\end{equation}
We further choose
\begin{equation*}
	\beta_1
	=
	\max\{1,u,2\zeta,2\xi\}.
	\label{eq:hybrid-beta-log-condition}
\end{equation*}
For $T\ge2$, this simplifies the logarithms in the bound to $\log(T)$, more precisely
\begin{equation}
	\log\left(
	1+\frac{\zeta T}{\beta_1\Lambda_0^{1/2}}
	\right)
	\le \log T,
	\qquad
	\log\left(
	1+\frac{\xi T}{\beta_1\Lambda_0^{1/3}}
	\right)
	\le \log T.
	\label{eq:hybrid-log-bounds}
\end{equation}

We tune the parameters in order to balance the multiplicative factors of the two dominant entropy-dependent contributions in
Proposition~\ref{prop:hybrid-master}, starting with the $\Lambda_T^{2/3}$ term as it dominates the bound in the adversarial and non-negligible cost regime. 

\paragraph{Paid-observations term} We start with the
$\Lambda_T^{2/3}$ term, which corresponds to the paid-observation
component. By \eqref{eq:hybrid-log-bounds}, its coefficient is bounded by
\begin{equation}
	3\xi
	+
	\frac{
		4Kd\nu\log(T)
	}{
		(1-\alpha)\xi
	}
	+
	\frac{4c}{\nu(1-\alpha)},
	\label{eq:hybrid-third-coefficient}
\end{equation}
which is balanced by choosing
%
\begin{equation}
	\nu
	:=
	\left(
	\frac{
		4c^2
	}{
		3Kd(1-\alpha)\log(T)
	}
	\right)^{1/3},
	\qquad
	\xi
	:=
	\left(
	\frac{
		16cKd\log(T)
	}{
		9(1-\alpha)^2
	}
	\right)^{1/3}.
	\label{eq:hybrid-nu-xi-choice}
\end{equation}
With this choice, each term in
\eqref{eq:hybrid-third-coefficient} equals $3\xi$, and hence the whole
coefficient is $9\xi$.

\paragraph{Costless-limit dominant contribution} We next tune the $\Lambda_T^{1/2}$ term, which dominates in the
full-information limit. Since
$u=4B_\star/(1-\alpha)$, its coefficient is bounded by
\begin{align*}
	4\zeta
	+
	\frac{
		4Kd/(1-\alpha)+cKu
	}{\zeta}\log(T)
	&=
	4\zeta
	+
	\frac{
		4K(d+cB_\star)\log(T)
	}{
		(1-\alpha)\zeta
	}.
\end{align*}
We balance these two contributions by taking
\begin{equation}
	\zeta
	:=
	\left(
	\frac{
		K(d+cB_\star)\log(T)
	}{
		1-\alpha
	}
	\right)^{1/2}.
	\label{eq:hybrid-zeta-choice}
\end{equation}
Both contributions then equal $4\zeta$, so that the resulting
coefficient is $8\zeta$.

\paragraph{Initial learning rate} Now that $\xi, \zeta$ are tuned, we can provide the explicit initial inverse-learning rate as 
\begin{equation}
	\beta_1
=	\max\left\{
	\frac{4B_\star}{1-\alpha},\,
	2\left(
	\frac{
		K(d+cB_\star)\log(T)
	}{
		1-\alpha
	}
	\right)^{1/2},\,
	2\left(
	\frac{
		16cKd\log(T)
	}{
		9(1-\alpha)^2
	}
	\right)^{1/3}
	\right\},
	\label{eq:hybrid-beta1-choice}
\end{equation}
where we removed the $1$ because $u\geq 1$ is guaranteed by $B_\star \geq d$ and $\alpha\in (0,1)$. This final tuning simultaneously guarantees the magnitude condition and
\eqref{eq:hybrid-log-bounds}.

\paragraph{Simplifying the bound} Substituting these choices into
Proposition~\ref{prop:hybrid-master} gives the simplifying expression
\begin{align}
	R_T
	\le&
	8\zeta\,\bbE[\Lambda_T^{1/2}]
	+
	9\xi\,\bbE[\Lambda_T^{2/3}]
	+
		\beta_1h_\star + 
\frac{4Kd}{(1-\alpha)\beta_1} + \frac{cK u }{\beta_1}+
	\frac{
		4Kd\nu
	}{
		(1-\alpha)\beta_1
	}
	\bbE[\Lambda_T^{1/3}].
	\label{eq:hybrid-tuned-before-young}
\end{align}
It remains to control the last, lower-order term. By Young's inequality,
\begin{align*}
	\frac{
		4Kd\nu
	}{
		(1-\alpha)\beta_1
	}
	\Lambda_T^{1/3}
	&\le
	\xi\Lambda_T^{2/3}
	+
	\frac{1}{4\xi}
	\left(
	\frac{
		4Kd\nu
	}{
		(1-\alpha)\beta_1
	}
	\right)^2.
\end{align*}
The balancing relation
\begin{equation*}
	\frac{
		4Kd\nu\log(T)
	}{
		(1-\alpha)\xi
	}
	=3\xi
\end{equation*}
implies
\begin{equation*}
	\frac{4Kd\nu}{1-\alpha}
	=
	\frac{3\xi^2}{\log(T)}.
\end{equation*}
Hence
\begin{align*}
	\frac{1}{4\xi}
	\left(
	\frac{
		4Kd\nu
	}{
		(1-\alpha)\beta_1
	}
	\right)^2
	&=
	\frac{9\xi^3}
	{4\log^2(T)\beta_1^2}
	=
	\frac{
		4cKd
	}{
		(1-\alpha)^2\log(T)\beta_1^2
	}  \leq 	\frac{
	cKd
}{
4B_\star^2 \log(T)
} \leq cK \; , 
\end{align*}
where the last equality follows from
\eqref{eq:hybrid-nu-xi-choice}, the inequality from $\beta_1 \geq \frac{4B_\star}{1-\alpha}$, and the last inequality uses the crude bound $4B_\star^2\log(T)\geq d$ for $T\geq 2$. While we will use the sharper inequality involving $\beta_1$ below, this shows that this term is non-increasing with $T$.

We therefore obtain the following
entropy-adaptive bound.

\begin{proposition}[Tuned entropy-adaptive bound]
	\label{prop:hybrid-tuned}
	Let $c\geq 0$ and $T\ge2$, and choose $u,\Lambda_0,\nu,\xi,\zeta$ and
	$\beta_1$ according to
	\eqref{eq:hybrid-u-choice},
	\eqref{eq:hybrid-Lambda0-choice},
	\eqref{eq:hybrid-nu-xi-choice},
	\eqref{eq:hybrid-zeta-choice}, and
	\eqref{eq:hybrid-beta1-choice}. Then
	\begin{equation}
		R_T
		\le
		8\zeta\,\bbE[\Lambda_T^{1/2}]
		+
		10\xi\,\bbE[\Lambda_T^{2/3}]
		+
		R_{\mathrm{low}},
		\label{eq:hybrid-adaptive-final}
	\end{equation}
for some $R_\mathrm{low}$ satisfying 
	\begin{equation}
		R_{\mathrm{low}}
		\leq 
		\beta_1 h_\star
		+
		\frac{
			4K(d+cB_\star)
		}{
			(1-\alpha)\beta_1
		}
		+
		\frac{
			4cKd
		}{
			(1-\alpha)^2
			\log(T)\beta_1^2
		}.
		\label{eq:hybrid-Rlow}
	\end{equation}
Replacing $\xi,\zeta,\beta_1$ by their values and taking
$\alpha=1-1/\log(T)$ yields an upper bound of order at most
\begin{equation}
	R_T
	\lesssim
	\sqrt{K(d+cB_\star)\log(T)\log(K)\; T}
	+
	(cKd\log(T))^{1/3}(\log(K))^\frac{2}{3} T^\frac{2}{3} + B_\star \log(K), 
	\label{eq:hybrid-adversarial-simplified}
\end{equation}
where the last term is the only term from $R_{\mathrm{low}}$ that is not cleanly absorbed by the polynomial terms. 
\end{proposition}

\begin{proof}
Eq.~\eqref{eq:hybrid-adaptive-final} is direct from the above computations, as well as the upper bound on $R_\mathrm{low}$. For the polynomial terms, we use that $\Lambda_T \leq \Lambda_0 + T h_\star$.

It remains to prove the last statement regarding the scaling of $R_{\mathrm{low}}$ in all parameters. The key argument is that, for each term, we will use a different property of $\beta_1$. 

We also use that, for $K\le T$ and $T\geq 3$, the choice of $\alpha$ makes the maximal entropy scale inversely with $\log(T)$. More precisely, using that $e^x \leq 1+xe^x$ for $x\geq 0$, we get 
\begin{align*}
	K^{1/\log(T)}-1 & \leq \frac{\log(K)}{\log(T)}e^\frac{\log(K)}{\log(T)} \leq e \; \frac{\log(K)}{\log(T)},
\end{align*}
so
\begin{equation*}
	h_\star
	=
	\frac{K^{1/\log(T)}-1}{\alpha}
	\leq
\frac{e}{\alpha}	\frac{\log K}{\log T}.
\end{equation*}
Hence
\begin{align*}
	\beta_1h_\star
	&\leq \frac{2e}{\alpha} 
	\left(
	2  B_\star
	+ \frac{\xi}{\log(T)} + \frac{\zeta}{\log(T)} 
	\right)\log K, \\
\end{align*}
which can be upper bounded independently of $T$ as follows, 

\[\beta_1 h_\star \lesssim \left(B_\star + \sqrt{K(d+cB_\star)}+(cKd)^{1/3}\right)\log(K)\;.\]
Then, using simultaneously
$\beta_1\gtrsim
\log(T)\sqrt{K(d+cB_\star)}$ and $\beta_1\geq u$ 
for the second term of $R_{\mathrm{low}}$ gives
\begin{equation*}
	\frac{
		4K(d+cB_\star)
	}{
		(1-\alpha)\beta_1
	} 
	\lesssim
	\sqrt{K(d+cB_\star)} \wedge K(1+c) \lesssim \frac{\zeta}{\log(T)} .
\end{equation*}
Similarly, using
$\beta_1\gtrsim (cKd)^{1/3}\log(T)$, the third term is upper bounded by
\begin{equation*}
	\frac{
		4cKd
	}{
		(1-\alpha)^2\log(T)\beta_1^2
	}
	\lesssim
	\frac{(cKd)^{1/3}}{\log T} \lesssim \frac{\xi}{\log(T)^2}.
\end{equation*}
The last two terms are therefore also non-increasing in $T$, and can be incorporated into the upper bound of the $\beta_1 h_\star$ term, up to increasing the hidden multiplicative constant. This is also the case for $\left(\sqrt{K(d+cB_\star)}+(cKd)^{1/3}\right)\log(K)$. We can thus conclude that $R_\mathrm{low}$ can be upper bounded independently of $T$, and that its only remaining component that needs  to appear in the detailed bounds is the $B_\star \log(K)$ term.
\end{proof}

\subsection{Corrupted stochastic bound}
\label{sec:le-hybrid-stochastic}

We now specialize the entropy-adaptive bound to the stochastic and
corrupted-stochastic regimes of Section~\ref{sec::setting}. Let
\begin{equation*}
	\pi^\star(x):=a^\star(x),
	\qquad
	\Delta:=\min\{\Delta_{\min},1\},
\end{equation*}
and define the probability of playing a suboptimal action by
\begin{equation*}
	e_t(x):=
	1-q_t(\pi^\star(x)\mid x),
	\qquad
	S_T:=
	\bbE\left[\sum_{t=1}^T e_t(X_t)\right].
\end{equation*}
The key point is that the entropy clock can be controlled by $S_T$,
while by definition $S_T$ is itself controlled by the regret through the stochastic
gap.
	
\begin{lemma}[Entropy and regret self-bounds]
	\label{lem:hybrid-entropy-self-bound}
	For every $\alpha\in(0,1)$,
	\begin{equation}
		\bbE\left[
		\sum_{t=1}^T h_t(X_t)
		\right]
		\le
		\frac{(K-1)^{1-\alpha}}{\alpha}
		T^{1-\alpha}S_T^\alpha.
		\label{eq:hybrid-entropy-self-bound}
	\end{equation}
Moreover, 
\begin{equation}
	S_T
	\le
	\frac{R_T+C_{\mathrm{corr}}}{\Delta}.
	\label{eq:hybrid-gap-self-bound}
\end{equation}
\end{lemma}

\begin{proof}
	We first apply Lemma~\ref{lem::entropy_ub} pointwise on each entropy $h_t(x)$, and then use Jensen's inequality to get
	\begin{align*}
		\bbE\left[\sum_{t=1}^T h_t(X_t)
	\right]	&\le
		\frac{(K-1)^{1-\alpha}}{\alpha}
		\sum_{t=1}^T
		\bbE[e_t(X_t)^\alpha]
		\\
		&\le
		\frac{(K-1)^{1-\alpha}}{\alpha}
		T^{1-\alpha}
		\left(
		\bbE\left[\sum_{t=1}^T e_t(X_t)\right]
		\right)^\alpha,
	\end{align*}
This proves
	\eqref{eq:hybrid-entropy-self-bound}.
	
	For the second claim, the stochastic gap assumption implies, for every
	context $x$,
	\begin{equation*}
		\sum_{a=1}^K q_t(a\mid x)
		\left(
		\langle x,\theta_a\rangle
		-
		\langle x,\theta_{\pi^\star(x)}\rangle
		\right)
		\ge
		\Delta e_t(x).
	\end{equation*}
	Replacing $\theta_a$ by $\theta_{t,a}$ changes the instantaneous
	difference by at most
	\begin{equation*}
		2X_{\max}
		\max_{a\in[K]}
		\norm{\theta_{t,a}-\theta_a}_2.
	\end{equation*}
	Summing over $t$ therefore gives
	\begin{equation*}
		\bbE\sum_{t=1}^T
		\left[
		\sum_{a=1}^K
		q_t(a\mid X_t)\ell_t(X_t,a)
		-
		\ell_t(X_t,\pi^\star(X_t))
		\right]
		\ge
		\Delta S_T-C_{\mathrm{corr}}.
	\end{equation*}
	The hindsight comparator has cumulative loss no larger than the fixed
	policy $\pi^\star$, and the observation costs are nonnegative. Hence
	\begin{equation*}
		R_T
		\ge
		\Delta S_T-C_{\mathrm{corr}},
	\end{equation*}
	which is equivalent to
	\eqref{eq:hybrid-gap-self-bound}.
\end{proof}

We can now combine these two self-bounds with the entropy-adaptive
guarantee. For every $r\in(0,1)$, subadditivity and Jensen's inequality give
\begin{equation}
	\bbE[\Lambda_T^r]
	\le
	\Lambda_0^r
	+
	\left(
	\bbE\left[\sum_{t=1}^T h_t(X_t)
	\right]\right)^r.
	\label{eq:hybrid-clock-moment}
\end{equation}

\begin{theorem}[Corrupted stochastic bound for \loc]
	\label{thm:le-spb-stochastic}
	Assume the stochastic or corrupted-stochastic conditions and the
	parameter choices of Proposition~\ref{prop:hybrid-tuned}, along with 
	\[T\geq 3, \quad  2\leq K \leq T, \quad \alpha=1-\frac{1}{\log(T)}\;.\] 
	Then
	\begin{align}
		R_T
		\lesssim& \quad
\frac{\zeta^2}{\Delta} + 	\frac{\xi^3}{\Delta^2} + 		\zeta
+
\xi		+
B_\star \log(K)
+
C_{\mathrm{corr}}  \;.		\label{eq:hybrid-stochastic-general}
	\end{align}
Instantiating with the values of $\zeta,\xi$ and dropping lower order terms, we finally obtain 
\begin{equation}
R_T\lesssim 	\frac{
		K(d+cB_\star)\log^2 T
	}{\Delta} + \frac{cKd \log^3(T)}{\Delta^2} + B_\star \log(K) + C_\text{corr} \;.
\end{equation}
\end{theorem}

\begin{proof} We start from Proposition~\ref{prop:hybrid-tuned} with 
	\[R_T \lesssim \; \zeta\,\bbE[\Lambda_T^{1/2}]
	+
	\xi\,\bbE[\Lambda_T^{2/3}]
	+
	B_\star \log(K),
	\]
	which by \eqref{eq:hybrid-clock-moment} immediately gives 
	\begin{align*}
	R_T \lesssim \zeta\Lambda_0^{1/2}
	+ \xi\Lambda_0^{2/3} +  \zeta \, \sqrt{	\bbE\left[\sum_{t=1}^T h_t(X_t)
		\right]} + \xi \, \left(	\bbE\left[\sum_{t=1}^T h_t(X_t)
	\right]\right)^\frac{2}{3} + B_\star \log(K) \;.  
	\end{align*}
We then apply Lemma~\ref{lem:hybrid-entropy-self-bound} to turn this into a self-bounding inequality. We remark that setting $\alpha = 1-\frac{1}{\log(T)}$ simplifies the expression, as we can use that 
\[\frac{(K-1)^{1-\alpha}}{\alpha} T^{1-\alpha} \lesssim 1 \;. \]  
Therefore, 	
\begin{align*}
		R_T
		\lesssim& \;
		\zeta
		+
		\xi		+ B_\star \log(K)	
		+
		\zeta \;		S_T^{\alpha/2} +
		\xi \;		S_T^{2\alpha/3} \\ 
		\lesssim & \; 		\zeta
		+
		\xi		+ B_\star \log(K)	
		+
		\zeta \;		S_T^{1/2} +
		\xi \;		S_T^{2/3},
	\end{align*}
where the last line is obtained by applying the inequality $x^p\leq 1+x^q$, valid $x\geq 0$ and $0<p\leq q$, to $S_T^{\alpha/2}$ and $S_T^{2\alpha/3}$. 

	Applying \eqref{eq:hybrid-gap-self-bound} to $S_T$ and setting
	\begin{equation*}
		Y:=R_T+C_{\mathrm{corr}},
	\end{equation*}
	we obtain an inequality of the form
	\begin{equation*}
		Y
		\leq
		a_1Y^{1/2}
		+
		a_2Y^{2/3}
		+
		b,
	\end{equation*}
	where the fact that the $\lesssim$ sign hides only universal constants implies that there exist universal non-negative constants $a_1',a_2', b'$ such that the above equation is valid with
	\begin{align*}
		a_1
		&= a_1' \,
		 \frac{\zeta}{\Delta^{1/2}} ,
		\\
		a_2 
		&= a_2'  \; \frac{
		\xi}{\Delta^{2/3}},
	\end{align*}
	and
	\begin{equation*}
		b
		= b' (
		\zeta + 
		\xi + B_\star \log(K) )
		+
		C_{\mathrm{corr}}.
	\end{equation*}
We then use that, in general, if
	\begin{equation*}
		y\le a_1y^{p_1}+a_2y^{p_2}+b,
		\qquad
		p_1,p_2\in(0,1),
	\end{equation*}
	then
	\begin{equation*}
		y
		\le
		3b
		+
		(3a_1)^{1/(1-p_1)}
		+
		(3a_2)^{1/(1-p_2)}.
	\end{equation*}
	Indeed, if $y$ were larger than the right-hand side, each of the
	three terms on the right-hand side of the first inequality would be
	strictly smaller than $y/3$, yielding a contradiction. Applying this
	with $p_1=1/2$ and $p_2=2/3$ proves
	\eqref{eq:hybrid-stochastic-general}.
\end{proof}

\subsection{Obstruction to tuning $1-\alpha$ independently of $\log(T)$}

The following result shows that making $1-\alpha$ depend on $\log(T)$ is unfortunately necessary to obtain a polylogarithmic stochastic bound for \loc. 

\begin{lemma}[Polynomial stochastic regret of \loc\ for fixed $\alpha$]
	\label{lem::unavoidable_log_loc}
	Fix $\alpha\in(0,1)$ independently of $T$. There exists a deterministic linear contextual bandit instance with $K=2$, $d=1$, and $c=0$
	such that \loc{} tuned with $\zeta
	= \sqrt{\frac{2\log T}{1-\alpha}}$ 
	satisfies
	\[R_T \gtrsim (\log T)^{\frac{1}{2-\alpha}}
T^{\frac{1-\alpha}{2-\alpha}},
	\]
	where the $\gtrsim$ sign hides constants depending on $1-\alpha$.
\end{lemma}

\begin{proof}
	Consider $\mathcal X=\{1\}$ and deterministic
	loss parameters
	\[
	\theta_1=0,
	\qquad
	\theta_2=\frac{1}{2},
	\]
Thus arm $1$ is uniquely optimal with gap
	$1/2$. Since $c=0$, all losses are observed at every round. Let $p_t$ denote the probability assigned to the suboptimal arm, and set
	$\delta:=1-\alpha>0$. Since the cumulative losses before round $t$ are
	$0$ and $(t-1)/2$, respectively, the first-order optimality condition
	for Tsallis-FTRL gives
	\[
	\frac{t-1}{2}
	=
	\beta_t
	\left(
	p_t^{-\delta}
	-
	(1-p_t)^{-\delta}
	\right).
	\]
	If $p_t\le 1/4$, then
	\[
	(1-p_t)^{-\delta}
	\le
	3^{-\delta}p_t^{-\delta},
	\]
	and we obtain
	\[
	p_t
	\ge
	(1-3^{-\delta})^{1/\delta}
	\left(
	\frac{2\beta_t}{t-1}
	\right)^{1/\delta}.
	\]	
	Furthermore, by monotonicity of $\Lambda_t$,
	\[
	\beta_t
	\ge
	\zeta\sum_{s=1}^{t-1}\Lambda_s^{-1/2}
	\ge
	\frac{\zeta(t-1)}{\sqrt{\Lambda_T}}.
	\]	
	Hence, at each round $t$ we have 
	\[ p_t \gtrsim \frac{\zeta^{1/\delta}}{\Lambda_T^\frac{1}{2\delta}} \wedge \frac{1}{4}\;.  \]
	Since this lower bound is constant and $R_T:=\frac{1}{2}\sum_{t=1}^T p_t$, we first obtain the (at least) polynomial bound
	\[
	R_T
	\gtrsim
	\zeta^{1/\delta}
	\frac{T}{\Lambda_T^\frac{1}{2\delta}} \wedge T.
	\]
	By concavity, we have that $\Lambda_T\lesssim \sum_{t=1}^T p_t^\alpha \leq T^{1-\alpha}R_T^\alpha$, which finally gives
	\[
	R_T^{1+\frac{\alpha}{2-2\alpha}} 
	\gtrsim \zeta^\frac{1}{1-\alpha} \;  T^{1-\frac{1-\alpha}{2-2\alpha}},	\]
	which proves the result by simplifying the powers.	
\end{proof}
	
	\section{Technical tools}\label{app::technical_tools}

In this appendix, we restate useful results from previous works and prove some technical results useful in our analyses. 

\subsection{FTRL and ghost-sample tools}

In this section we present useful results that serve to develop the main FTRL/linear contextual bandit proof framework of the paper. 

\begin{lemma}[\citealp{NeuO20}, Lemma~3] \label{lem:ghost-sample-neu}
    Let $\pi^\star$ be a fixed stochastic policy and $\cD$ be a probability distribution over $\cX$. Let $\spr*{\cH_t}_{t \ge 0}$ be the full interaction-history filtration. Assume that, conditional on $\cH_{t - 1}$, the context $X_t$ is distributed according to $\cD$ and is independent of the learner's fresh randomization at round $t$. Let $X_0 \sim \cD$ be independent of $\cH_T$ and of all learner randomization. For any $t \in \sbr*{T}$ and any action $a \in \sbr*{K}$, suppose that $\pi_t$ is $\cH_{t - 1}$-measurable, $\theta_{t,a}$ is $\cH_{t - 1}$-measurable, and $\wh{\theta}_{t, a}$ is an estimator such that $\bbE_t \sbr{\wh{\theta}_{t, a}} = \theta_{t, a}$. Then, 
    \begin{align*}
        \bbE \sbr*{\sum_{t, a} \spr*{\pi_t \spr*{a \given X_t} - \pi^\star \spr*{a \given X_t}} \inp*{X_t, \theta_{t, a}}}
        =
        \bbE \sbr*{\sum_{t, a} \spr*{\pi_t \spr*{a \given X_0} - \pi^\star \spr*{a \given X_0}} \inp{X_0, \wh{\theta}_{t, a}}}.
    \end{align*}
\end{lemma}

\begin{lemma}[\citealp{BanditBook}, Exercise~28.12.a] \label{lem::ftrl_regret_decomp}
    Let $\cA \subseteq \bbR^d$ be non-empty and convex and $y_1, \ldots, y_n \in \cL \subseteq \bbR^d$ be a sequence of loss vectors. Let $F_1, \ldots, F_n, F_{n+1}$ be a sequence of proper closed convex functions that are finite and continuous on $\cA$ and differentiable on $\operatorname{relint}(\cA)$. For $t \in \sbr*{n + 1}$, define $\Phi_t = F_t + \sum_{s=1}^{t-1} \inp*{\cdot, y_s}$, and assume $a_t \in \argmin_{a \in \cA} \Phi_t \spr*{a}$ is well-defined and belongs to $\operatorname{relint}(\cA)$. Then, for any $a \in \cA$,
    \begin{align*}
        \sumtn \inp*{a_t - a, y_t} &\le \sumtn \spr*{\inp*{a_t - a_{t+1}, y_t} - D_{F_t} \spr*{a_{t+1}, a_t}}\\
        &\quad+ F_{n+1} \spr*{a} - F_1 \spr*{a_1} + \sumtn \spr*{F_t \spr*{a_{t+1}} - F_{t+1} \spr*{a_{t+1}}}.
    \end{align*}
\end{lemma}

\begin{proof}[Proof of Lemma~\ref{lem::ftrl_regret_decomp}]
    For convenience, denote $S_t = \sumst y_s$ for any $t$ and $S_0 = 0$. By definition of $\Phi_t$, we have $\nabla \Phi_t \spr*{a_t} = \nabla F_t \spr*{a_t} + S_{t-1}$, and by the first-order optimality condition of $a_t$, we have that for any $z \in \cA$,
    \begin{equation*}
        \inp*{z - a_t, \nabla \Phi_t \spr*{a_t}} \ge 0.
    \end{equation*}
    In particular, for $z = a_{t+1}$, we have
    \begin{equation} \label{eq:ftrl-decomp-at-optimality}
        \inp*{a_{t+1} - a_t, \nabla F_t \spr*{a_t} + S_{t-1}} \ge 0.
    \end{equation}
    Furthermore, observe that
    \begin{align*}
        \Phi_t \spr*{a_{t+1}} - \Phi_t \spr*{a_t} &= F_t \spr*{a_{t+1}} - F_t \spr*{a_t} + \inp*{a_{t+1} - a_t, S_{t-1}}\\
        &= D_{F_t} \spr*{a_{t+1}, a_t} + \inp*{a_{t+1} - a_t, \nabla F_t \spr*{a_t} + S_{t-1}},
    \end{align*}
    where we used the definitions of the potential $\Phi_t$ and the Bregman divergence $D_{F_t}$. By Equation~\eqref{eq:ftrl-decomp-at-optimality}, we have
    \begin{equation} \label{eq:ftrl-decomp-potential-increase}
        \Phi_t \spr*{a_{t+1}} - \Phi_t \spr*{a_t} \ge D_{F_t} \spr*{a_{t+1}, a_t}.
    \end{equation}
    Next, we compare $\Phi_{t+1}$ and $\Phi_t$ at $a_{t+1}$. We have
    \begin{align*}
        \Phi_{t+1} \spr*{a_{t+1}} &= F_{t+1} \spr*{a_{t+1}} + \inp*{a_{t+1}, S_t}\\
        &= F_{t+1} \spr*{a_{t+1}} - F_t \spr*{a_{t+1}} + \Phi_t \spr*{a_{t+1}} + \inp*{a_{t+1}, y_t}.
    \end{align*}
    Rearranging, we have
    \begin{align*}
        \inp*{a_{t+1}, y_t} &= \Phi_{t+1} \spr*{a_{t+1}} - \Phi_t \spr*{a_{t+1}} + F_t \spr*{a_{t+1}} - F_{t+1} \spr*{a_{t+1}}\\
        &\leq \Phi_{t+1} \spr*{a_{t+1}} - \Phi_t \spr*{a_t} - D_{F_t} \spr*{a_{t+1}, a_t} + F_t \spr*{a_{t+1}} - F_{t+1} \spr*{a_{t+1}},
    \end{align*}
    where we used Equation~\eqref{eq:ftrl-decomp-potential-increase} in the last step. Summing over $t$, the sum telescopes and we have
    \begin{equation*}
        \sumtn \inp*{a_{t+1}, y_t} \le \Phi_{n+1} \spr*{a_{n+1}} - \Phi_1 \spr*{a_1} - \sumtn D_{F_t} \spr*{a_{t+1}, a_t} + \sumtn \spr*{F_t \spr*{a_{t+1}} - F_{t+1} \spr*{a_{t+1}}}.
    \end{equation*}
    Since $a_{n+1}$ minimizes $\Phi_{n+1}$ over $\cA$, we have $\Phi_{n+1} \spr*{a_{n+1}} \le \Phi_{n+1} \spr*{a} = F_{n+1} \spr*{a} + \inp*{a, S_n}$ for any $a \in \cA$. We also have $\Phi_1 \spr*{a_1} = F_1 \spr*{a_1}$ since $S_0 = 0$. Therefore,
    \begin{equation*}
        \sumtn \inp*{a_{t+1}, y_t} \le F_{n+1} \spr*{a} - F_1 \spr*{a_1} + \inp*{a, S_n} - \sumtn D_{F_t} \spr*{a_{t+1}, a_t} + \sumtn \spr*{F_t \spr*{a_{t+1}} - F_{t+1} \spr*{a_{t+1}}}.
    \end{equation*}
    Rearranging, we have
    \begin{equation*}
        \sumtn \inp*{a_{t+1} - a, y_t} \le F_{n+1} \spr*{a} - F_1 \spr*{a_1} - \sumtn D_{F_t} \spr*{a_{t+1}, a_t} + \sumtn \spr*{F_t \spr*{a_{t+1}} - F_{t+1} \spr*{a_{t+1}}}.
    \end{equation*}
    Finally, we can write $a_t - a = \spr*{a_t - a_{t+1}} + \spr*{a_{t+1} - a}$ and conclude.
\end{proof}

\begin{lemma}[Negative Tsallis entropy]\label{lem::negative_tsallis_entropy}
    Let $\alpha \in \spr*{0, 1}$ and define the Tsallis entropy $H_\alpha \colon \bbRnnK \to \bbR$ by
    \begin{equation*}
        H_\alpha \spr*{q} = \frac{1}{\alpha} \sumaK \spr*{q_a^\alpha - q_a}.
    \end{equation*}
    Then $-H_\alpha$ is convex on $\bbRnnK$ and differentiable on $\bbRpK$. Moreover, for any $q \in \bbRpK$,
    \begin{equation*}
        \nabla \spr*{-H_\alpha} \spr*{q} = \frac{1}{\alpha} \bfone - q^{\alpha - 1},
        \qquad
        \nabla^2 \spr*{-H_\alpha} \spr*{q} = \spr*{1 - \alpha} \Diag \spr*{q^{\alpha - 2}} \succeq 0.
    \end{equation*}
    Consequently, for any $p, q \in \bbRpK$, the Bregman divergence induced by $-H_\alpha$ is
    \begin{equation*}
        D_{-H_\alpha} \spr*{p, q}
        = \frac{1}{\alpha} \sumaK \spr*{q_a^\alpha - p_a^\alpha + \alpha q_a^{\alpha - 1} \spr*{p_a - q_a}}
        \geq 0,
    \end{equation*}
    with equality if and only if $p = q$.
\end{lemma}

\begin{proof}[Proof of Lemma~\ref{lem::negative_tsallis_entropy}]
    Let $f_\alpha \spr*{u} = \frac{1}{\alpha} \spr*{u - u^\alpha}$ for $u \geq 0$. Since $-H_\alpha \spr*{q} = \sumaK f_\alpha \spr*{q_a}$, it is enough to study the scalar function $f_\alpha$. For any $u > 0$,
    \begin{equation*}
        f_\alpha'' \spr*{u} = \spr*{1 - \alpha} u^{\alpha - 2} > 0.
    \end{equation*}
    Thus $f_\alpha$ is convex on $\bbR_{>0}$, and by continuity it is convex on $\bbR_{\geq 0}$. Hence $-H_\alpha$ is convex on $\bbRnnK$ and differentiable on $\bbRpK$. The displayed expressions for the gradient and Hessian follow by differentiating coordinate-wise.

    It remains to compute the Bregman divergence. For $p, q \in \bbRpK$,
    \begin{align*}
        D_{-H_\alpha} \spr*{p, q}
        &= -H_\alpha \spr*{p} + H_\alpha \spr*{q} - \inp*{\nabla \spr*{-H_\alpha} \spr*{q}, p - q} \\
        &= \frac{1}{\alpha} \sumaK \spr*{q_a^\alpha - p_a^\alpha + \alpha q_a^{\alpha - 1} \spr*{p_a - q_a}}.
    \end{align*}
    Since the Bregman divergence of a convex differentiable function is nonnegative, $D_{-H_\alpha} \spr*{p, q} \geq 0$. The equality condition follows from the strict convexity of $-H_\alpha$ on $\bbRpK$.
\end{proof}

\subsection{Tsallis entropy inequalities}

In this section we introduce inequalities involving Tsallis-related quantities, including the Tsallis entropy itself. 

\begin{lemma}[Tsallis entropy lower bound] \label{lem:tsallis-entropy-lower-bound}
    Let $\alpha \in \spr*{0, 1}$ and $q \in \simplex_K$. If $I \in \argmax_{a \in \sbr*{K}} q_a$ and $s = \min \scbr*{q_{I}, 1 - q_{I}}$, then
    \begin{equation*}
        H_\alpha \spr*{q}
        \ge \frac{1 - \alpha}{4\alpha}
s^\alpha \;.
    \end{equation*}
\end{lemma}

\begin{proof}[Proof of Lemma~\ref{lem:tsallis-entropy-lower-bound}]
    We consider two cases. If $q_I \le 1 / 2$, then $s = q_{I}$ and
    \begin{equation*}
    \begin{aligned}
        H_\alpha \spr*{q}
        &\ge
        \frac{1}{\alpha} \spr*{q_{I}^\alpha - q_{I}}
        =
        \frac{1}{\alpha} q_{I}^\alpha \spr*{1 - q_{I}^{1 - \alpha}}
        \ge
        \frac{1}{\alpha} s^\alpha \spr*{1 - 2^{\alpha - 1}}.
    \end{aligned}
    \end{equation*}
    If $q_{I} > 1 / 2$, then
    $s = 1 - q_{I} = \sum_{a \ne I} q_a$, and subadditivity of
    $u \mapsto u^\alpha$ on $\bbR_+$ gives
    $\sum_{a \ne I} q_a^\alpha \ge s^\alpha$. Therefore
    \begin{equation*}
    \begin{aligned}
        H_\alpha \spr*{q}
        &\ge
        \frac{1}{\alpha}
        \sum_{a \ne I } \spr*{q_a^\alpha - q_a}
        \ge
        \frac{1}{\alpha} \spr*{s^\alpha - s}
        =
        \frac{1}{\alpha} s^\alpha \spr*{1 - s^{1 - \alpha}}
        \ge
        \frac{1}{\alpha} s^\alpha \spr*{1 - 2^{\alpha - 1}}.
    \end{aligned}
    \end{equation*}
    In both cases, $1 - 2^{\alpha - 1} \ge \spr*{1 - \alpha} / 4$, which proves the result.
\end{proof}

\begin{lemma}[Tsallis entropy upper bound]\label{lem::entropy_ub} For any $\alpha \in (0,1)$ and every distribution $q \in \simplex_K$, for any arm $a^\star \in [K]$, we have 
	\[H_\alpha(q) \leq \frac{(K-1)^{1-\alpha}}{\alpha} (1-q_{a^\star})^\alpha \;. \]
\end{lemma}
\begin{proof} Let $s = 1-q_{a^\star}$.
	By concavity of $u\mapsto u^\alpha$, we first have
	\begin{equation*}
		\sum_{a\neq a^\star}
		q_a^\alpha
		\le
		(K-1)^{1-\alpha}s^\alpha.
	\end{equation*}
	Moreover $(1-s)^\alpha-(1-s)\le s$, as the expression is maximized in $\alpha=0$. We thus obtain
	\begin{align*}
		H_\alpha(q)
		&=
		\frac{1}{\alpha}
		\left[
		(1-s)^\alpha-(1-s)
		+
		\sum_{a\neq a^\star}
		\left(q_a^\alpha-q_a\right)
		\right]
		\\
		&\leq
		\frac{1}{\alpha}
		\left[s
		+ (K-1)^{1-\alpha} s^\alpha - s\right]
		\\
		&=
		\frac{(K-1)^{1-\alpha}}{\alpha}s^\alpha.
	\end{align*}
\end{proof}

\begin{lemma}[\citealp{ito2024competitive}, Lemma 22, \citealp{BOBWhardproblems}, Lemma 14] \label{lem:stability-tsallis-bound}
    Consider a probability distribution $q \in \operatorname{relint}(\simplex_K)$ and denote $\bar{I} \in \argmax_{i \in \sbr*{K}} q_i$. For $\alpha \in \spr*{0, 1}$, denote $H_\alpha \colon \bbRnnK \to \bbR$ the Tsallis entropy defined in Lemma~\ref{lem::negative_tsallis_entropy}. Furthermore, let $\ell \in \bbR^K$ be a vector such that, for all $i \in \sbr*{K}$,
    \begin{equation*}
        \abs*{\ell_i} \leq \frac{1 - \alpha}{4} \frac{1}{\min \scbr*{q_{\bar{I}}, 1 - q_{\bar{I}}}^{1 - \alpha}}.
    \end{equation*}
    Then, the following bound holds
    \begin{equation*}
        \max_{p \in \simplex_K} \scbr*{\inp*{\ell, q - p} - D_{- H_\alpha} \spr*{p, q}} \leq \frac{4}{1 - \alpha} \spr*{\sum_{i \neq \bar{I}} q_i^{2 - \alpha} \ell_i^2 + \min \scbr*{q_{\bar{I}}, 1 - q_{\bar{I}}}^{2 - \alpha} \ell_{\bar{I}}^2}.
    \end{equation*}
\end{lemma}

\begin{lemma}[Deterministic entropy growth for composite Tsallis FTRL, \citealp{ito2024competitive}, Lemma 24, \citealp{BOBWhardproblems}, Lemma 15] \label{lem::ftrl_smooth_bound}
    Let $\alpha \in \spr*{0, 1}$ and $\bar\alpha = 1 - \alpha$. Fix $L, \ell \in \bbR^K$, $0 < \beta \le \beta'$, and $\bar\beta > 0$. Let $q, r \in \Delta_K$ be defined by
    \begin{equation*}
        q
        \in
        \argmin_{p \in \Delta_K}
        \scbr*{
            \inp*{L, p}
            +
            \beta \spr*{- H_\alpha \spr*{p}}
            +
            \bar\beta \spr*{- H_{\bar\alpha} \spr*{p}}
        },
    \end{equation*}
    and
    \begin{equation*}
        r
        \in
        \argmin_{p \in \Delta_K}
        \scbr*{
            \inp*{L + \ell, p}
            +
            \beta' \spr*{- H_\alpha \spr*{p}}
            +
            \bar\beta \spr*{- H_{\bar\alpha} \spr*{p}}
        }.
    \end{equation*}
    Let
    \begin{equation*}
        I \in \argmax_{a \in \sbr*{K}} q_a,
        \qquad
        s \coloneqq \min \scbr*{q_I, 1 - q_I}.
    \end{equation*}
    If
    \begin{equation}
        \label{eq:penalty-det-loss-condition}
        \norm*{\ell}_\infty
        \le
        \max \scbr*{
            \frac{1 - \spr*{\sqrt{2}}^{\alpha - 1}}{2}\,\beta s^{\alpha - 1},
            \frac{1 - \spr*{\sqrt{2}}^{\bar\alpha - 1}}{2}\,
            \bar\beta s^{\bar\alpha - 1}
        },
    \end{equation}
    and
    \begin{equation}
        \label{eq:penalty-det-beta-condition}
        0
        \le
        \beta' - \beta
        \le
        \max \scbr*{
            \spr*{1 - \spr*{\sqrt{2}}^{\alpha - 1}}\beta,
            \frac{1 - \spr*{\sqrt{2}}^{\bar\alpha - 1}}{\sqrt{2}}\,
            \bar\beta s^{\bar\alpha - \alpha}
        },
    \end{equation}
    then
    \begin{equation*}
        H_\alpha \spr*{r} \le 2 H_\alpha \spr*{q}.
    \end{equation*}
\end{lemma}

\begin{lemma}[Useful bound] \label{lem:ub_CalphaZ}
	Let $\alpha \in (0,1)$ and $K \geq 2$. Then
	\begin{equation} \sup_{q \in \triangle_K} \frac{Z(q)}{H_\alpha(q)}
		=
		\frac{\alpha}
		{2^{1-\alpha}\bigl(2^{1-\alpha}-1\bigr)} \leq 		\frac{2\alpha}{1-\alpha}.
		\label{lem:CZ-exact}
	\end{equation}
\end{lemma}

\begin{proof}
	Let $\delta = 1-\alpha \in (0,1)$ and define
	\begin{equation*}
		g(u) = u^\alpha-u.
	\end{equation*}
	Then $\alpha H_\alpha(q)
	=
	\sum_{a=1}^K g(q_a)$. 
	Fix $q\in\Delta_K$ and let $I\in\argmax_a q_a$. Recall that
	\begin{equation*}
		Z(q)
		=
		\sum_{a\neq I}q_a^{1+\delta}
		+
		s^{1+\delta},
		\qquad
		s=\min\{q_I,1-q_I\}.
	\end{equation*}
	For $u\in(0,1/2]$,
	\begin{equation*}
		\frac{u^{1+\delta}}{g(u)}
		=
		\frac{u^{2\delta}}{1-u^\delta}.
	\end{equation*}
	The right-hand side is increasing in $u$, and hence
	\begin{equation}
		u^{1+\delta}
		\leq
		\frac{1}{2^\delta(2^\delta-1)}g(u),
		\qquad
		u\in[0,1/2].
		\label{eq:coordinate-Z-H}
	\end{equation}
	
	We distinguish two cases. If $q_I\leq 1/2$, then $s=q_I$ and $Z(q)
	=
	\sum_{a=1}^K q_a^{1+\delta}$.
	Applying~\eqref{eq:coordinate-Z-H} coordinatewise yields
	\begin{equation*}
		Z(q)
		\leq
		\frac{1}{2^\delta(2^\delta-1)}
		\sum_{a=1}^K g(q_a)
		=
		\frac{\alpha}{2^\delta(2^\delta-1)}
		H_\alpha(q).
	\end{equation*}
	
	Suppose now that $q_I>1/2$, and write $r=1-q_I=s<1/2$. Then
	\begin{equation*}
		Z(q)
		=
		r^{1+\delta}
		+
		\sum_{a\neq I}q_a^{1+\delta}.
	\end{equation*}
	Since $g$ is concave on $[0,1]$, $g(1)=0$, and
	\begin{equation*}
		g(1/2)
		=
		\frac{2^\delta-1}{2},
	\end{equation*}
	the chord between $1/2$ and $1$ gives
	\begin{equation*}
		g(q_I)
		\geq
		2(1-q_I)g(1/2)
		=
		r(2^\delta-1).
	\end{equation*}
	As $r\leq 1/2$,
	\begin{equation*}
		r^{1+\delta}
		\leq
		2^{-\delta}r
		\leq
		\frac{1}{2^\delta(2^\delta-1)}g(q_I).
	\end{equation*}
	Moreover, $q_a\leq 1/2$ for every $a\neq I$, so
	\eqref{eq:coordinate-Z-H} gives
	\begin{equation*}
		Z(q)
		\leq
		\frac{1}{2^\delta(2^\delta-1)}
		\sum_{a=1}^K g(q_a)
		=
		\frac{\alpha}{2^\delta(2^\delta-1)}
		H_\alpha(q).
	\end{equation*}
	Thus, in both cases we obtained
	\begin{equation*}
		\frac{Z(q)}{H_\alpha(q)}
		\leq
		\frac{\alpha}{2^\delta(2^\delta-1)}.
	\end{equation*}
	
	For the reverse inequality, take
	$q=(1/2,1/2,0,\ldots,0)$. Then
	\begin{equation*}
		Z(q)=2^{-\delta},
		\qquad
		H_\alpha(q)=\frac{2^\delta-1}{\alpha},
	\end{equation*}
	and therefore
	\begin{equation*}
		\frac{Z(q)}{H_\alpha(q)}
		=
		\frac{\alpha}{2^\delta(2^\delta-1)}.
	\end{equation*}
	This proves~\eqref{lem:CZ-exact}. Finally,
	$2^\delta-1=e^{\delta\log 2}-1\geq\delta\log 2$
	and $2^\delta\geq1$, which gives
	\begin{equation*}
		\frac{Z(q)}{H_\alpha(q)}
		\leq
		\frac{\alpha}{(1-\alpha)\log 2}\leq 		\frac{2\alpha}{1-\alpha}.
	\end{equation*}
\end{proof}

\subsection{SPB matching}

In this part we simply restate the main result from \cite{BOBWhardproblems}, that we use to prove the BOBW guarantees of \agg.

\begin{theorem}[SPB matching: summation and BOBW transfer, \citealp{BOBWhardproblems}]
\label{thm:spb-transfer}
Let $(z_t)_{t\ge1}$ and $(u_t)_{t\ge1}$ be nonnegative adapted sequences.
Let $(\widehat h_{t+1})_{t\ge1}$ be a predictable sequence of positive
available upper bounds on the next penalty component. Assume that, for all
$t\le T$,
\[
    z_t\le z_{\max},
    \qquad
    u_t\le u_{\max},
    \qquad
    0<\widehat h_{t+1}\le \widehat h_{\max}.
\]
Let $\beta_1>0$, and define the inverse learning rates by the Rule~2
SPB-matching update
\[
    \beta_{t+1}
    =
    \beta_t
    +
    \frac{1}{\widehat h_{t+1}}
    \left(
        2\sqrt{\frac{z_t}{\beta_t}}
        +
        \frac{u_t}{\beta_t}
    \right).
\]
For shorthand, define
\[
    b_t
    :=
    2\sqrt{\frac{z_t}{\beta_t}}
    +
    \frac{u_t}{\beta_t}.
\]
Then there exists a universal numerical constant $C>0$ such that, for every
$T\ge1$,
\begin{equation*}
    \sum_{t=1}^T b_t
    \le
    C
    \left(
        (z_{\max}\widehat h_{\max})^{1/3}T^{2/3}
        +
        \sqrt{u_{\max}\widehat h_{\max}T}
        +
        \sqrt{\frac{z_{\max}}{\beta_1}}
        +
        \frac{u_{\max}}{\beta_1}
        +
        \beta_1\widehat h_{\max}
    \right).
\end{equation*}
More generally, for every $\varepsilon\ge 1/T$,
\begin{equation*}
\begin{aligned}
    \sum_{t=1}^T b_t
    \le
    C\bigg[
    &
    \left(
        \sum_{t=1}^T
        \sqrt{
            z_t\widehat h_{t+1}\log(\varepsilon T)
        }
    \right)^{2/3}
    +
    \left(
        \frac{\sqrt{z_{\max}\widehat h_{\max}}}{\varepsilon}
    \right)^{2/3}                                      \\
    &+
    \sqrt{
        \sum_{t=1}^T
        u_t\widehat h_{t+1}\log(\varepsilon T)
    }
    +
    \sqrt{\frac{u_{\max}\widehat h_{\max}}{\varepsilon}}
    +
    \sqrt{\frac{z_{\max}}{\beta_1}}
    +
    \frac{u_{\max}}{\beta_1}
    +
    \beta_1\widehat h_{\max}
    \bigg].
\end{aligned}
\end{equation*}
The same statements hold after taking expectations if the sequences are random.

Now suppose that a regret quantity $R_T$ satisfies, for some nonnegative
constant $R_0$,
\begin{equation*}
    R_T
    \le
    R_0
    +
    2\,
    \mathbb E\!\left[
        \sum_{t=1}^T b_t
    \right].
\end{equation*}
Then the adversarial-regime bound
\begin{equation}
    \label{eq:spb-transfer-adversarial-regret}
    R_T
    \le
    C
    \left(
        (z_{\max}\widehat h_{\max})^{1/3}T^{2/3}
        +
        \sqrt{u_{\max}\widehat h_{\max}T}
        +
        \kappa
    \right)
\end{equation}
holds, where
\begin{equation*}
	\kappa
	:=
	R_0
	+
	\frac{\sqrt{z_{\max}}}{\beta_1}
	+
	\frac{u_{\max}}{\beta_1}
	+
	\beta_1 \wh{h}_{\max}.
\end{equation*}

Furthermore, suppose that the following self-bounding hypotheses hold: there
exist a nonnegative adapted sequence $(e_t)_{t\ge1}$, a gap parameter
$\Delta\in(0,1]$, a corruption level $C_{\mathrm{corr}}\ge0$, and constants
$\rho_1,\rho_2>0$ such that, for all $t$,
\begin{equation}
    \label{eq:spb-transfer-self-sequence}
    \sqrt{z_t\widehat h_{t+1}}
    \le
    \sqrt{\rho_1}\,e_t,
    \qquad
    u_t\widehat h_{t+1}
    \le
    \rho_2 e_t,
\end{equation}
and such that the regret satisfies the self-bounding lower bound
\begin{equation}
    \label{eq:spb-transfer-self-lower}
    R_T
    \ge
    \Delta\,
    \mathbb E\!\left[
        \sum_{t=1}^T e_t
    \right]
    -
    C_{\mathrm{corr}}.
\end{equation}
Let $\rho:=\max\{\rho_1,\rho_2\}$ and
\[
    \tau
    :=
    \Delta^{-3}
    +
    \frac{C_{\mathrm{corr}}}{\Delta}.
\]
Then, whenever $T\ge 1$,
\begin{equation}
    \label{eq:spb-transfer-bobw}
    R_T
    \le
    C
    \left(
        \frac{\rho}{\Delta^2}
        \log_+(T\Delta^3)
        +
        \left(
            \frac{C_{\mathrm{corr}}^2\rho}{\Delta^2}
            \log_+\!\left(\frac{T\Delta}{C_{\mathrm{corr}}}\right)
        \right)^{1/3}
        +
        \kappa'
    \right),
\end{equation}
where $\log_+(x):=\max\{1,\log x\}$, the second term is interpreted as zero
when $C_{\mathrm{corr}}=0$, and
\[
    \kappa'
    :=
    \kappa
    +
    \left(
        (z_{\max}\widehat h_{\max})^{1/3} \tau^{2/3}
        +
        \sqrt{u_{\max}\widehat h_{\max}\tau}
    \right),
\]
so $\kappa'$ includes the adversarial regret bound corresponding to the case $T\leq \tau$.
\end{theorem}

\begin{proof}
    This is a restatement of the Rule~2 part of the SPB-matching summation theorem of Tsuchiya and Ito~\cite[Theorem~6]{BOBWhardproblems}, together with their best-of-both-worlds transfer theorem~\cite[Theorem~7]{BOBWhardproblems}. The only notational change is that we write $\widehat h_{t+1}$ for the available upper bound on the next penalty component. In applications in which the algorithm uses the update with $\widehat h_{t+1}$ and proves that the true next penalty component is at most $\widehat h_{t+1}$, the theorem applies directly. The stochastic and corrupted-stochastic statement follows from the same self-bounding argument as in the proof of their Theorem~7.
\end{proof}

	\section{Complete pseudo-codes and implementation remarks}\label{app::algos}

In this section we present the pseudo-code of the algorithms presented in this work, for generic admissible parameters. We refer to the corresponding theorems for the precise tuning used in our analysis. 

To simplify the presentation, both use a Tsallis-FTRL subroutine introduced in \Cref{alg:tsallis-ftrl-routine}, which takes as input previous parameter estimates and a context in order to return the corresponding Tsallis-FTRL allocation, as well as relevant quantities used by our algorithms. 

We introduce the pseudo-code of \agg{} in \Cref{alg:clipped-aggregate-spb-iw}, and the pseudo-code of \loc{} in \Cref{alg:kuroki-style}.

\begin{algorithm}[ht]
\caption{Aggregate SPB--matching with FTRL (\agg)}
\label{alg:clipped-aggregate-spb-iw}
\begin{algorithmic}[1]
\REQUIRE number of arms $K$, cost $c>0$, $\alpha\in[1/2,1)$,
parameters $\beta_1>0$ and $\bar\beta>0$ satisfying the conditions of
\Cref{thm:exact-iw-aggregate-spb}, and exact access to a finite-support context
distribution $\cD$ on $\cX$.
\STATE Set $\bar\alpha \leftarrow 1 - \alpha$ and $H_\rho \spr*{q} := \rho^{-1} \sumaK \spr*{q_a^\rho - q_a}$.
\STATE Set $\Sigma := \bbE_{X \sim \cD} \sbr*{X X\transpose}$, and $B_\star := \max_{x, y \in \cX} \abs*{x\transpose \Sigma^{-1} y}$.
\STATE Initialize $\widehat\Theta_{0,a}^{\mathtt{IW}} \leftarrow 0 \in \bbR^d$ for all $a \in \sbr*{K}$.
\FOR{$t = 1, 2, \ldots$}
    \FOR{each $x \in \cX$}
        \STATE
        \vspace{-5mm}
        \begin{align*}
        	\spr{
        		q_t(\cdot\mid x),
        		h_t(x),
        		I_t(x),
        		s_t(x),
        		(z_{t,a}(x))_{a\in[K]},
        		Z_t(x)
        	}\leftarrow
        	\textsc{Tsallis-FTRL}
        	\spr{
        		(\widehat\Theta_{t-1,a}^{\mathtt{IW}})_{a\in[K]},
        		x,\beta_t,\alpha,\bar\beta
        	}.
        \end{align*}
    \ENDFOR
    \STATE Set $\htmax \leftarrow \max_{x \in \cX} H_\alpha \spr*{q_t \spr*{\cdot \given x}}$. $\qquad$ (maximum entropy over contexts)
    \STATE Set $z_t \leftarrow \frac{4 c K}{1 - \alpha} \bbE_{X \sim \cD} \sbr*{X\transpose \Sigma^{-1} X Z_t \spr*{X}}$. $\qquad$ (aggregate stability term)
    \STATE Set $u_t^{\mathrm{mag}} \leftarrow \frac{8 c K B_\star}{1 - \alpha} \spr*{\max_{x \in \cX}s_t \spr*{x}}^{1 - \alpha}$ $\qquad$ (maximal magnitude)
    \STATE Set $\widetilde u_t \leftarrow u_t^{\mathrm{mag}} + z_t / \spr*{cK}$. $\qquad$ (clipping-aware magnitude)
    \STATE Set the SPB-matching parameters
    \begin{equation*}
        \gamma_t \leftarrow \min \scbr*{\sqrt{\frac{z_t}{\beta_t}} + \frac{u_t^{\mathrm{mag}}}{\beta_t}, c K}, \qquad p_t \leftarrow \frac{\gamma_t}{cK}, \qquad b_t \leftarrow 2\sqrt{\frac{z_t}{\beta_t}} + \frac{\widetilde u_t}{\beta_t}.
    \end{equation*}
    \STATE Observe context $X_t$ and sample $A_t \sim q_t \spr*{\cdot \given X_t}$.
    \STATE Independently include each arm $a \in \sbr*{K}$ in $O_t$ with probability $p_t$.
    \STATE Incur mean loss $\ell_t \spr*{X_t, A_t} + c \abs*{O_t}$ and observe $Y_t(a)$ for all $a \in O_t$.
    \FOR{$a = 1, \ldots, K$}
        \STATE Set $\thetataiw \leftarrow \mathds{1} \spr*{a \in O_t} p_t^{-1} \Sigma^{-1} X_t\,Y_t(a)$.
        \STATE Update $\Thetataiw \leftarrow \widehat\Theta_{t-1, a}^{\mathtt{IW}} + \thetataiw$.
    \ENDFOR
    \IF{$\htmax > 0$}
        \STATE Update
        \begin{equation*}
            \beta_{t+1} \leftarrow \beta_t + \frac{b_t}{2 \htmax}.
        \end{equation*}
    \ELSE
        \STATE Set $\beta_{t+1} \leftarrow \beta_t$.
    \ENDIF
\ENDFOR
\end{algorithmic}
\end{algorithm}

\begin{algorithm}[ht]
	\caption{Cumulative-Entropy SPB--matching with FTRL (\loc)}
	\label{alg:kuroki-style}
	\begin{algorithmic}[1]
		\REQUIRE number of arms $K$, cost $c \ge 0$, $\alpha \in (0,1)$,
		parameters $\Lambda_0,\beta_1,\zeta,\xi,\nu,u$ satisfying the conditions
		of \Cref{thm:le-spb}, and exact access to a finite-support context
		distribution $\cD$ on $\cX$.
		\STATE Set
		\[
		\Sigma \leftarrow \bbE_{X\sim\cD}[XX\transpose],
		\qquad
		B_\star \leftarrow
		\max_{x,y\in\cX}\abs{x\transpose\Sigma^{-1}y}.
		\]
		\STATE Initialize
		$\widehat\Theta_{0,a}^{\mathtt{IW}}\leftarrow 0\in\bbR^d$
		for all $a\in[K]$.
		\FOR{$t=1,2,\ldots$}
		\FOR{each $x\in\cX$}
\STATE
\vspace{-5mm}
\begin{align*}
	\spr{
		q_t(\cdot\mid x),
		h_t(x),
		I_t(x),
		s_t(x),
		(z_{t,a}(x))_{a\in[K]},
		Z_t(x)
	}\leftarrow
	\textsc{Tsallis-FTRL}
	\spr{
		(\widehat\Theta_{t-1,a}^{\mathtt{IW}})_{a\in[K]},
		x,\beta_t,\alpha,\bar\beta=0
	}.
\end{align*}		
\FOR{$a=1,\ldots,K$}
\IF{$\nu=0$}
\STATE Set $p_{t,a}(x)\leftarrow1$.
\ELSE
\STATE Set
\[
p_{t,a}(x)
\leftarrow
\min\left\{
1,
\frac{z_{t,a}(x)}
{\nu\Lambda_{t-1}^{1/3}}
+\frac{u}{\beta_t}
\right\}.
\]
\ENDIF
\ENDFOR
		\ENDFOR
		\STATE Observe $X_t$ and sample
$A_t\sim q_t(\cdot\mid X_t)$ .
				\STATE Set
		$\Lambda_t\leftarrow\Lambda_{t-1}+h_t(X_t) \qquad \text{(update the cumulative entropy)}$.
		\STATE Update
		\[
		\beta_{t+1}
		\leftarrow
		\beta_t
		+\zeta\Lambda_t^{-1/2}
		+\xi\Lambda_t^{-1/3}.
		\]
		
		\STATE Independently include each $a\in[K]$ in $O_t$
		with probability $p_{t,a}(X_t)$.
		\STATE Incur mean loss
		$\ell_t(X_t,A_t)+c|O_t|$.
		\FOR{$a \in O_t$}
		\STATE  Observe $Y_t(a)$, and set
		\[
		\Sigma_{t,a}
		\leftarrow
		\bbE_{X\sim\cD}
		[p_{t,a}(X)XX\transpose], \]
				\[
		\widehat\theta_{t,a}^{\mathtt{IW}}
		\leftarrow
		\Sigma_{t,a}^{-1}X_tY_t(a).
		\]
		\STATE Update
		\[
		\widehat\Theta_{t,a}^{\mathtt{IW}}
		\leftarrow
		\widehat\Theta_{t-1,a}^{\mathtt{IW}}
		+
		\widehat\theta_{t,a}^{\mathtt{IW}}.
		\]
		
		\ENDFOR
		\FOR{$a \notin O_t$}
		\STATE Update
		\[
		\widehat\Theta_{t,a}^{\mathtt{IW}}
		\leftarrow
		\widehat\Theta_{t-1,a}^{\mathtt{IW}}
		.		\]
		\ENDFOR
		\ENDFOR
	\end{algorithmic}
\end{algorithm}

\begin{algorithm}[ht]
	\caption{Tsallis-FTRL local routine}
	\label{alg:tsallis-ftrl-routine}
	\begin{algorithmic}[1]
		\REQUIRE cumulative estimates
		$\spr*{\widehat\Theta_a}_{a\in[K]}$,
		context $x$, inverse learning rate $\beta$,
		parameters $\alpha\in(0,1)$ and $\bar\beta\ge0$.
		\STATE Set $\bar\alpha \leftarrow 1-\alpha$.
		\STATE Compute
		\begin{equation*}
			q \leftarrow
			\argmin_{p\in\Delta_K}
			\scbr{
				\sumaK p_a \inp{x,\widehat\Theta_a}
				-\beta H_\alpha(p)
				-\bar\beta H_{\bar\alpha}(p)
			}.
		\end{equation*}
		\STATE Set $h \leftarrow H_\alpha(q)$.
		\STATE Set $I \leftarrow \argmax_{a\in[K]} q_a$.
		\STATE Set $s \leftarrow \min\{q_I,1-q_I\}$.
		\FOR{$a=1,\ldots,K$}
		\STATE Set
		\begin{equation*}
			z_a \leftarrow
			\begin{cases}
				q_a^{2-\alpha}, & a\neq I,\\
				s^{2-\alpha}, & a=I.
			\end{cases}
		\end{equation*}
		\ENDFOR
		\STATE Set $Z\leftarrow\sum_{a=1}^K z_a$.
		\STATE \textbf{return}
		$(q,h,I,s,(z_a)_{a\in[K]},Z)$.
	\end{algorithmic}
\end{algorithm}

	\subsection{Implementation remarks for the exact-IW algorithms}
\label{app::implem}

The regret guarantees in this work are stated for exact
importance-weighted estimators. In particular, the algorithms assume
exact access to the distributional quantities entering their observation
rules and estimators. This includes $\Sigma^{-1}$ (for \agg), $B_\star$ or a valid
upper bound thereof, and the expectations over $X\sim\cD$ appearing in
the algorithms. The purpose of this appendix is to clarify the
computational implications of these assumptions. Approximation errors
are not covered by our theoretical guarantees.

For concreteness, consider first a finite-support distribution
$\cD$ over
$\cX=\{x_1,\ldots,x_M\}$. In this case, all expectations below reduce
to finite sums. For more general context distributions, the same exact
algorithms require an oracle capable of evaluating the corresponding
expectations.

\paragraph{\agg.}
The main computational advantage of
Algorithm~\ref{alg:clipped-aggregate-spb-iw} is that its observation
probability is independent of both the arm and the context. Each arm is
observed with probability $p_t$, and therefore the only matrix inverse required by the estimator is the fixed
matrix $\Sigma^{-1}$, which can be computed once before the interaction.

The time-dependent cost instead comes from the aggregate SPB
quantities. At every round, the FTRL policy and its associated local
statistics must be evaluated over the support of $\cD$ in order to
compute
\begin{equation*}
	z_t
	=
	\frac{4cK}{1-\alpha}
	\bbE_{X\sim\cD}
	\left[
	X\transpose\Sigma^{-1}X\,Z_t(X)
	\right],
\end{equation*}
as well as
\begin{equation*}
	m_t
	=
	\max_{x\in\cX}s_t(x),
	\qquad
	h_t^{\max}
	=
	\max_{x\in\cX}h_t(x).
\end{equation*}
Thus, in the finite-support implementation, \agg\ requires one
support-wide evaluation of the Tsallis-FTRL statistics per round, but
no time-varying matrix inversion. Once these quantities are available,
$u_t^{\mathrm{mag}}$, $\widetilde u_t$, $\gamma_t$, and $b_t$ are
scalar computations.

Notice that the exact maxima $m_t$ and $h_t^{\max}$ play a structural
role in the analysis: they provide deterministic upper bounds used in
the magnitude and entropy-growth conditions. Consequently, replacing
the support-wide computations by Monte Carlo estimates is not
immediate, since underestimating these quantities may invalidate the
one-sided inequalities used by the SPB argument.

\paragraph{\loc.}
The exact implementation of \loc\ requires weighted covariance
inversions, but only for arms that are actually observed. Indeed, if
$a\notin O_t$, then
$\widehat\theta_{t,a}^{\mathtt{IW}}=0$, so neither
$\Sigma_{t,a}$ nor its inverse needs to be formed at round $t$.
Consequently, the number of time-varying covariance constructions and
inversions at round $t$ is $|O_t|$, with conditional expectation
\begin{equation*}
	\bbE_t[|O_t|]
	=
	\bbE_{X\sim\cD}
	\left[
	\sum_{a=1}^K p_{t,a}(X)
	\right].
\end{equation*}
In the worst case this can still be of order $K$, but in stochastic
regimes it decreases together with the observation rate.

Additionally, when $c=0$ then
$p_{t,a}(x)=1$ for every $(t,a,x)$, so that
$\Sigma_{t,a}=\Sigma$ for all arms and rounds, and the single fixed
inverse $\Sigma^{-1}$ can again be reused.

Thus, with exact distributional information, \loc\ is generally more
computationally demanding than \agg\ in the adversarial paid-observation regime, but this cost can be milder in the stochastic regime.

\paragraph{Sampling-based approximations.} However, as discussed in the main paper, the matrix geometric resampling (MGR) technique used in previous works \citep{NeuO20, BOBWlinear} could replace the exact-distribution access by sampling access to the context distribution $\cD$. Therefore, an estimate of
$\Sigma_{t,a}^{-1}$ could be obtained without explicitly forming the covariance matrix. In both works, $O(\log(T))$ samples (up to dimension-based factors) are sufficient to make the bias induced by sampling negligible in the final order of the regret bounds.

Such an approximation introduces truncation and estimation errors. We expect these
additional terms to be controlled using arguments similar to those of
\citet{BOBWlinear}, together with an appropriate retuning of the
algorithmic parameters, but we do not pursue this extension here.

For \agg, matrix-geometric resampling is less directly relevant:
the estimator already depends only on the fixed matrix $\Sigma^{-1}$.
The main oracle requirements are instead the aggregate expectation
defining $z_t$ and the worst-context quantities $m_t$ and
$h_t^{\max}$. Approximating these quantities while preserving the
one-sided SPB and magnitude conditions requires a different sensitivity
analysis, which is not covered by previous works.

\paragraph{Memory.}
Both algorithms maintain the cumulative parameter estimates
$\widehat\Theta_{t,a}^{\mathtt{IW}}\in\bbR^d$ for every arm, requiring
$O(Kd)$ persistent memory. In addition, \agg\ stores the fixed matrix
$\Sigma^{-1}$, requiring $O(d^2)$ memory. An exact implementation of
\loc\ may additionally store the at most $K$ current covariance matrices of observed arms and
their inverses, requiring $O(Kd^2)$ working memory, although these
matrices can in principle be processed sequentially at the expense of
additional computation.

For a finite-support distribution, storing the context support itself
requires an additional $O(Md)$ memory, together with its probability
weights. There is no need to retain past per-round IW estimates:
maintaining their cumulative sums is sufficient for FTRL, so the memory
requirement does not grow with $T$.
	
\end{document}